\documentclass[superscriptaddress, onecolumn, nofootinbib]{revtex4-2}
\makeatletter

\usepackage{amsmath,amsfonts,bm}

\def\eqref#1{equation~\ref{#1}}

\def\1{\bm{1}}

\DeclareMathAlphabet{\mathsfit}{\encodingdefault}{\sfdefault}{m}{sl}
\SetMathAlphabet{\mathsfit}{bold}{\encodingdefault}{\sfdefault}{bx}{n}

\usepackage{amsmath,amssymb, amsthm, mathtools}
\usepackage{url}
\usepackage{bm}
\usepackage{braket}
\usepackage{xcolor}
\definecolor{crimson}{RGB}{165,28,48}
\usepackage[colorlinks=true,allcolors=crimson]{hyperref}
\usepackage{microtype}
\usepackage{tcolorbox}

\newcommand{\rsgd}{r_{\rm SGD}}
\newcommand{\neff}{\eta_{\rm eff}}
\newcommand{\neffi}{{\eta_{\rm eff}^\infty}}
\newcommand{\hS}{{\hat{\mathcal{S}}}}
\newcommand{\hB}{{\hat{\mathcal{B}}}}
\newcommand{\hK}{{\hat{\mathcal{K}}}}
\newcommand{\hT}{{\hat{\mathcal{T}}}}
\newcommand{\hR}{\hat{R}_3}

\DeclareMathOperator*{\tr}{Tr}

\usepackage{graphicx}

\begin{document}

\title[A Solvable Model of Adaptive Learning Rate Rescaling: Acceleration, Stability \& Scaling]{A Solvable Model of Adaptive Learning Rate Rescaling: Acceleration, Stability \& Scaling}

\author{Itay Lavie}
\email{itaylavie@g.harvard.edu}
\affiliation{John A. Paulson School of Engineering and Applied Sciences, Harvard University, Cambridge, MA}

\author{Clarissa Lauditi$^\dagger$}
\affiliation{John A. Paulson School of Engineering and Applied Sciences, Harvard University, Cambridge, MA}

\author{Cengiz Pehlevan$^\dagger$}
\affiliation{John A. Paulson School of Engineering and Applied Sciences, Harvard University, Cambridge, MA}
\affiliation{Center for Brain Science, Harvard University, Cambridge, MA}
\affiliation{Kempner Institute for the Study of Natural and Artificial Intelligence, Harvard University, Cambridge, MA}

\begin{abstract}
A recurring design principle in modern optimizers is to decouple update magnitude from the raw gradient norm, yet its consequences for learning-curve and resource scaling remain unclear. We isolate this mechanism by studying normalized SGD in a random-feature model with power-law teacher and data covariance. Fixed-norm updates induce an effective learning rate that grows as gradients shrink. We derive a dynamical mean-field theory (DMFT) describing the joint dependence of the loss on training time, model width and batch size. Normalization initially accelerates SGD, mapping the power-law exponent $r_{\rm SGD}<1$ to $2r_{\rm SGD}/(1-r_{\rm SGD})$, with exponential convergence at $r_{\rm SGD}=1$ and formal finite-time convergence for $r_{\rm SGD}>1$. At finite step size, however, the same feedback ultimately breaks the acceleration and leads to marginal stability. The late-time theory yields width-limited, edge-of-stochastic-stability (EoSS), and deterministic edge-of-stability (EoS) regimes. These phases determine when larger batches or wider models reduce serial training time at comparable compute. We quantify in which of these phases increased batch size or width can compensate the excess compute use per step by fewer optimization steps to target loss. Linearized ResNet experiments on CIFAR-5M support the predicted acceleration, breakdown, and resource-scaling trends. Together, these results connect normalization-induced acceleration, EoS effects, and width--batch allocation within a solvable theory.
\end{abstract}

\maketitle
\begingroup
\renewcommand{\thefootnote}{\fnsymbol{footnote}}
\footnotetext[2]{Equal senior authorship}
\endgroup

\section{Introduction}
Optimization governs how efficiently neural networks convert training compute into improved performance, yet its role in shaping neural scaling laws remains incompletely understood. These laws describe how performance improves with training time, model size, and data, providing a quantitative basis for allocating compute across these resources~\citep{hoffmann2022training,kaplan2020scaling}. Several theoretical studies connect these laws to the spectral structure of data and tasks: learning proceeds progressively through the data spectrum, while finite training time, model size, and data impose different spectral cutoffs~\citep{bahriExplainingNeuralScaling2024,maloneySolvableModelNeural2022,bordelon2020spectrum,bordelonDynamicalModelNeural2024,paquette4+3PhasesComputeOptimal2024}. Only recently have studies begun to examine how these spectral learning dynamics interact with optimizer design~\citep{paquettePhasesMuonWhen2026,xiaoExactRiskCurves2025}, and many aspects of this interaction remain poorly understood.

Modern adaptive optimizers such as RMSProp, Adam, SOAP, and Muon induce trajectory-dependent effective learning rates through gradient-dependent preconditioning, even when the base learning rate is fixed~\citep{TielemanHinton2012, kingmaAdamMethodStochastic2014, vyasSOAPImprovingStabilizing2025,jordan2024muon}. This rescaling changes both update direction and magnitude; grafting makes their distinct roles explicit by combining the direction of one optimizer with the magnitude of another \citep{agarwalDisentanglingAdaptiveGradient2020}. We isolate the feedback between update magnitude and learning dynamics by studying normalized SGD, which preserves the stochastic-gradient direction while fixing the update norm. With the normalization used below, this is equivalent to SGD with an endogenous scalar learning rate $\eta_{\mathrm{eff}}(t)=\eta\ /|\mathbf g(t)\|_2$, providing a tractable setting for studying how adaptive rescaling affects convergence and stability.

Gradient normalization creates a tension between acceleration and stability. As progressively weaker spectral modes are learned, the gradient norm decays and the effective learning rate grows, potentially accelerating progress through the spectral tail. At finite step size, however, the same growth eventually brings the leading spectral mode to the discrete-time stability threshold.  The deterministic edge-of-stability (EoS) corresponds to marginal stability of the leading Hessian mode~\citep{cohen2021gradient}, while the edge-of-stochastic-stability (EoSS) is associated with a stability limit imposed by minibatch fluctuations~\citep{andreyev2024edge}. The spectral tail therefore controls the transient acceleration, while the spectral edge controls its breakdown. 

These stability limits also shape the computational value of increasing training resources. A larger batch can support a larger effective learning rate, but each update is more expensive~\citep{mccandlish2018empirical,shallueMeasuringEffectsData2019,zhang2025does,meterez2026defense}; increasing width changes both approximation capacity and the spectrum that controls stability. Understanding these trade-offs requires distinguishing gains in effective step size from gains per unit compute. Our model makes this distinction explicit and connects the resulting predictions to experiments with linearized ResNets on CIFAR-5M~\citep{nakkiranDeepBootstrapFramework2021}, using spectral exponents measured from the neural tangent kernel and target alignment.

We study a power-law random-feature model, which captures key features of empirically observed neural scaling laws while remaining analytically tractable~\citep{maloneySolvableModelNeural2022,bordelonDynamicalModelNeural2024}. Within this setting, we ask:
\begin{tcolorbox}[colframe=crimson, opacityback=0.95] 
\textit{How does adaptive learning rate shapes acceleration, and how do capacity and stability limits determine the benefits of increasing batch size and model width?}
\end{tcolorbox}
Our main contributions are:

\begin{itemize}
    \item We derive a dynamical mean field theory (DMFT) for normalized SGD in a high-dimensional power-law random-feature model, characterizing the joint scaling of the loss with training time, model width, batch size, and finite step size.
    \item We characterize source- and capacity-dependent transient scaling of norm-SGD: normalization accelerates the SGD exponent $r_{\mathrm{SGD}}$ to $2r_{\mathrm{SGD}}/(1-r_{\mathrm{SGD}})$ for $r_{\mathrm{SGD}}<1$, while $r_{\mathrm{SGD}}=1$ and $r_{\mathrm{SGD}}>1$ yield exponential and finite-time convergence, respectively. 
    \item DMFT yields a spectrum-dependent width-batch phase diagram: in overparameterized models, a width-dependent critical batch separates stochastic edge-of-stability (EoSS) from deterministic edge-of-stability (EoS), whereas underparameterized models are limited by finite-width error.
    
    \item We characterize compute--speed trade-offs across batch size and width: in the EoSS phase, a nearly linear batch-scaling regime reduces the steps to a target loss at almost unchanged compute. In the width-limited phase, we identify spectral regimes where the stationary effective learning rate grows superlinearly with width, giving rise to potential width speedups.
    
    \item We test the theory using linearized ResNets on CIFAR-5M. Measured kernel and target exponents yield predictions consistent with the accelerated loss decay and its crossover toward SGD-like scaling, while batch and width sweeps support the predicted efficiency trends in the explored regime.
    
\end{itemize}

We use norm-SGD to isolate how the feedback between gradient normalization and learning dynamics shapes convergence and stability. Our aim is to understand this mechanism rather than advocate norm-SGD as a practical optimizer.

\subsection{Related Work}
\paragraph{Theories of Neural Scaling Laws.} Empirical scaling laws have been documented across several domains~\citep{hestness2017deep}, quantifying the dependence of loss on model size, data, and compute~\citep{kaplan2020scaling,hoffmann2022training}. Theoretical explanations derive power-law exponents from kernel spectra and target alignment, or power-law random feature models~\citep{spigler2020asymptotic,bordelon2020spectrum,canatar2021spectral,bahriExplainingNeuralScaling2024,maloneySolvableModelNeural2022} or hierarchal data model~\citep{cagnetta2026deriving}; with recent works extending the spectral framework to training dynamics, compute-optimal regimes and feature learning ~\citep{bordelonDynamicalModelNeural2024,paquette4+3PhasesComputeOptimal2024,bordelonHowFeatureLearning2024}. 

\paragraph{Optimizers and Scaling.} Whether optimizer gains persist with model size and compute has itself become a scaling question. Comparisons among AdamW, Shampoo, SOAP and Muon show that reported gains depend on tuning, model scale, and training horizon~\citep{loshchilov2017decoupled,gupta2018shampoo,liu2025muon,magakyan2026spectral,wen2026fantastic,harariAdamSOAPMuon2026}. Time dependent batch-size and learning rate schedules further shape finite-time scaling through update scale and gradient noise, batch-size schedules, and coupled schedules~\citep{smith2017don,mccandlish2018empirical,defazio2023and,zhang2025does,tissue2026scaling,meterezSeesawAcceleratingTraining2026,meterez2026anytime}. High-dimensional theory further resolves optimizer-dependent preconditioning and batch effects~\citep{xiaoExactRiskCurves2025,paquettePhasesMuonWhen2026}. Grafting and norm-based views separate update direction from magnitude~\citep{agarwalDisentanglingAdaptiveGradient2020,bernsteinOldOptimizerNew2024}. Following this decomposition, we study gradient normalization to isolate magnitude control while leaving the SGD direction unchanged.

\vspace{-4pt}
\paragraph{Critical Batch Size and Compute Efficiency.}
Increasing batch size trades more computation per update for fewer serial updates, with diminishing returns beyond a problem-dependent ``critical batch size''~\citep{mccandlish2018empirical,shallueMeasuringEffectsData2019}. In LLM pretraining, prior work showed that critical batch size scales primarily with training data size rather than model size~\citep{zhang2025does}. Batch-size schedules can also replace learning-rate decay while preserving similar learning curves~\citep{smith2017don}; Seesaw develops this connection theoretically for noisy linear regression and extends it to normalized SGD under a variance-dominated approximation~\citep{meterezSeesawAcceleratingTraining2026}. Spectral analyses provide complementary perspective: high-dimensional theory identifies a spectrum-dependent batch threshold beyond which the convergence rate of SGD with momentum saturates~\citep{leeTrajectoryMiniBatchMomentum2022}. We study these thresholds and limits in normalized SGD, quantifying spectrum-dependent departures from linear effective-step scaling and separating the resulting efficiency consideration from the EoSS-EoS phase transition at which the effective learning rate freezes.

\paragraph{Training Stability and large step size effects.} Large learning rates can lead to faster optimization, while driving training toward the boundary of discrete-time stability. On real objectives, GD often exhibits progressive sharpening, followed by edge-of-stability dynamics with persistent oscillations along sharp directions~\citep{cohen2021gradient,aroraUnderstandingGradientDescent2022,cohen2022adaptive,cohenUnderstandingOptimizationDeep2024}. In SGD, sampling fluctuations modify the stability criterion, producing an edge of stochastic stability governed by batch-dependent directional curvature and related critical step-size transitions~\citep{xing2018walk,mulayoffExactMeanSquare2024,jastrzebski2020break,andreyev2024edge,paquetteSGDLargeAveragecase2021,agarwalaHighDimensionalAnalysis2025}. Because curvature determines the stable learning-rate range, its scaling with model size is also central to optimal learning-rate transfer. The $\mu$P framework establishes such transfer across width and depth~\citep{yang2021tensor,yang2021tuning,bordelon2022self,yang2024tensor,bordelon2024depthwise,chizat2025hidden,dey2026don}, with recent work seeking mechanistic explanations for it through width-consistent sharpness and top spectral modes~\citep{noci2024super,ghosh2026understanding,lauditi2026spectral}. 

\section{Model}
We study online normalized SGD in a power-law random-feature model commonly used in theories of neural scaling laws~\citep{maloneySolvableModelNeural2022,bahriExplainingNeuralScaling2024,bordelonDynamicalModelNeural2024,paquette4+3PhasesComputeOptimal2024, bodin2021model}. At each step $t$, a fresh batch of zero-mean inputs $\{\bm \psi_\mu^t\}_{\mu=1}^{B}\in\mathbb{R}^d$ is drawn independently with covariance $\mathbb{E}\left[\bm\psi_\mu^t\bm\psi_\rho^{s\top}\right]
=\delta_{\mu\rho}\delta_{ts}\bm \Lambda$, where $\Lambda_{kk}=\lambda_k=k^{-b}$.
A frozen random matrix $\bm A\in\mathbb{R}^{N\times d}$ projects the inputs onto $N$ features, and $\bm w^t\in\mathbb{R}^N$ is trained on top of them. The target coefficients obey $\beta_k^2=k^{-2a}$ in the covariance eigenbasis, with $2a+b>1$ ensuring finite initial loss. Denoting the gradient by $\bm g$, the loss, predictor, and weight updates are
\begin{align}
    L^t& :=\frac{1}{2B}\sum_{\mu=1}^B \left(\bm \beta^\top\bm \psi^t_{\mu}-f^t\right)^{2},
    &
    f_\mu^t &:=\frac{1}{\sqrt{N}}\bm w^{t\top}\bm A\bm \psi_{\mu}^t,
    &
    \bm w^{t+1}&=\bm w^{t}-\eta \sqrt{d}\left(\bm g^{\top}\bm g\right)^{-1/2}\bm g.
\label{eq:model_def}
\end{align}
We take the proportional limit $d,N,B\to\infty$ at fixed width and batch ratios $\nu:= \tfrac{N}{d}$ and $\alpha:= \tfrac{B}{d}$.

In this limit, we use DMFT to reduce the high-dimensional dynamics to effective stochastic processes for individual spectral modes~\citep{sompolinsky1981dynamic,de1978dynamics,crisanti2018path,mignacco2020dynamical,heliasStatisticalFieldTheory2020}. Their correlation and response functions, random at finite $d,N,B$, concentrate to deterministic limits. Equal-time correlations determine observables such as the loss and squared gradient norm $C_3^{tt}$, hence $\eta_{\mathrm{eff}}^t=\eta/\sqrt{\nu C_3^{tt}}$, while responses encode how perturbations propagate across steps. 
The derivation and closed set of equations are given in App.~\ref{app:dmft}.

\section{Continuous Time Dynamics}
\begin{figure}[t]
    \centering
    \includegraphics[width=1\linewidth]{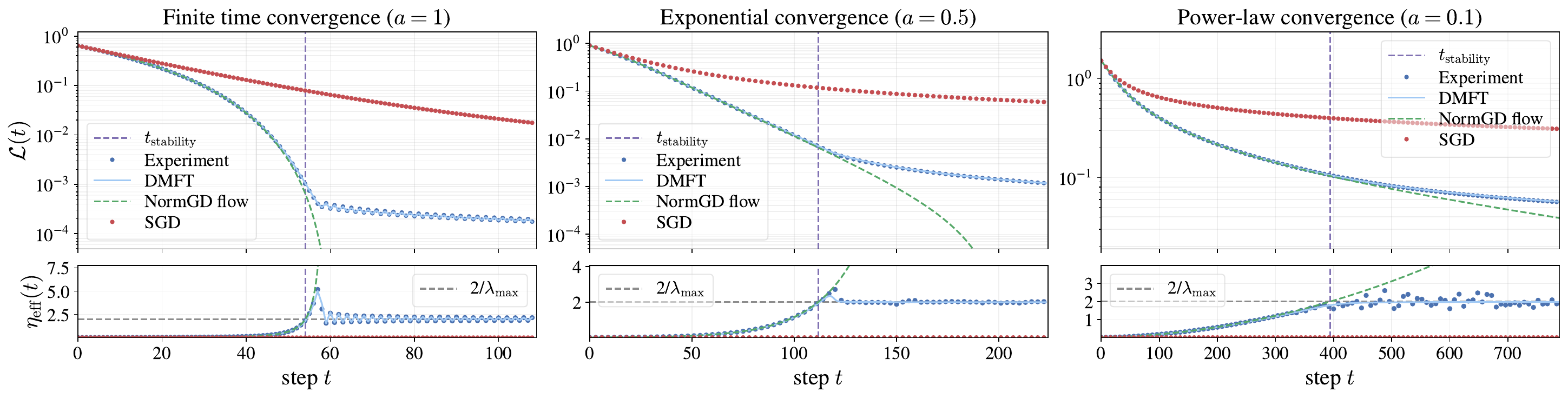}
    \caption{\textbf{Three transient convergence regimes.} Loss (\textbf{top}) and effective learning rate (\textbf{bottom}) at large $\alpha,\nu$. Before $t_{\mathrm{stab}}$, simulations and DMFT follow the continuous-time prediction: formal finite-time convergence for $a>1/2$ (\textbf{left}), exponential for $a=1/2$ (\textbf{center}), and accelerated power-law convergence for $a<1/2$ (\textbf{right}). SGD uses $\eta_{\mathrm{SGD}}=\eta_{\mathrm{eff}}(0)$ to match the initial update norm. At $t_{\mathrm{stab}}$, $\eta_{\mathrm{eff}}$ reaches $2/\lambda_{\max}$ and the discrete dynamics depart from the flow prediction.}
    \label{fig:normGF}
\end{figure}
We first isolate the transient dynamics from the finite-width and minibatch effects that govern late-time behavior. As $\eta\to 0$, normalized SGD approaches $\partial_t  \bm w(t) = - \sqrt{d}\left(\bm g^{\top} (t)\bm g(t)\right)^{-1/2}\bm g (t)$. These updates are parallel to ordinary gradient flow (GF) everywhere but have a different magnitude. Indeed, defining a new clock $\tau$ through $\frac{d \tau}{dt} :=  \left( \frac{g^{\top}(t) g(t)}{d} \right)^{-1/2}$ gives $\partial_\tau\bm w(\tau)=-\bm g(\tau)$. In this limit, normalization is therefore equivalent to reparameterizing time, with the effective learning rate increasing as the gradient norm decays. Consequently, it changes the temporal scaling of the loss while leaving the width- and data-limited exponents unchanged from GF (see e.g.~\citep{bordelonDynamicalModelNeural2024}). To extract this temporal scaling, we subsequently take $\alpha,\nu\to\infty$. In the $\tau$-clock, the residual in mode $k$ decays as $\exp(-\lambda_k\tau)$, so its relaxation time is $\tau_k\sim\lambda_k^{-1}\sim k^b$. At time $\tau$, modes $k\lesssim k_\tau\sim\tau^{1/b}$ have been learned, leaving the loss dominated by the unresolved tail. Since each such mode contributes $\lambda_k\beta_k^2\sim k^{-(2a+b)}$, summing this tail gives $L(\tau)\sim k_\tau^{-(2a+b-1)}\sim\tau^{-r_{\mathrm{SGD}}}$, where $r_{\mathrm{SGD}}=(2a+b-1)/b$. Reexpressing the same trajectory in physical time gives
\begin{equation}
    L\left(t\right)\sim\begin{cases}
\left(T-t\right)^{2\frac{\rsgd}{\rsgd-1}} & a>1/2\\
e^{-2\sqrt{b}t} & a=1/2\\
t^{-2\frac{\rsgd}{1-\rsgd}} & a<1/2
\end{cases}
\end{equation}
(see Appendix~\ref{app:transient_normSGD} for details). Thus, normalization produces accelerated power-law convergence for $a<1/2$, exponential convergence at $a=1/2$, and formal finite-time convergence for $a>1/2$. Figure~\ref{fig:normGF} verifies these predictions at large but finite $\alpha,\nu=10$. Defining $t_{\mathrm{stab}}$ as the first step in which the top mode loses stability $\neff > 2/ \lambda_{\rm max}$ we see that for $t<t_{\mathrm{stab}}$, simulations and DMFT track the continuous-time prediction and outperform constant learning-rate SGD, whose learning rate is set to $\eta_{\mathrm{eff}}(0)$ to match the initial update norm. The lower panels show $\eta_{\mathrm{eff}}(t)$ growing toward $2/\lambda_{\max}$: once this stability threshold is reached, the discrete dynamics depart from the continuous-time prediction and enter the stability-controlled regime analyzed in the next section. Notably, $t_{\mathrm{stab}}$ itself can be found within the flow framework, as up to that time the trajectory is still well approximated by the flow. 

\section {Discrete-Time Dynamics: Edge of Stability and Breakdown of the Continuous Description}
At finite step size, the effective learning rate grows as the gradient decays until finite width, minibatch fluctuations, or deterministic instability cut off the acceleration. Figure~\ref{fig:normGF} shows $\eta_{\mathrm{eff}}^t$ approaching a constant or stationary mean, denoted by $\neffi$. Norm-SGD then has the same late-time update rule as SGD with constant step size $\neffi$, with $\neffi$ selected autonomously by the dynamics through three different mechanisms, as we show below. Because the feature map is untrained, the curvature is time-independent; therefore the response to perturbations is stationary, and we  seek time-translation-invariant (TTI) solutions.

\begin{figure}[t]
    \centering
    \includegraphics[width=0.31\linewidth]{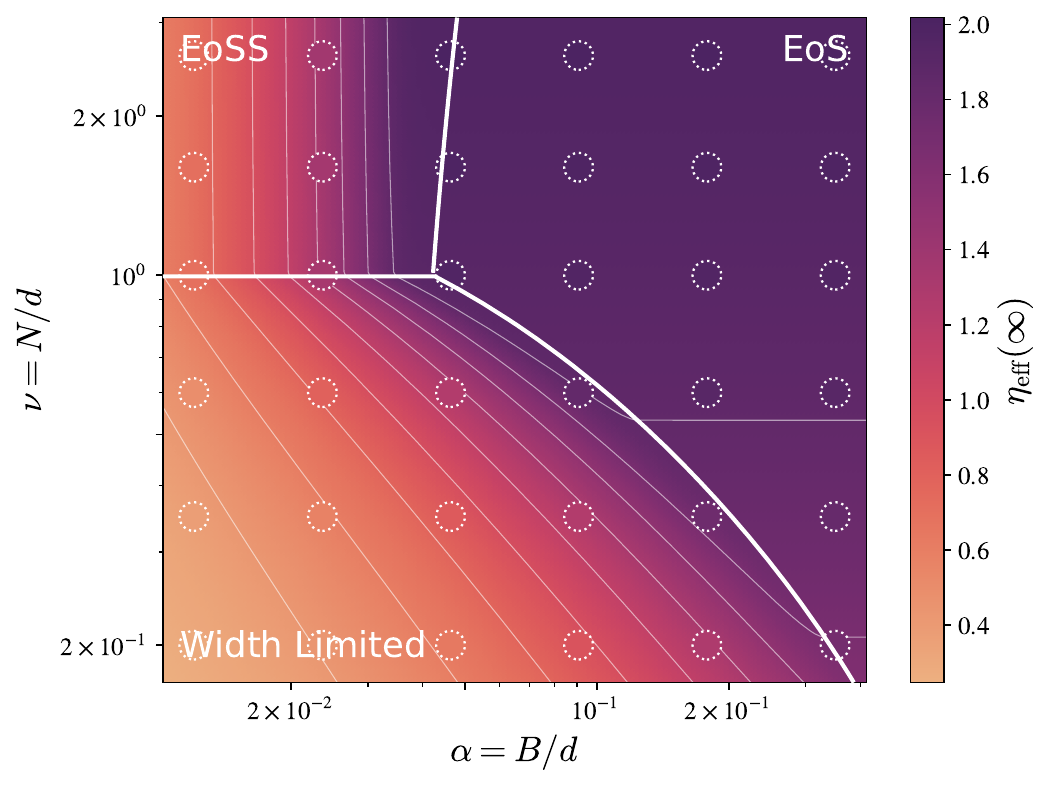}
    \includegraphics[width=0.33\linewidth]{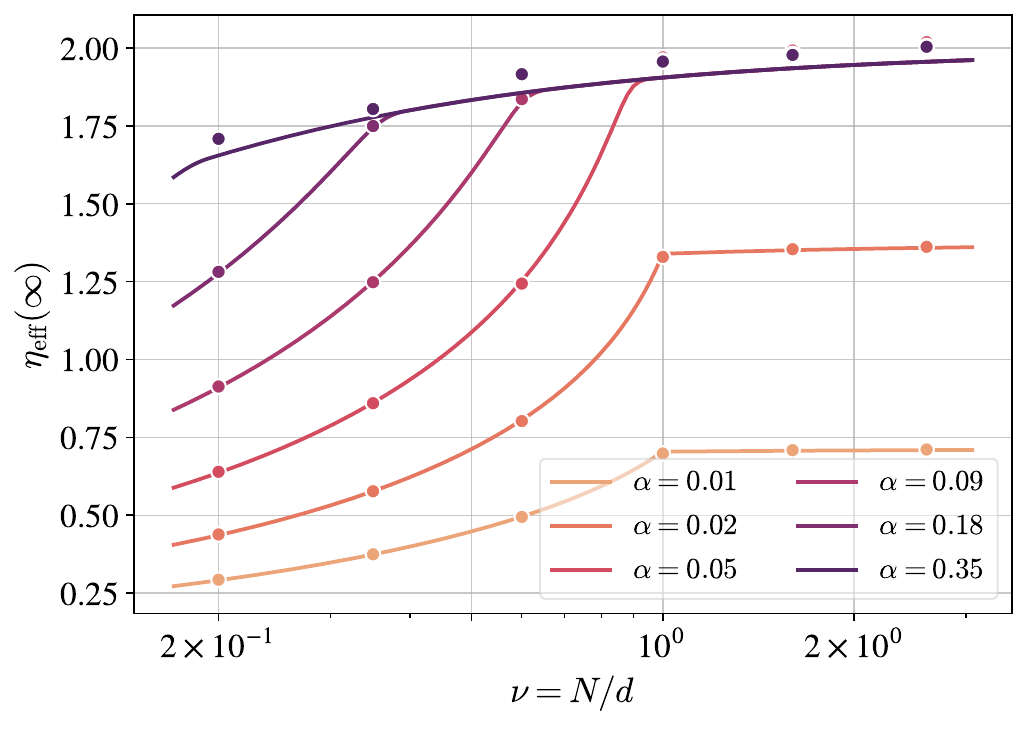}
    \includegraphics[width=0.33\linewidth]{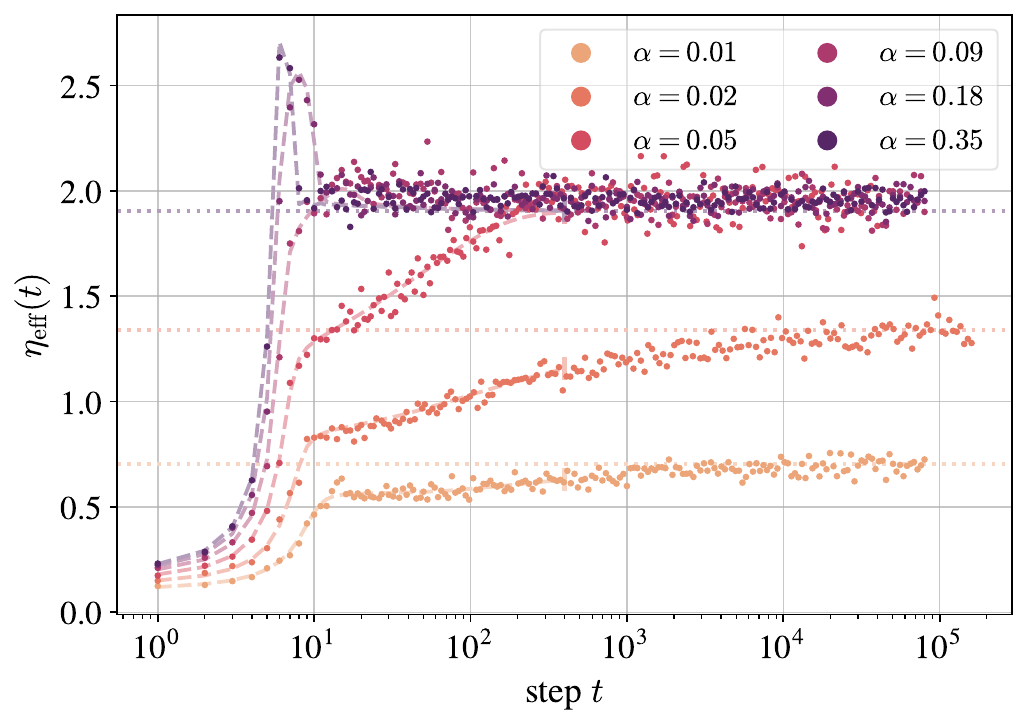}
    \caption{\textbf{Three late-time regimes: width limited, EoSS, and EoS.} \textbf{Left:} DMFT phase diagram in the $(\alpha,\nu)$ plane. Color denotes $\eta_{\mathrm{eff}}^\infty$, solid curves phase boundaries, and perforated circles denote norm-SGD simulations; 
    colors that blend well indicate a good match to the theory.
    Increasing $\nu$ removes the width bottleneck, while increasing $\alpha$ suppresses minibatch noise and drives the EoSS--EoS transition; within EoS, the stability threshold depends weakly on $\nu$. \textbf{Center:} $\eta_{\mathrm{eff}}^\infty$ versus $\nu$, showing width-limited behavior and the phase boundary between batch-dependent EoSS branches to the common EoS branch as $\alpha$ increases. \textbf{Right:} $\eta_{\mathrm{eff}}^t$ at $\nu=1$. Dots are simulations, dashed curves are DMFT, and horizontal dotted lines are stationary predictions.
    }
    \label{fig:EoS}
\end{figure}

Let $R_3^{ts}$ denote the causal DMFT response associated with the gradient field whose equal-time correlation $C_3^{tt}$ determines the squared gradient norm. In a stationary state, $R_3^{ts}=R_3(t-s)$. Its Z-transform (see App~\ref{app:z-transform} for an introduction to the Z-transform and its use here) obeys
\begin{equation}
\hat{R}_{3}\left(z\right)=\left(1+\frac{1}{\nu}\left\langle \frac{\lambda_{k}\neffi}{z-1+\lambda_{k}\neffi\hat{R}_{3}\left(z\right)}\right\rangle _{k}\right)^{-1},
\qquad
\hat R_3(z)\to1\quad\text{as }z\to\infty 
\label{eq:R3_theta}
\end{equation}
where $\langle\cdot\rangle_k=d^{-1}\sum_k(\cdot)$ and the large-$z$ condition selects the causal branch. Equation~\ref{eq:R3_theta} determines the response for a candidate value of $\neffi$. We determine $\neffi$ by imposing the gradient norm square $C_3^\infty$ and hence $\neffi$ has a finite size as $t\to\infty$ and solve for it self-consistently. The resulting self-consistency condition admits three distinct stationary solutions, or phases, as derived in Appendix~\ref{app:pole_hunt} and illustrated in Fig.~\ref{fig:EoS}. (1) In the under-parametrized regime $\nu<1$, the random-features miss part of the teacher, producing irreducible error sustained by bias and noise. In the over-parameterized regime $\nu>1$ or large batch $\alpha$, the effective learning rate is instead stability-limited: (2) at small batch, collective minibatch fluctuations produce an edge of stochastic stability (EoSS), whereas at sufficiently large batch (3) the leading mode of the Hessian $\bm H=\bm A\bm\Lambda\bm A^\top/N$ becomes marginally stable, producing a deterministic edge of stability (EoS). The three phases are determined by
\begin{align}
    \underbrace{\nu=\left\langle \frac{\lambda_{k} \neffi \rho}{1+\lambda_{k} \neffi \rho}\right\rangle _{k} ~\And~ \nu<1}_\text{Phase 1: Width approximation error bias},
    &&
    \underbrace{\alpha = \left\langle \frac{\lambda_{k} \neffi \hat{R}_3(-1)}{2-\lambda_{k} \neffi \hat{R}_3(-1)}\right\rangle _{k}}_\text{Phase 2: EoSS},
    &&
    \underbrace{\nu = \left\langle \left( \frac{ \lambda_{k} \neffi \hat{R}_3(-1)}{2-\lambda_{k} \neffi \hat{R}_3(-1)} \right)^2 \right\rangle_k}_\text{Phase 3: EoS}.
\label{eq:stability_phases}
\end{align}
Each of the conditions in Eq.~\ref{eq:stability_phases} together with Eq.~\ref{eq:R3_theta}, and additionally Eq.~\ref{eq:bias_eta} for Phase 1, determines $\neffi$, $\hR$, and $\rho$ for Phase 1.

These conditions have direct spectral interpretation through random matrix theory. As we show in Appendix~\ref{app:rmt_interp}, $\hat R_3$ is related to the Stieltjes transform of the random-feature Hessian $\bm H$, and the EoS phase condition can be shown to be equivalent to $\neffi=\frac{2}{\lambda_{\text{max}}(\nu)}$,
where $\lambda_{\text{max}}(\nu)$ is the deterministic equivalent of the upper edge of the spectrum of $\bm H$. Similarly, the EoSS condition can be compared with the usual mini-batch mean-square stability criterion for a given realization of the Hessian: $B = \tr \left[ \tfrac{\eta \bm H}{2-\eta\bm H}\right]$~\citep{mulayoffExactMeanSquare2024}. Remarkably, the crossing from EoSS to EoS happens at finite critical batch size $\alpha_c$, rather than $\alpha\to\infty$, reflecting the fact that there is only a marginal $1/d$ benefit in stability by increasing $\alpha$ beyond $\alpha_c$. $\alpha_c$ itself can be found analytically by comparing the conditions for phases 2 and 3 above at fixed $\nu$ and solving for $\alpha$\footnote{This $\alpha_c$ is indeed critical in the statistical physics sense and marks a phase transition; but it is distinct from ``critical batch size'' introduced by~\citep{mccandlish2018empirical}.}.

\begin{figure}[t!]
    \centering
    \includegraphics[width=0.31\linewidth]{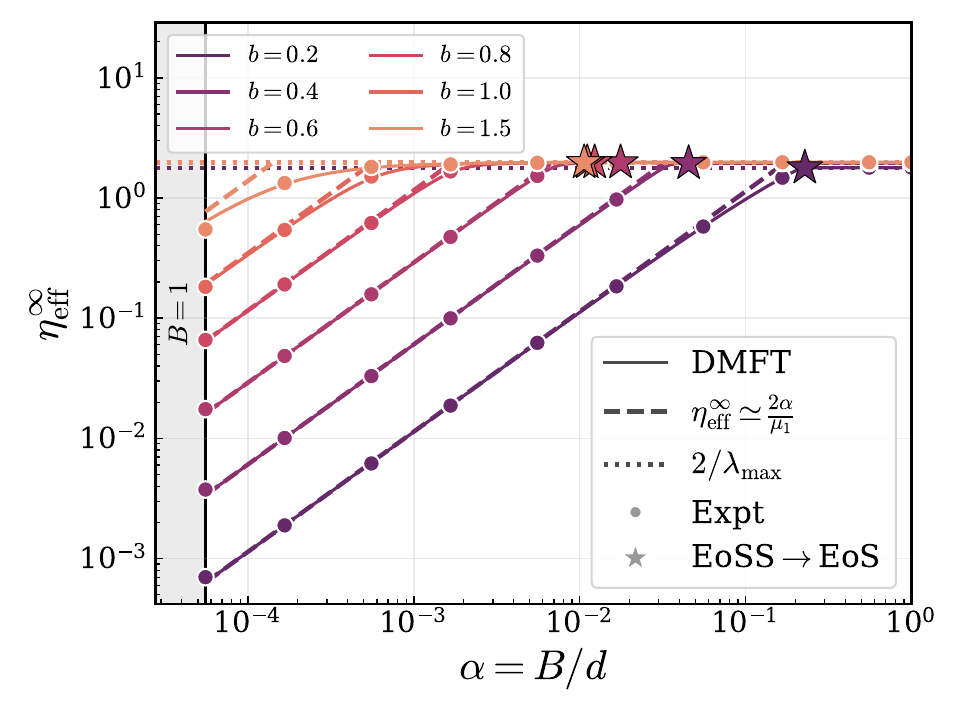}
  \includegraphics[width=0.31\linewidth]{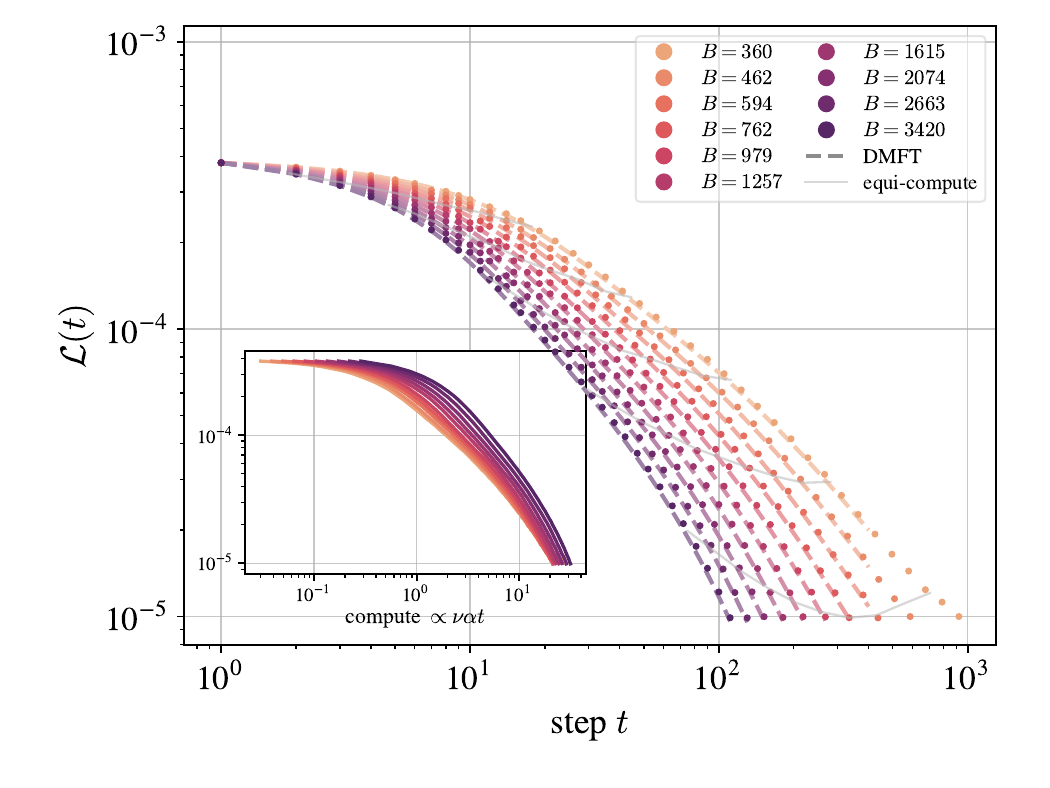}
  \includegraphics[width=0.31\linewidth]{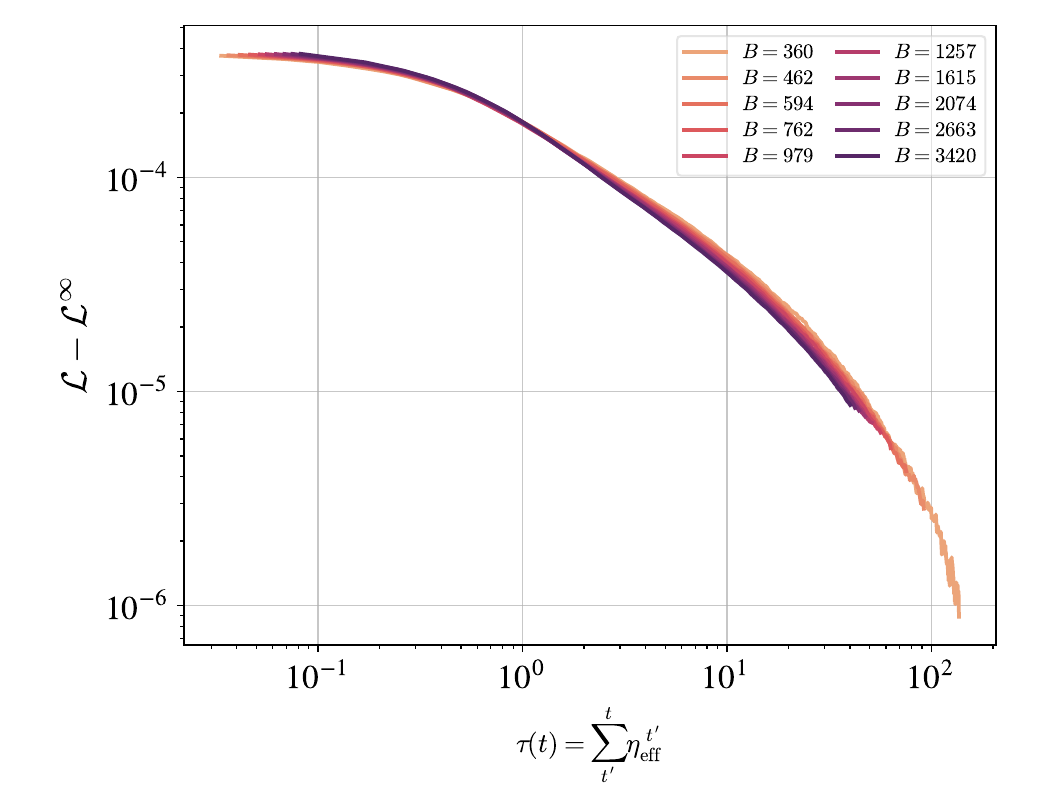}
    \caption{\textbf{Spectral control of batch-size scaling and saturation.}
\textbf{Left:} Stationary effective learning rate $\eta_{\mathrm{eff}}^\infty$ versus batch ratio $\alpha=B/d$, at fixed width ratio $\nu=N/d$. Solid curves show DMFT predictions and circles show simulations. Dashed lines indicate the small-batch linear approximation $2\alpha/\mu_1$, with $\mu_1=\langle\lambda_k\rangle_k$, and dotted lines indicate the curvature ceiling $2/\lambda_{\max}(\nu)$. Stars mark the EoSS--EoS transition; the vertical line marks $B=1$. Flatter spectra support a more extended increase in effective learning rate before saturation. 
\textbf{Center:} Training different batch sizes $\alpha$ on $a=0.45$, $b=0.2$ data. With equal compute, a larger batch size achieves the same level of performance faster (fewer steps). The inset shows the loss as a function of compute with a nearly collapsed curve across all batch sizes within a certain range.
\textbf{Right:} The excess loss above the loss floor in \eqref{eq:EoSS_loss_floor}. Training curves for different batches \emph{collapse} to one master curve when measures in time $\tau(t) = \sum_{t'}^t \neff^{t'}$, implying that $\neff^{t'},\, \alpha$ together control compute vs. time tradeoff. 
    }
    \label{fig:eta_vs_eoss}
\end{figure}

\section{Batch Size and Width Speedups}
While compute efficiency in power-law random-feature models has been studied extensively (for the case of SGD see~\cite{bordelonDynamicalModelNeural2024,paquette4+3PhasesComputeOptimal2024}), another aspect is how to reduce the number of optimization steps needed to reach a target loss without substantially increasing compute, which we will refer to as \emph{speed} or speedup. The three phases identified above yield distinct regimes: \vspace{-0.3cm}
\begin{itemize}
    \item In Phase 1, increasing batch size speeds up optimization, but the reduction in steps does not offset the higher compute per step. Increasing width, however, can reduce the required steps enough to reach some target losses faster at roughly the equal total compute.
    \item In Phase 2, increasing batch size can result in speedup with roughly equal compute, but increasing width cannot.
    \item In Phase 3, neither larger batch size nor wider networks can lead to speedup with approximately equal compute if the model is over-parameterized ($\nu\geq1$) or the total parameter count in the model is larger than the batch size ($\nu \geq \alpha$). 
\end{itemize}
We quantify these results using $Q_\alpha (\alpha) := \neffi/\alpha$ (or $Q_\nu (\nu) := \neffi/\nu$) which measure whether the potential steps saving due to the increase $\neffi$ can compensate the increase in compute due to $\alpha$ ($\nu$ resp.). In all phases above these quantities would depend on the data and task through $a,b$, enabling speed up in some scenarios an forbidding it in others.

The absence of approximately equal-compute speedup follows from the scaling of the effective learning rate. In the small-step Phase-1 regime, increasing batch size yields only square-root gains; in Phase 2, increasing width yields gains slower than square-root. These gains cannot offset the linear increase in per-step compute. In Phase 3, increased batch size provides no effective-step gain, while the fractional gain from increasing $\nu$ scales at most as $\sqrt{\nu}$ leading to a significant compute inefficiency. The derivations are given in Appendix~\ref{subapp:no_speedup}.

\subsection{Batch Speedup at the Edge of Stochastic Stability}
We study this trade-off on the stationary EoSS branch by varying $\alpha=B/d$ at fixed $\nu=N/d$ and input spectrum. Assuming a per-step cost proportional to $\alpha$, we measure stationary effective time per unit compute by $Q_\alpha(\alpha)=\eta_{\mathrm{eff}}^\infty(\alpha)/\alpha$. Linear growth of the effective learning rate would reduce the required number of updates proportionally, leaving the total compute unchanged. Throughout EoSS, however, $Q_\alpha'(\alpha)<0$ for all $b$: larger batches yield only sublinear effective-step gains. Thus, accumulating the same effective time requires fewer updates but more compute; nearly linear scaling keeps this overhead small.

Nevertheless, the departure from linear scaling can be arbitrarily small. Expanding the Phase 2 condition from Eq.~\ref{eq:stability_phases} at small $\alpha$ gives
\begin{equation}
    \eta_{\mathrm{eff}}^\infty
=\frac{2\alpha}{\mu_1}
\left[1-A\alpha+\mathcal O(\alpha^2)\right],
\qquad
A:=\frac{\mu_2}{\mu_1^2}+\frac1\nu
\label{eq:neff_B_star}
\end{equation}
where $\mu_j=\langle\lambda_k^j\rangle_k$. Both the relative efficiency penalty and the logarithmic slope deficit $1-\partial\log\eta_{\mathrm{eff}}^\infty/\partial\log\alpha$ therefore scale like $A\alpha$. Allowing a small compute overhead $\delta$ gives the local estimate $\alpha_\delta\simeq\delta/A$, thus defining a notion of approximately identical compute use, with $\delta=0$ giving linear scaling -- exactly the same compute and loss with less optimization steps.

When progress toward a target loss follows a common effective-time description across batch sizes (as the excess loss in Fig~\ref{fig:eta_vs_eoss}, right panel), $\alpha_\delta$ characterizes the departure from equal compute for target loss. It thus provides a local analogue of empirical critical batch sizes~\citep{zhang2025does}. Unlike $\alpha_\delta$, which quantifies a tolerated departure from linear scaling, the EoSS--EoS boundary $\alpha_c$ marks saturation of the stationary effective learning rate~\citep{meterez2026defense}.

The spectrum controls how quickly this benefit deteriorates. For $\lambda_k=k^{-b}$, larger $b$ concentrates curvature in fewer leading directions, increasing $\mu_2/\mu_1^2$ and the initial efficiency penalty. Flatter spectra spread curvature across more directions and support a more sustained benefit from increased batch size. Figure~\ref{fig:eta_vs_eoss} (left) illustrates this distinction: steeper spectra allow larger effective steps at small $\alpha$, but saturate sooner. Dimension determines whether this regime accommodates large physical batches. The leading linear approximation reveals the dependence on spectrum and dimension
\begin{equation}
    \eta_{\mathrm{lin}}^\infty
=\frac{2\alpha}{\mu_1}
\sim
\begin{cases}
2(1-b)\alpha d^b, & 0\le b<1\\
2\alpha d/\zeta(b), & b\geq1
\end{cases}
\end{equation}
where $\zeta(b)=\sum_{k=1}^{\infty}k^{-b}, \, b>1$. At fixed physical batch size $B$, increasing $d$ reduces the leading effective-step gain for $b\leq1$, because the total curvature $\sum_k\lambda_k$ grows with dimension. For $b>1$, this sum converges to $\zeta(b)$, making the linear prediction $\eta_{\mathrm{lin}}^\infty$ independent of $d$.

\begin{figure}[t]
    \centering
    \includegraphics[width=0.64\linewidth]{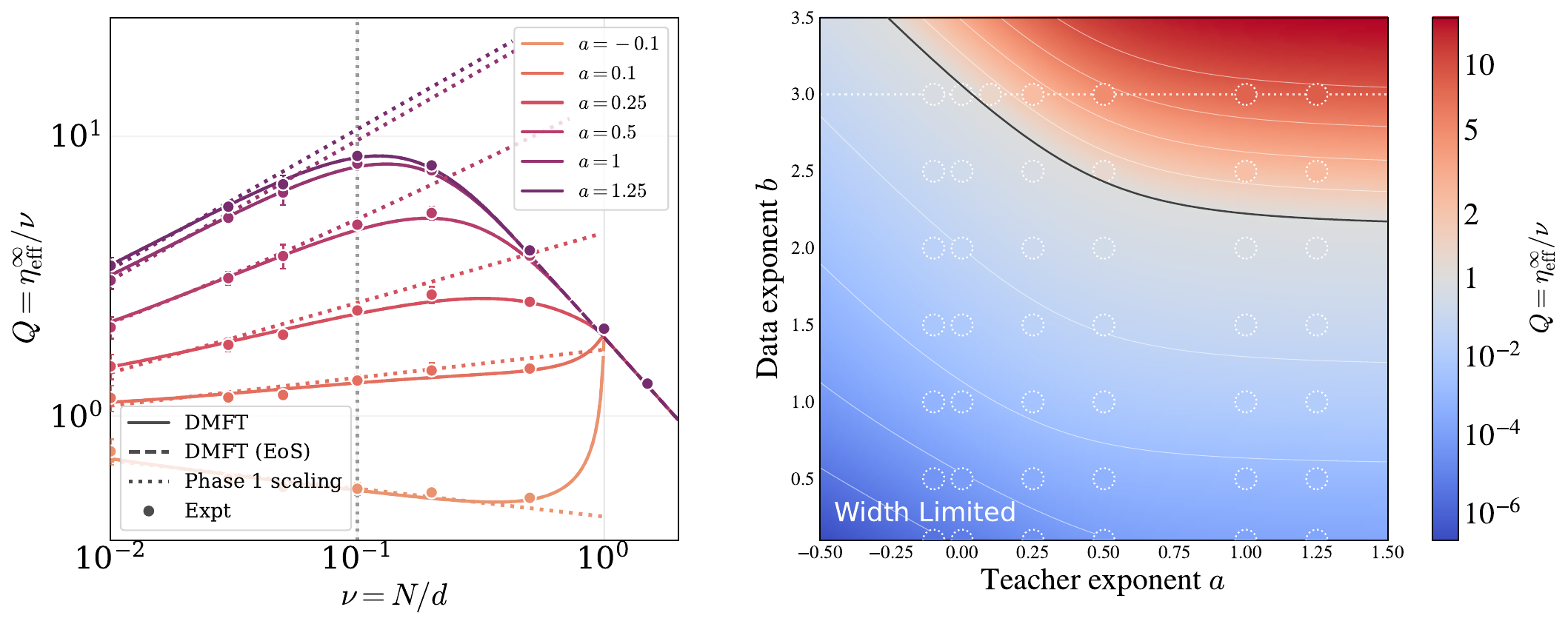}
    \includegraphics[width=0.33\linewidth]{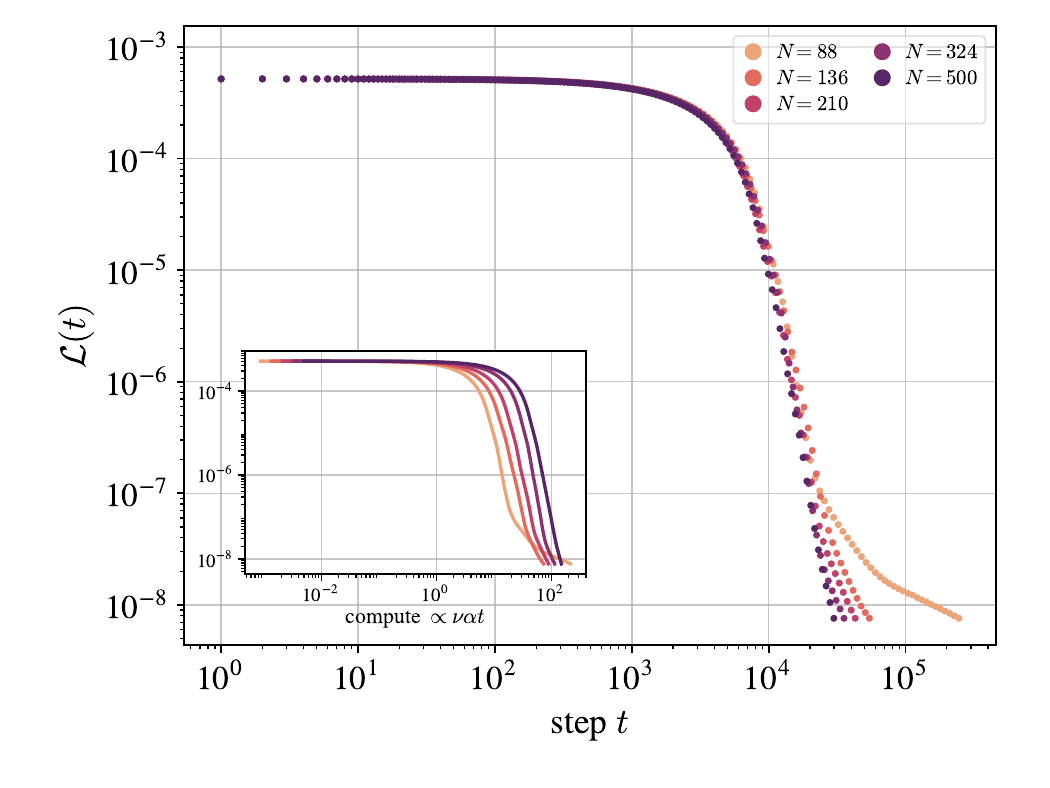}
    \caption{\textbf{Spectral dependence of width efficiency.}
Stationary effective time per unit compute, quantified by $Q_\nu=\eta_{\mathrm{eff}}^\infty/\nu$.
\textbf{Left:} Width dependence of $Q_\nu$ at fixed $b=3$, for different teacher exponents $a$. Solid curves show DMFT predictions before EoS, dashed curves indicate the EoS branch, and markers show simulations. Increasing $Q_\nu$ corresponds to superlinear growth of the effective learning rate with width; a maximum identifies an optimal width according to this stationary efficiency measure.
\textbf{Center:} Dependence of $Q_\nu$ on the teacher and data exponents at fixed $\nu=0.1$. The background shows DMFT predictions, while circles show simulation estimates using the same logarithmic color scale. Dotted reference lines indicate the fixed width and data exponent linking the two panels. Both panels use $d=2000$, $B=200$, and fixed bare learning rate $\eta=10^{-5}$. \textbf{Right:} Loss as a function of steps (inset: compute) showing an example where wider networks train faster, while requiring approximately the same compute as narrower networks.}
    \label{fig:width_dep}
\end{figure}

Eventually, the effective learning rate reaches the curvature ceiling at the EoSS–EoS transition, marked by stars in Figure~\ref{fig:eta_vs_eoss} (left). Beyond this point, further batching provides no effective-step gain and the efficiency deteriorates as $Q_\alpha\propto\alpha^{-1}$.

The right panel of Fig.~\ref{fig:eta_vs_eoss} shows that the loss curves for different batch sizes approximately collapse when the excess loss $\mathcal L-\mathcal L^\infty$ is plotted against cumulative effective time, $\tau(t):=\sum_{t'}^{t}\neff^{t'}$. Here, $\mathcal L^\infty$ is the corresponding loss floor from \eqref{eq:EoSS_loss_floor}.
This collapse indicates that reaching a fixed target excess loss requires approximately the same effective time $\tau_*$ across batch sizes.
The middle panel of Fig.~\ref{fig:eta_vs_eoss} demonstrates the compute-speed tradeoff. The main plot shows the loss as a function of step size, showing that larger batches get to the same target loss faster, while the contour connects equal-compute points. The inset shows the same loss curves as a function of compute ($\propto BN t$) with very little spread in compute around $B=1,000$ indicating significant savings in number of steps.

\subsection{Width Speedup at the Width Limited Phase}

Next, we vary the width ratio $\nu=N/d$ at fixed batch ratio $\alpha$, base learning rate $\eta$, input and teacher spectra $b,a$ (resp.). Similarly, assuming a per-step cost proportional to $\nu$, we measure stationary width efficiency by $Q_\nu(\nu)=\neffi(\nu)/\nu$. Whereas batch efficiency decreases throughout EoSS, width efficiency can increase on the width-limited branch: adding parameters can raise the stationary effective learning rate more than proportionally to their computational cost. This behavior is controlled by the residual target power $P_\nu:=\tfrac1d\sum_{k=1}^d
\tfrac{\lambda_k\beta_k^2}{1+\lambda_k\neffi\rho}$ left unrepresented at finite width, where the product $\neffi\rho$ is determined by the width constraint in Eq.~\ref{eq:stability_phases} and it is independent of $\eta$. As derived in Appendix~\ref{appendix:width_speedup}, taking $\eta \ll 1$ at fixed $0<\nu<1$ gives \vspace{-0.2cm}
\begin{equation}
    \eta_{\mathrm{eff}}^\infty(\nu)\sim \eta\sqrt{\frac{\alpha}{\mu_1P_\nu}}.
\end{equation}
Since $\mu_1$ is independent of width, reducing $P_\nu$ increases the stationary effective learning rate as $P_\nu^{-1/2}$. Efficiency improves only if this increase outweighs the additional cost per update, which requires $Q'_\nu \geq 0$. The data spectrum and teacher alignment jointly determine whether this condition holds. For $b>1$ and $2a+b>1$, the width scalings in Appendix~\ref{appendix:width_speedup} yield
\begin{equation}\label{eq::Q_nu}
    Q_\nu(\nu)\sim
\begin{cases}
\nu^{(2a+b-3)/2}, & (1-b)/2<a<1/2\\
\nu^{(b-2)/2}, & a\geq1/2
\end{cases}
\end{equation}
up to log factors of $\nu$. For flatter spectra, $0<b\leq1$, the corresponding asymptotic scalings instead predict decreasing efficiency within this width window for all admissible teacher exponents $a$. 

Figure~\ref{fig:width_dep} (left) shows stronger width-efficiency gains for larger $a$, which concentrates target power in the leading data modes. The curves reach an interior maximum at $\nu_*<1$, while beyond this maximum, further widening no longer produces a proportional increase in the effective learning rate. Here, the decreasing efficiency on the EoS branch, where $Q_\nu'(\nu)<0$ (see Appendix~\ref{subapp:no_speedup}), limits the width benefit. The middle panel shows the dependence of $Q$ on both $a,b$ when fixing $\nu=0.1$, visualizing that higher $a,b$ favor larger widths, while low $a,b$ favor narrow widths while they are not capacity limited. Finally, the left panel of Fig.~\ref{fig:width_dep} shows a case where larger widths lead to faster training at approximately the same compute budget for the same target loss. 
In this regime, the loss floor $L^\infty$ plays an important role, and narrow models can reach their capacity limit before the width speedup can be realized. In these cases, wider models trivially reach a target loss faster and using less compute, implying that some of the benefit is due to the lower loss floor.  
In Fig.~\ref{fig:width_speed_all} in Appendix~\ref{appendix:width_speedup}, we show narrower widths that exhibit a capacity floor under the same parameters as in the right panel.

\begin{figure}[t]
    \centering
    \includegraphics[width=0.4\linewidth]{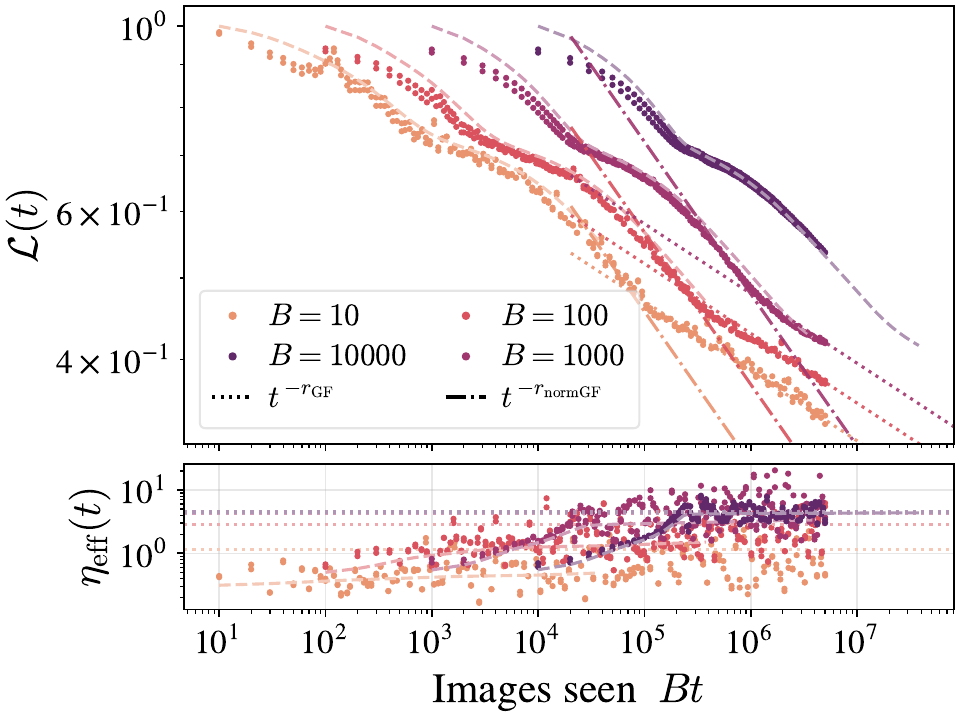}
    \includegraphics[width=0.4\linewidth]{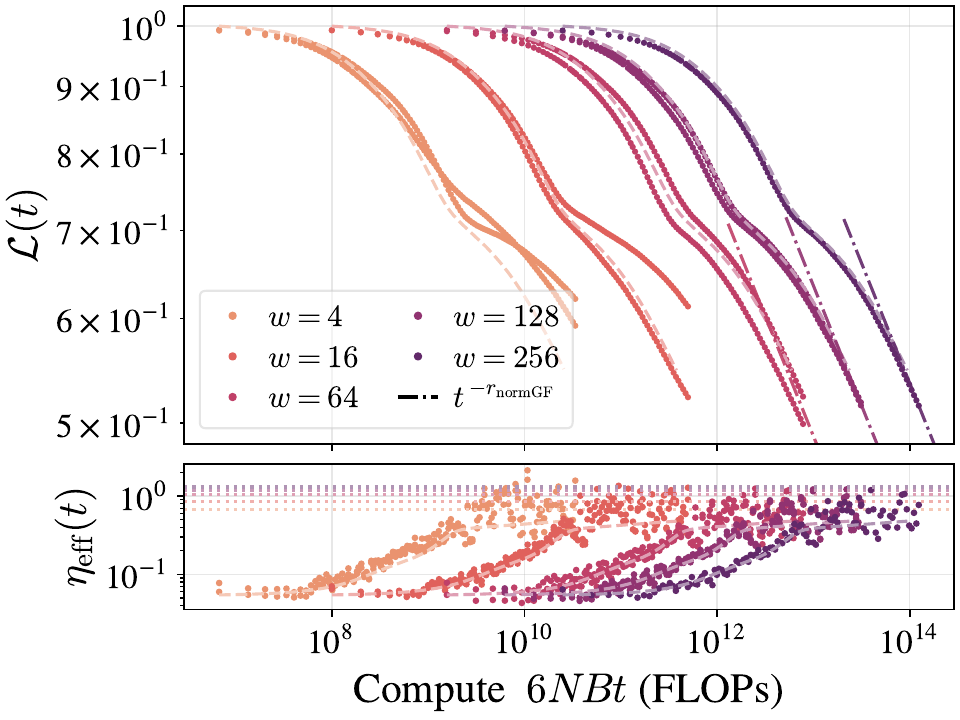}
    \caption{\textbf{Theory predicts training curves exponents for training with linearized networks.} Loss (\textbf{top}) and effective learning rate (\textbf{bottom}). Before $\neff$ hits its ceiling $\neffi$, the loss decreases with an accelerated power law, after the ceiling is reached the power law degrades to the standard SGD power law. Throughout the narrowest width and smallest batch size are optimal, as predicted in the previous section for these $a,b$. (\textbf{left}) Loss as a function of images seen, $Bt$, for different batch sizes and fixed $w=128$, $\eta=0.1$ (proportional to compute). (\textbf{right}) Loss as a function of compute, calculated as $6NBt$, for different widths and fixed $B=1,000$, $\eta=0.01$}
    \label{fig:resnet_cifar}
\end{figure} 

\vspace{-4pt}
\section{Experiment with real data and networks}
\vspace{-4pt}
In this section, we experiment with training real ResNet in the lazy (kernel) regime on the CIFAR-5M binary classification task of classifying animate vs inanimate while training with normalized SGD. To match the setting to our theory, we extract $a,b$ from the infinite-width deterministic NTK and treat the linearized neural networks eNTK as random features drawn from the NTK. In our experiments, we measured $b\simeq1.8$ and $a\simeq-0.325$, close to the measurements of~\citep{bordelonDynamicalModelNeural2024} for the same architecture.

In Fig.~\ref{fig:resnet_cifar}, right panel, we show the results when varying the width in the range $w=4$ ($N=1,120$ random features) to $w=256$ ($N=4,135,936$) with fixed $B=1,000$ for a single pass with no repetitions. On the left panel we show the results for the same setting when fixing $w=128$ and varying the batch size from $B=10$ to $B=10,000$. Since the NTK is infinite-dimensional, while any network has a finite parameter count and batch size, we adopt our theory to this dimension-free regime ($d\to\infty$ while $N,B$ are finite) in Appendix~\ref{app:d-free}. This also means that $\nu<1$ and phase 2 (EoSS) is inaccessible. As discussed in the previous section, this implies that the smallest batch size $B$ is always compute optimal, which we confirm in the left panel. Additionally, based on results from the previous section, we expect that as long as the model is time-constrained rather than parameter-constrained, narrow networks will be more compute-efficient for our measured exponents $a,b$, which the experiments in the right panel confirm.

Additionally, we add to the figures rulers showing the normalized gradient flow accelerated exponent, and intersecting rulers showing the breakdown of acceleration and the change to the standard SGD exponent when the effective learning rate reaches marginal stability. Throughout, our deterministic DMFT dynamics track the stochastic dynamics well, even though the DMFT assumes Gaussian features and only uses the NTK spectrum, the NTK target coefficients, and $N,B$, with no information on the eNTK or data, architecture, and task beyond those four quantities.

\vspace{-5pt}
\section{Outlook}
\vspace{-5pt}
Normalized SGD provides a solvable setting in which acceleration, stability, and resource allocation can be studied together. In continuous time, normalization preserves the gradient-flow trajectory while accelerating its clock, yielding accelerated power-law, exponential, or formal finite-time convergence depending on the teacher exponent. However, in discrete dynamics, the growing effective learning rate ultimately breaks this acceleration down through one of three mechanisms: width-limited noise and approximation error, a noise-driven EoSS, or deterministic EoS dynamics.

We relate these phases to compute efficiency and speedup. We find a critical batch ratio $\alpha_c$ that marks the phase transition between EoSS and EoS, and distinguish it from $\alpha_\delta$, the maximal batch ratio that allows approximately linear scaling of learning rate and batch size -- which allows speedup with equal compute.  We quantify the dependence of these scales on width and on data exponents. 

Experiments with linearized ResNets on CIFAR-5M support the predicted acceleration and its breakdown, and connect the spectral description to observed batch- and width-dependent compute trends. Extending this connection beyond the kernel regime requires accounting for the evolution of features, target alignment, and curvature during training. Further directions include determining how learning-rate and batch-size schedules alter the trade-off, and relating effective-time efficiency to wall-clock performance under realistic hardware costs. Additionally, developing similar trade-off of compute--parameter count instead of compute-speedup could accelerate inference at the cost of increased compute cost in the training phase. Finally, extending the analysis to optimizers that also modify update direction would clarify which conclusions persist beyond global gradient normalization.
\\
\paragraph*{Acknowledgments}
The authors thank Alex Meterez for the useful discussion. I.L. and C.L. are supported by DARPA grant AIQ-HR00112520041. C.P. is supported by
an NSF CAREER Award (IIS-2239780), DARPA grants DIAL-FP-038 and AIQ-HR00112520041, and
the Simons Collaboration on the Physics of Learning and Neural Computation. This work has been made possible in part by a gift from the
Chan Zuckerberg Initiative Foundation to establish the Kempner Institute for the Study of Natural and
Artificial Intelligence. 

\clearpage

\bibliography{main}

\clearpage
\appendix
\counterwithin{equation}{section}
\section{A Short Introduction to the Z-transform}
\label{app:z-transform}

The Z-transform packages the sequence $f^0,f^1,f^2,\ldots, f^t, \ldots$ into a function of
a complex variable, with each power of $z^{-1}$ labeling one time step; the result is a generating function for a sequence, which can also be seen as the discrete-time analogue of the Laplace transform under the substitution $z=e^{s}$.
This transformation diagonalizes operators that have a shift symmetry, or as they are more commonly called time translation invariant (TTI) operators. Practically, the transform turns shifts and convolutions into algebraic manipulations.

\subsection{From a sequence to a function}

For a causal sequence $f^t$, defined for $t\geq0$, the Z-transform is
\begin{equation}
    \hat f(z):=\sum_{t=0}^{\infty}f^t z^{-t}
    =f^0+f^1z^{-1}+f^2z^{-2}+\cdots,
    \label{eq:z_transform_definition}
\end{equation}
where first equality is the definition and the second makes its bookkeeping role explicit. For example, the transform of the sequence $1,a,a^2,\ldots$ is the
geometric series
\begin{equation}
    \sum_{t=0}^{\infty}a^t z^{-t}
    =\frac{z}{z-a},
    \qquad |z|>|a|.
    \label{eq:z_exponential_example}
\end{equation}
The condition $|z|>|a|$ is the \emph{region of convergence} of this series.
For a causal sequence, the region of convergence is the exterior of a circle
whose radius is set by the growth rate of the sequence. Keeping track of this
region tells us where algebraic manipulations and contour integrals are valid.

In particular, $z\to1$ corresponds to the small-$s$ limit $s\to0$ in Laplace transform or the small frequency limit in Fourier transform $\omega\to0$, which is why the behavior near $z=1$ contains information about long times. 

More generally, the sequence can be recovered from its transform by extracting the
coefficient of $z^{-t}$
\begin{equation}
    f^t=\oint_{\gamma}\frac{d z}{2\pi i z}\,
    z^{t}\hat f(z),
    \label{eq:z_transform_inverse}
\end{equation}
where the contour $\gamma$ is a circle lying in the region of convergence.
This is the familiar coefficient-extraction formula for a power series. 
In practice, one rarely use this method to invert the series directly; it is more useful as a way to understand how the singularities of $\hat f$ determine the behavior of $f^t$.

\subsection{Shifts and recurrences}

A change of summation index gives the transform of a forward shift,
\begin{equation}
    \sum_{t=0}^{\infty} f^{t+1}z^{-t}
    =z\bigl(\hat f(z)-f^0\bigr).
    \label{eq:z_forward_shift}
\end{equation}
It follows that the forward difference $(Df)^t=f^{t+1}-f^t$ satisfies
\begin{equation}
    \widehat{Df}(z)=(z-1)\hat f(z)-zf^0.
    \label{eq:z_transform_difference}
\end{equation}
Thus $z-1$ plays the role of the perhaps more familiar Laplace variable $s$ for a time derivative,
while the remaining term carries the initial condition.

This shift rule turns linear recurrences into algebraic equations. For example,
if
\begin{equation}
    x^{t+1}=a x^t+b^t,
\end{equation}
with prescribed input $b^t$, transforming both sides and using
\eqref{eq:z_forward_shift} gives
\begin{equation}
    \hat x(z)=\frac{zx^0+\hat b(z)}{z-a}.
    \label{eq:z_recurrence_example}
\end{equation}
The denominator already reveals the natural mode $a^t$.

\subsection{Time-translation invariance and diagonalization}

Consider a causal matrix $K^{ts}$ acting on time indices. It is
time-translation invariant, or TTI, when it depends only on the lag between
its indices,
\begin{equation}
    K^{ts}=K(t-s),
    \qquad K(u)=0\quad\text{for }u<0.
\end{equation}
Its action on a sequence is therefore a convolution,
\begin{equation}
    (Kf)^t:=\sum_{s=0}^{t}K(t-s)f^s,
    \qquad
    \hat K(z):=\sum_{u=0}^{\infty}K(u)z^{-u}.
\end{equation}
Ignoring the initial boundary for a moment, a geometric sequence $f^s=z^s$
is an eigenvector of every TTI matrix:
\begin{equation}
    \sum_s K(t-s)z^s
    =z^t\sum_u K(u)z^{-u}
    =\hat K(z)z^t.
    \label{eq:z_tti_eigenvector}
\end{equation}
The Z-transform is therefore a change to the common eigenbasis of TTI
matrices, and $\hat K(z)$ is the corresponding eigenvalue. This is the precise
sense in which it diagonalizes the time dynamics. Matrix products become
products of their Z-transforms; equivalently, convolution with a sequence
obeys
\begin{equation}
    \widehat{Kf}(z)=\hat K(z)\hat f(z).
    \label{eq:z_transform_convolution}
\end{equation}

For example, the delayed running-sum matrix has $K(u)=\mathbf 1_{u>0}$, so
\begin{equation}
    \hat K(z)
    =\sum_{u=1}^{\infty}z^{-u}
    =\frac{1}{z-1},
    \qquad |z|>1.
    \label{eq:z_transform_running_sum}
\end{equation}
Together with \eqref{eq:z_transform_difference}, this shows that $z-1$
and $1/(z-1)$ represent discrete differentiation and integration, just as
$s$ and $1/s$ do in the Laplace transform.

\subsection{Reading dynamics from singularities}

Equation~\eqref{eq:z_exponential_example} gives the basic dictionary: a pole
at $z=a$ produces a mode proportional to $a^t$. Such a mode decays when
$|a|<1$ and grows when $|a|>1$. On the unit circle it is marginal: $a=1$
gives a constant, $a=-1$ a period-two oscillation, and $a=e^{i\omega}$ an
undamped oscillation at frequency $\omega$. The unit circle is therefore the
discrete-time stability boundary. More generally, a pole of order $m$ adds a factor $t^{m-1}$ to the exponential mode.

If a sequence approaches a finite limit as $t \to \infty$, its asymptotic value can be obtained using the final-value theorem
\begin{equation}
    \lim_{t\to\infty}f^t
    =\lim_{z\to1}(z-1)\hat f(z).
    \label{eq:z_final_value}
\end{equation}
Thus a nonzero constant at late times appears as a simple pole at $z=1$, with
coefficient isolated by multiplying by $z-1$. A growing mode or an undamped
oscillation violates the pole condition and prevents the limit from existing.

The same picture can be phrased in frequency language. Evaluating a stable
transform at $z=e^{\epsilon+i\omega}$ and taking $\epsilon\downarrow0$ gives
its response to a discrete-time frequency $\omega$. The point $z=1$ is zero
frequency, while $z=-1$ is the highest discrete-time frequency and corresponds
to sign alternation from one step to the next.

\subsection{Beyond TTI matrices}

An ordinary Z-transform suffices for a TTI matrix because $K^{ts}=K(t-s)$
depends on only one time lag. A general two-time matrix $F^{t_1t_2}$, such as
a correlation matrix, is not necessarily TTI and retains two independent time
indices. We therefore introduce one transform variable for each index:
\begin{equation}
    \tilde F(z_1,z_2)
    :=\sum_{t_1,t_2=0}^{\infty}
    F^{t_1t_2}z_1^{-t_1}z_2^{-t_2}.
    \label{eq:double_z_transform}
\end{equation}
The double transform preserves the diagonalization of TTI multiplication on
either side. If $K$ and $L$ are TTI matrices, then
\begin{equation}
    \widetilde{KFL^{\top}}(z_1,z_2)
    =\hat K(z_1)\hat L(z_2)\tilde F(z_1,z_2).
    \label{eq:double_z_composition}
\end{equation}

For our use, the equal-time diagonal is especially important. If $F$ is naturally diagonal in time, $F^{t_1t_2}=\delta^{t_1t_2}f^{t_1}$, then
\begin{equation}
    \tilde F(z_1,z_2)=\hat f(z_1z_2).
    \label{eq:diagonal_double_transform}
\end{equation}
because the two time indices are equal. For a general $F$, define
\begin{equation}
    f^t:=F^{tt},
    \qquad
    \hat f(\zeta):=\sum_{t=0}^{\infty}F^{tt}\zeta^{-t}.
\end{equation}
We keep the product $z_1z_2=\zeta$ fixed by writing
\begin{equation}
    z_1=\sqrt{\zeta}\,w,
    \qquad
    z_2=\frac{\sqrt{\zeta}}{w}.
\end{equation}
The double transform thus also have the degree of freedom $w^{t_2-t_1}$. Integrating along the unit circle removes every nonzero power of $w$, leaving only $t_1=t_2$:
\begin{align}
    \hat f(\zeta)
    &=\oint_{|w|=1}\frac{d w}{2\pi i w}\,
      \tilde F\left(\sqrt{\zeta}\,w,
                    \frac{\sqrt{\zeta}}{w}\right)
    \nonumber\\
    &=\int_0^{2\pi}\frac{d\theta}{2\pi}\,
      \tilde F\left(\sqrt{\zeta}e^{i\theta},
                    \sqrt{\zeta}e^{-i\theta}\right).
    \label{eq:equal_time_projection}
\end{align}
Here $\zeta$ follows the behavior averaged across time, while $w=e^{i\theta}$ resolves relative-time difference and we use the angular integration to project onto zero relative time and hence onto the diagonal.

\subsection{Use in the steady-state analysis}

We now specialize these facts to the calculation in this paper. Once
$\eta_{\rm eff}^t$ approaches a constant, the matrices $\Theta$ and $R_3$
become TTI. The Z-transforms diagonalize the time indices, turning
matrix products and inverses into scalar algebra. In particular,
$\hat\Theta(z)=\eta_{\rm eff}/(z-1)$, while the physical branch of the self-consistent equation for $\hat R_3$ is selected by
$\hat R_3(z)\to R_3(0)=1$ as $z\to\infty$.

The correlation functions remain genuine two-time objects. We therefore solve
for their double transforms and apply \eqref{eq:equal_time_projection} to
obtain the equal-time loss and gradient correlations $\hat c_1(\zeta)$ and
$\hat c_3(\zeta)$, and the final-value theorem gives
\begin{equation}
    c_3(\infty)
    =\lim_{\zeta\to1}(\zeta-1)\hat c_3(\zeta),
    \qquad
    \eta_{\rm eff}^{\infty}
    =\frac{\eta}{\sqrt{\nu c_3(\infty)}}.
    \label{eq:z_effective_lr_final_value}
\end{equation}
The transformed DMFT equations contain several possible sources for this
simple pole at $\zeta=1$. Which one becomes singular distinguishes the
width-limited, edge-of-stability, and stochastic
edge-of-stability phases.

\section{Deriving the Dynamical Mean-Field Theory Equations}
\label{app:dmft}
Starting from the loss
\begin{equation*}
    L=\frac{1}{2B}\sum_{\mu=1}^{B}\left(\beta^{\top}\psi_{\mu}-\frac{1}{\sqrt{N}}w^{\top}A\psi_{\mu}\right)^{2}
\end{equation*}
The gradient is
\begin{align*}
g & =\frac{\partial L}{\partial w_{k}}=-\frac{1}{\sqrt{N}}A\frac{1}{B}\psi^{T}\psi\left(\beta-\frac{1}{\sqrt{N}}A^{T}w\right)
\end{align*}
so the weight updates are
\[
w_{t+1}=w_{t}-\sqrt{d}\eta\left(g^{T}g\right)^{-1/2}g
\]
multiplying by $\frac{1}{\sqrt{N}}A^{T}$ from the left 
\[
\frac{1}{\sqrt{N}}A^{T}w_{t+1}=\frac{1}{\sqrt{N}}A^{T}w_{t}-\sqrt{d}\eta\frac{1}{\sqrt{N}}A^{T}\left(g^{T}g\right)^{-1/2}g
\]
and subtracting our equation form $\beta$
\[
\beta-\frac{1}{\sqrt{N}}A^{T}w_{t+1}=\beta-\frac{1}{\sqrt{N}}A^{T}w_{t}+\sqrt{d}\eta\frac{1}{\sqrt{N}}A^{T}\left(g^{T}g\right)^{-1/2}g.
\]
uUsing the following fields
\begin{align}
\mathbb{R}^{d}\ni h_{0}^{t}&:=\beta-\frac{1}{\sqrt{N}}A^{\top}w^{t}
&
\mathbb{R}^{B}\ni h_{1}^{t}&:=\frac{1}{\sqrt{\alpha d}}\psi_{t}h_{0}^{t}
&
\mathbb{R}^{d} \ni h_{2}^{t}&:=\frac{1}{\sqrt{B}}\psi_{t}^{\top}h_{1}^{t} 
 \nonumber\\
\mathbb{R}^{N}\ni h_{3}^{t}&:=\frac{1}{\sqrt{\nu d}}Ah_{2} 
&
\mathbb{R}^{d} \ni h_{4}^{t}&:=\frac{1}{\sqrt{N}}A^{\top}h_{3}^{t}
&
\label{eq:fields_def}
\end{align}
and the dynamics close the loop by
\[
h_{0}^{t+1}=h_{0}^{t}-\sqrt{d}\eta\left(h_{3}^{t\top}h_{3}^{t}\right)^{-1/2}h_{4}^{t}.
\]

\subsection{Field Averaging Using the Cavity Method}
Below, we use the cavity method~\citep{mezardInformationPhysicsComputation2009a} to find a mean-field description for the fields $\{h_i\}_{i=0}^4$ by adding a small perturbation to the rows and columns of $\psi$ and $A$ and reading off their collective average effect on a fresh coordinate in the same field. Along the way, we introduce average response functions $R_{i\to j}$ that denote the effect of a perturbation in a field $h_i$ on the field $h_j$; these response functions are unknown and have to be solved for self-consistently.
\paragraph{Adding a row to $\psi$}
Throughout, we use $\tilde{\cdot}$ to denote a perturbed quantity and $\bar{\cdot}$ to denote a fresh independent entry in the field. We ignore $1/d$ corrections and use $\to$ to denote the large $d$ limit after employing the law of large numbers (LLN) and the central limit theorem (CLT) when summing independent quantities.
\begin{align*}
h_{1}^{t} & =\frac{1}{\sqrt{\alpha d}}\psi_{t}h_{0}^{t}+\frac{1}{\sqrt{\alpha d}}\psi_{t}\tilde{h}_{0}^{t}\\
\bar{h}_{1}^{t} & =\frac{1}{\sqrt{\alpha d}}\tilde{\psi}_{t}h_{0}^{t}+\frac{1}{\sqrt{\alpha d}}\tilde{\psi}_{t}\tilde{h}_{0}^{t}\\
h_{2}^{t} & =\frac{1}{\sqrt{B}}\psi_{t}^{T}h_{1}^{t}+\frac{1}{\sqrt{B}}\psi_{t}^{T}\delta h_{1}^{t}+\tilde{\psi}_{t}^{T}\bar{h}_{1}^{t}
\end{align*}
Reading the effect of the perturbation on the cavity 
\begin{align*}
\bar{h}_{1}^{t}&=\frac{1}{\sqrt{\alpha d}}\tilde{\psi}_{t}^{T}h_{0}^{t}+\frac{1}{\sqrt{\alpha d}}\frac{1}{\sqrt{B}}\intop\tilde{\psi}_{t}^{T}\frac{\partial h_{0}^{t}}{\partial h_{2}^{t'}}\tilde{\psi}_{t'}\bar{h}_{1}^{t'}\,dt'.
\end{align*}
Using the fact that the new row $\tilde{\psi}$ is independent of the old fields, we can use CLT on the first expression and LLN on the second, keeping in mind that batches are independent 
\begin{align*}
\bar{h}_{1}^{t} &\to u_{1}+\frac{1}{\alpha}\intop^{t}\delta\left(t-t'\right)R_{2\to0}\left(t,t'\right)\bar{h}_{1}^{t'}\left(t'\right)\,dt'
=
u_{1}\left(t\right)+\frac{1}{\alpha}R_{2\to0}\left(t,t\right)\bar{h}_{1}\left(t\right)
\end{align*}
with $u_1 \sim \mathcal{N}\left(0,\delta^{tt'}\frac{1}{\alpha}C_{0}^{tt'}\right)$
\[
\left\langle u_{1}^{t\top}u_{1}^{t'}\right\rangle =\frac{1}{\alpha d}\left\langle \left(\psi_{t}^{T}h_{0}^{t}\right)^{\top}\psi_{t}^{T}h_{0}^{t}\right\rangle =\tr\left[\left\langle \psi_{t}\psi_{t}^{\top}\right\rangle \left\langle h_{0k}^{t}h_{0k}^{t\top}\right\rangle \right]=\frac{1}{\alpha d}\sum_{k=1}^{d}\delta^{tt'}\lambda_{k}\left\langle h_{0k}^{t}h_{0k}^{t'}\right\rangle =:\delta^{tt'}\frac{1}{\alpha}C_{0}^{tt'},
\]
and the response
\[
R_{2\to0}=\frac{1}{d}\sum_{k}\lambda_{k}\left\langle \frac{\partial h_{0k}^{t}}{\partial h_{2k}^{t'}}\right\rangle =\frac{1}{d}\sum_{k}\lambda_{k}R_{2\to0k}.
\]
However, since the causal chain runs $h_{0}^{t}\to h_{1}^{t}\to h_{2}^{t}\to h_{3}^{t}\to h_{4}^{t}\to h_{0}^{t+1}$ we see that $h_{2}^{t}$ only effects $h_{0}^{t+1}$ which forces $R_{2\to0}^{tt}=0$

\paragraph{Adding a column to $\psi$}
Similarly, adding a column to $\psi$ we find the following perturbed fields
\begin{align*}
h_{1}^{t} & =\frac{1}{\sqrt{\alpha d}}\psi_{t}h_{0}^{t}+\frac{1}{\sqrt{\alpha d}}\psi_{t}\tilde{h}_{0}^{t}+\tilde{\psi}_{t}\bar{h}_{0}^{t}\\
h_{2}^{t} & =\frac{1}{\sqrt{B}}\psi_{t}^{T}h_{1}^{t}+\frac{1}{\sqrt{B}}\psi_{t}^{T}\delta h_{1}^{t}\\
\bar{h}_{2}^{t} & =\frac{1}{\sqrt{B}}\tilde{\psi}_{t}^{T}h_{1}^{t}+\frac{1}{\sqrt{B}}\tilde{\psi}_{t}^{T}\delta h_{1}^{t}.
\end{align*}
Again, reading off the effect of the perturbation on the cavity, we find
\[
\bar{h}_{2k}^{t}=\frac{1}{\sqrt{B}}\psi_{tk}^{T}h_{1}^{t}+\frac{1}{\sqrt{B}}\frac{1}{\sqrt{\alpha d}}\intop\tilde{\psi}_{tk}^{T}\frac{\partial h_{1}^{t}}{\partial h_{1}^{t'}}\tilde{\psi}_{t'k}\bar{h}_{0}^{t'}\,dt'
\]
and in the high-dimensional limit
\[
\bar{h}_{2k}^{t} \to
u_{2k}+\lambda_{k}\intop^{t}\delta\left(t-t'\right)R_{1}\left(t,t'\right)\bar{h}_{0}\left(t'\right)\,dt'=u_{2}\left(t\right)+\lambda_{k}R_{1k}\left(t,t\right)\bar{h}_{0}\left(t\right)
\]
with
\[
\left\langle u_{2}^{t}u_{2}^{t'}\right\rangle =\delta^{tt'}\lambda_{k}\left\langle h_{1k}^{t}h_{1k}^{t'}\right\rangle =:\delta^{tt'}\lambda_{k}C_{1}^{tt'}
\]
\[
R_{1k}=\lambda_{k}\left\langle \frac{\partial h_{1k}^{t}}{\partial h_{1k}^{t'}}\right\rangle 
\]
We can again use causality to simplify the response $R_{1}^{tt}$. As we have seen above $h_{1}$ is just
noise uncorrelated in time, so the only contribution is the immediate
one, thus $R_{1}=1$

\paragraph{Cavity results}

Repeating the same process for a column and a row of $A$ and collecting all the results together, we find
\begin{align*}
h_{0}^{t+1} & =h_{0}^{t}-\eta\left(C_{3}^{tt}\right)^{-1/2}h_{4}^{t}\\
h_{1}^{t} & =u_{1}^{t} & u_{1}^{t} & :=\mathcal{N}\left(0,\delta^{tt'}\frac{1}{\alpha}C_{0}^{tt'}\right)\\
h_{2k}^{t} & =u_{2k}^{t}+\lambda_{k}h_{0k}^{t} & u_{2k}^{t} & :=\mathcal{N}\left(0,\delta^{tt'}\lambda_{k}C_{1}^{tt'}\right)\\
h_{3}^{t} & =u_{3}+\frac{1}{\nu}\sum_{t'=0}^{t}R_{4\to2}^{tt'}h_{3}^{t'} & u_{3} & :=\mathcal{N}\left(0,\frac{1}{\nu}C_{2}\right)\\
h_{4k}^{t} & =u_{4k}+\sum_{t'=0}^{t}R_{3}^{tt'}h_{2k}^{t'} & u_{4} & :=\mathcal{N}\left(0,C_{3}\right)
\end{align*}
\begin{align*}
C_{0} & :=\frac{1}{d}\sum_{k}\lambda_{k}\left\langle h_{0k}h_{0k}\right\rangle  & R_{2\to0}^{D}:= & \frac{1}{d}\sum_{k}\lambda_{k}\left\langle \frac{\delta h_{0k}}{\delta u_{2k}}\right\rangle \\
C_{1} & :=\left\langle h_{1}h_{1}\right\rangle  & R_{1}:= & \left\langle \frac{\delta h_{1}}{\delta u_{1}}\right\rangle \\
C_{2} & :=\frac{1}{d}\sum_{k}\left\langle h_{2k}h_{2k}\right\rangle  & R_{4\to2}:= & \frac{1}{d}\sum_{k}\left\langle \frac{\delta h_{2k}}{\delta u_{4k}}\right\rangle \\
C_{3} & :=\left\langle h_{3}h_{3}\right\rangle  & R_{3}:= & \left\langle \frac{\delta h_{3}}{\delta u_{3}}\right\rangle.
\end{align*}
Therefore, the dynamics can be described as
\[
h_{0}^{t+1}=h_{0}^{t}-\eta\left(C_{3}^{tt}\right)^{-1/2}\left(u_{4k}+\sum_{t'=0}^{t}R_{3}^{tt'}\left(u_{2k}^{t'}+\lambda_{k}h_{0k}^{t'}\right)\right)
\]

The test $\mathcal{L}$ and train $L$ loss can be expressed as 
\[
\mathcal{L}\left(t\right)=\frac{1}{d}\sum_{k}\lambda_{k}h_{0k}^{2}\left(t\right)=C_{0}\left(t,t\right);\quad L\left(t\right)=C_{1}\left(t,t\right)
\]
and since it is online, we only have to account for the difference
in normalization (normalized by $P$ vs by $d$) 
\[
C_{0}\left(t,t\right)=\alpha C_{1}\left(t,t\right)
\]
Which can also be seen more directly by considering $h_{1}^{t}=u_{1}^{t}\Rightarrow\left\langle h_{1}^{t}h_{1}^{t'}\right\rangle =\left\langle u_{1}^{t}u_{1}^{t'}\right\rangle \Rightarrow C_{1}^{tt'}=\delta^{tt'}\frac{1}{\alpha}C_{0}^{tt'}$.

However, these dynamics still describe the random field $h_1$ and injected colored noises $u_{2,k},~u_{4,k}$. We would like to have a fully deterministic description in terms of the concentration response and correlation functions.

\subsection{Closing the DMFT equations}
Here, we derive a closed set of equations in terms of correlations and responses. Taking the derivative of the $h_{3}$ equation by $u_{3}$ 
\[
\frac{\delta}{\delta u_{3}^{s}}h_{3}^{t}=\frac{\delta}{\delta u_{3}}\left(u_{3}+\frac{1}{\nu}\sum_{t'=0}^{t}R_{4\to2}^{tt'}h_{3}^{t'}\right)
\]
and using $R_3 = \tfrac{\delta h_{3}^{t}}{\delta u_{3}^{s}}$
\[
R_{3}^{ts}=\delta^{ts}+\frac{1}{\nu}\sum_{t'=0}^{t}R_{4\to2}^{tt'}R_{3}^{ts}.
\]
Similarly taking a derivative $h_{2}$  w.r.t. $u_{4}$
\[
R_{4\to2k}^{ts}=\lambda_{k}R_{4\to0k}^{ts}
\]
repeating the same process for $R_{4\to0}$ and $R_4$
\[
\left(\delta^{t+1,t'}-\delta^{t,t'}\right)R_{4\to0k}^{t',s}=-\eta\left(\nu C_{3}^{tt}\right)^{-1/2}R_{4k}^{ts}
\]
\[
R_{4k}^{ts}=\delta^{ts}+\sum_{t'=0}^{t}R_{3}^{tt'}R_{4\to2k}^{t's}.
\]
Using matrix notation in time domain, and introduction the integration operator $\Theta^{ts} = \eta 1_{s<t} (\nu c_3^s)^{-1/2}$
\[
\Theta=\left(\begin{array}{ccccc}
0 & 0 & 0 & 0 & 0\\
\frac{\eta}{\sqrt{\nu C_{3}^{00}}} & 0 & 0 & 0 & 0\\
\frac{\eta}{\sqrt{\nu C_{3}^{00}}} & \frac{\eta}{\sqrt{\nu C_{3}^{11}}} & dots & \vdots & \vdots\\
\frac{\eta}{\sqrt{\nu C_{3}^{00}}} & \frac{\eta}{\sqrt{\nu C_{3}^{11}}} & dots & 0 & \vdots\\
\frac{\eta}{\sqrt{\nu C_{3}^{00}}} & \frac{\eta}{\sqrt{\nu C_{3}^{11}}} & dots & \frac{\eta}{\sqrt{\nu C_{3}^{tt}}} & dots
\end{array}\right)
\]
we can solve the above equations for $R_3$ to find a self-consistent equation
\begin{equation}
R_{3}=I-\frac{1}{\nu d}\sum_{k=1}^{d}\lambda_{k}\left(\Theta \left[I+\lambda_{k}R_{3}\Theta \right]^{-1}\right)R_{3}.
\label{eq:time_domain_R3_eq}
\end{equation}
We can also close the equation of $h_{0}$
\[
h_{0k}=h_{0k}^{0}j-\Theta\left(u_{4k}+R_{3}\left(u_{2k}+\lambda_{k}h_{0k}\right)\right)
\]
with $j$ the all ones time vector. We can formally solve this to
find a fully explicit equation for $h_{0}$
\[
h_{0k}=\left[I+\lambda_{k}\Theta R_{3}\right]^{-1}\left(h_{0k}^{0}j-\Theta\left(u_{4k}+R_{3}u_{2k}\right)\right).
\]
Similarly for $h_{2}$
\[
h_{2k}^{t}=u_{2k}^{t}+\lambda_{k}\left[I+\lambda_{k}\Theta R_{3}\right]^{-1}\left(h_{0k}^{0}j-\Theta\left(u_{4k}+R_{3}u_{2k}\right)\right)
\]

\subsubsection{Correlations}

Starting with the $h_{3}$ equation we see
\[
h_{3}^{t}-\frac{1}{\nu}\sum_{t'=0}^{t}R_{4\to2}^{tt'}h_{3}^{t'}=u_{3}
\]
multiplying by the same equation at time $s$ and taking expectation
\[
\left\langle \left(h_{3}^{t}-\frac{1}{\nu}\sum_{t'=0}^{t}R_{4\to2}^{tt'}h_{3}^{t'}\right)\left(h_{3}^{s}-\frac{1}{\nu}\sum_{s'=0}^{t}R_{4\to2}^{ss'}h_{3}^{s'}\right)\right\rangle =\left\langle u_{3}^{t}u_{3}^{s}\right\rangle 
\]
\[
\nu C_{3}^{ts}-\sum_{s'=0}^{t}R_{4\to2}^{ss'}C_{3}^{s't}-\sum_{t'=0}^{t}R_{4\to2}^{tt'}C_{3}^{t's}+\frac{1}{\nu}\sum_{t'=0}^{t}\sum_{s'=0}^{t}R_{4\to2}^{tt'}R_{4\to2}^{ss'}C_{3}^{t's'}=C_{2}
\]
using $R_{3}=I+\frac{1}{\nu}R_{4\to2}R_{3}$ and simplifying we can rewrite this as
\[
C_{3}=\frac{1}{\nu}R_{3}C_{2}R_{3}^{\top}.
\]
Similarly calculating $C_{2}$ directly
\[
h_{2k}^{t}=u_{2k}^{t}+\lambda_{k}\left[I+\lambda_{k}\Theta R_{3}\right]^{-1}\left(h_{0k}^{0}j-\Theta\left(u_{4k}+R_{3}u_{2k}\right)\right)
\]
\begin{align*}
\left\langle h_{2k}h_{2k}^{\top}\right\rangle &=\left\langle \left(u_{2k}+\lambda_{k}\left[I+\lambda_{k}\Theta R_{3}\right]^{-1}\left(h_{0k}^{0}j-\Theta\left(u_{4k}+R_{3}u_{2k}\right)\right)\right)
\right.\\&\qquad\left.
\left(u_{2k}^{t}+\lambda_{k}\left[I+\lambda_{k}\Theta R_{3}\right]^{-1}\left(h_{0k}^{0}j-\Theta\left(u_{4k}+R_{3}u_{2k}\right)\right)\right)^{\top}\right\rangle 
\end{align*}
we can simplify it to
\begin{align*}
C_{2k}&=\left(\lambda_{k}c_{1}-\lambda_{k}^{2}c_{1}\left(\left[I+\lambda_{k}\Theta R_{3}\right]^{-1}\Theta R_{3}\right)^{\top}-\lambda_{k}^{2}\left[I+\lambda_{k}\Theta R_{3}\right]^{-1}\Theta R_{3}c_{1}
\right. \\ 
&\left.\qquad +\lambda_{k}^{2}\left[I+\lambda_{k}\Theta R_{3}\right]^{-1}\left(\left(\left\langle \left(h_{0k}^{0}\right)^{2}\right\rangle jj^{\top}+\Theta\left(C_{3}+\lambda_{k}R_{3}c_{1}R_{3}^{\top}\right)\Theta^{\top}\right)\right)\left[I+\lambda_{k}\Theta R_{3}\right]^{-\top}\right).
\end{align*}
We also have that 
\[
C_{1}^{tt'}=\delta^{tt'}c_{1}^{t}=\delta^{tt'}\frac{1}{\alpha}C_{0}^{tt'}.
\]
Finally a direct calculation for $C_{0k}=\left\langle h_{0k}h_{0k}^{\top}\right\rangle $ gives 
\begin{align*}
C_{0k}&=\left\langle \left(\left[I+\lambda_{k}\Theta R_{3}\right]^{-1}\left(h_{0k}^{0}j-\Theta\left(u_{4k}+R_{3}u_{2k}\right)\right)\right)\left(\left[I+\lambda_{k}\Theta R_{3}\right]^{-1}\left(h_{0k}^{0}j-\Theta\left(u_{4k}+R_{3}u_{2k}\right)\right)\right)^{\top}\right\rangle 
\\
&=\left[I+\lambda_{k}\Theta R_{3}\right]^{-1}\left(\left(\left\langle \left(h_{0k}^{0}\right)^{2}\right\rangle jj^{\top}+\Theta\left(C_{3}+\lambda_{k}R_{3}c_{1}R_{3}^{\top}\right)\Theta^{\top}\right)\right)\left[I+\lambda_{k}\Theta R_{3}\right]^{-\top}
\end{align*}
which concludes the all required correlation functions.

\subsection{Results}
\label{subapp:DMFT_closed}

To get a more intutive understanding we can define $H_{k}:=\left[I+\lambda_{k}\Theta R_{3}\right]^{-1}$
and $\left\langle f_{k}\right\rangle _{k}:=\frac{1}{d}\sum_{k}f_{k}$,
$\left\langle \left(h_{0k}^{0}\right)^{2}\right\rangle _{h^{0}}:=\beta_{k}^{2}$ and re-write the $C_0$ equation
\begin{align*}
C_{0k} & =\beta_{k}^{2}H_{k}j\left(jH_{k}\right)^{\top}+H_{k}\Theta\left(C_{3}+\lambda_{k}R_{3}c_{1}R_{3}^{\top}\right)\left(\Theta H_{k}\right)^{\top}\\
C_{0} & =\left\langle \lambda_{k}\left(\beta_{k}^{2}H_{k}j\left(jH_{k}\right)^{\top}+H_{k}\Theta\left(C_{3}+\lambda_{k}R_{3}c_{1}R_{3}^{\top}\right)\left(\Theta H_{k}\right)^{\top}\right)\right\rangle _{k}
\end{align*}
we now see that the first term propagates forward the initial conditions, and the
second propagates forward the current state. 

For $C_{2}$ we may notice
that $I-\lambda_{k}H_{k}\Theta R_{3}=H_{k}\left(H_{k}^{-1}-\lambda_{k}\Theta R_{3}\right)=H_{k}\left(I+\lambda_{k}\Theta R_{3}-\lambda_{k}\Theta R_{3}\right)=H_{k}$
so 
\begin{align*}
C_{2k} & =\left(\lambda_{k}\left(I-\lambda_{k}\left(H_{k}\Theta R_{3}\right)\right)c_{1}\left(I-\lambda_{k}\left(H_{k}\Theta R_{3}\right)^{\top}\right)+\lambda_{k}^{2}H_{k}\left(\beta_{k}^{2}jj^{\top}+\Theta C_{3}\Theta^{\top}\right)H_{k}^{\top}\right)\\
 & =\lambda_{k}^{2} \beta_k^2 H_{k}j\left(jH_{k}\right)^{\top}+\lambda_{k}H_{k}c_{1}H_{k}^{\top}+\lambda_{k}^{2}H_{k}\Theta C_{3}\left(\Theta H_{k}\right)^{\top}
 \\
C_{2} & =\left\langle \lambda_{k}^{2}\beta_k^2  H_{k}j\left(jH_{k}\right)^{\top}+\lambda_{k}H_{k}\left(c_{1}+\lambda_{k}\Theta C_{3}\Theta^{\top}\right)H_{k}^{\top}\right\rangle _{k}
\end{align*}

we can see the first term again propagates the initial conditions
and the second propagates forward the current state.

To complete these we have the simpler relations
\begin{equation}
C_{1}^{tt'}=\delta^{tt'}c_{1}^{t}=\delta^{tt'}\frac{1}{\alpha}C_{0}^{tt'}
\label{eq:C0_C1_c1_relation}
\end{equation}
\[
C_{3}=\frac{1}{\nu}R_{3}C_{2}R_{3}^{\top}.
\]
The diagonal of $C_{2}$ takes an especially simple form pretty simple
\[
C_{2k}^{tt}=\left\langle \left(u_{2k}^{t}+\lambda_{k}h_{0k}^{t}\right)\left(u_{2k}^{t}+\lambda_{k}h_{0k}^{t}\right)\right\rangle =\left\langle u_{2k}^{t}u_{2k}^{t}\right\rangle +2\lambda_{k}\left\langle u_{2k}^{t}h_{0k}^{t}\right\rangle +\lambda_{k}^{2}\left\langle h_{0k}^{t}h_{0k}^{t}\right\rangle 
\]
since $u_{2k}^{t},h_{0k}^{t}$ are uncorrelated by causality we have
(defining $C_{0k}^{tt'}:=\left\langle h_{0k}^{t}h_{0k}^{t}\right\rangle $)
\[
\left\langle u_{2k}^{t}u_{2k}^{t}\right\rangle +\lambda_{k}^{2}\left\langle h_{0k}^{t}h_{0k}^{t}\right\rangle =\lambda_{k}C_{1}^{tt}+\lambda_{k}^{2}\left\langle h_{0k}^{t}h_{0k}^{t}\right\rangle =\lambda_{k}C_{1}^{tt}+\lambda_{k}^{2}C_{0k}^{tt}
\]
now averaging over modes to get $C_{2}^{tt}$
\[
c_{2}^{t}=C_{2}^{tt}=C_{1}^{tt}\frac{1}{d}\sum_{k}\lambda_{k}+\frac{1}{d}\sum_{k}\lambda_{k}^{2}C_{0k}^{tt}=\bar{\lambda}C_{1}^{tt}+\frac{1}{d}\sum_{k}\lambda_{k}^2C_{0k}^{tt}
\]
using the equal time identity of $C_{0},C_{1}$
\[
c_{2}^{t}=C_{2}^{tt}=\frac{\left\langle \lambda_{k}\right\rangle _{k}}{\alpha}C_{0}^{tt}+\left\langle \lambda_{k}^2C_{0k}^{tt}\right\rangle _{k}
\]
We can also see the first term is SGD noise as it vanishes for $\alpha\to1$

As the operators in these equations are causal (and therefore lower triangular) we can solve them numerically by stepping through time.

\section{Random Matrix Theory Interpretation}
\label{app:rmt_interp}

In this section we show that the Z-transform of the response function can be interpreted as the companion Stieltjes transform to $\Lambda^{1/2}A^{T}A\Lambda^{1/2}$, giving another interpretation to its role, and showing that the EoS condition is equivalent to the familiar sharpness stability condition $\neff = 2/\lambda_{\rm max}$; where $\lambda_{\rm max}$ is the largest eigenvalue fo the random matrix $\Lambda^{1/2}A^{T}A\Lambda^{1/2}$.

We start by showing the equivalence to the Stieltjes transform; by definition
\[
h_{0k}=h_{0k}^{0}j-\Theta h_{4}.
\]
now applying the cavity argument only to the $\psi$ randomness
\begin{align*}
h_{3}^{t} & =\frac{1}{\sqrt{\nu d}}Ah_{2}\\
 & =\frac{1}{\sqrt{\nu d}}A\left(u_{2}\left(t\right)+\Lambda\bar{h}_{0}\left(t\right)\right)\\
 & =\frac{1}{\sqrt{\nu d}}A\left(u_{2}\left(t\right)+\Lambda\left(h_{0k}^{0}j-\Theta h_{4}\right)\right)\\
 & =\frac{1}{\sqrt{\nu d}}A\left(u_{2}\left(t\right)+\Lambda\left(h_{0k}^{0}j-\Theta\frac{1}{\sqrt{N}}A^{T}h_{3}^{t}\right)\right).
\end{align*}
To find the response operator, we introduce a small source $u_{3}$ to $h_3$ such that $h_{3}^{t}=\frac{1}{\sqrt{\nu d}}Ah_{2}+u_{3}$ and
take a derivative w.r.t. to it
\begin{align*}
\frac{\delta h_{3}}{\delta u_{3}} & =I-\Lambda\frac{1}{\sqrt{\nu d}}A\Theta\frac{1}{\sqrt{N}}A^{T}\frac{\delta h_{3}}{\delta u_{3}}
\end{align*}
where $\frac{\delta u_{2}}{\delta u_{3}}=0$ since they are two unrelated noise sources. We can now solve for $R_{3}$
\[
\left(I+\frac{1}{\sqrt{\nu d}}A\Lambda\Theta\frac{1}{\sqrt{N}}A^{T}\right)\frac{\delta h_{3}}{\delta u_{3}}=I
\]
\[
\frac{\delta h_{3}}{\delta u_{3}}=\left[I+A\Lambda\Theta\frac{1}{N}A^{T}\right]^{-1}
\]
Noting that $\Theta$ is an operator in time domain, whereas $A,\Lambda$ are operators in feature and input space we can commute them to find
\[
\frac{\delta h_{3}}{\delta u_{3}}=\left[I+\Theta\frac{A\Lambda A^{T}}{N}\right]^{-1}
\]
Z-transforming the equations and then multiplying by $\hat{\Theta}$
\[
\hat{\Theta}\frac{\delta h_{3}}{\delta u_{3}}=\left[\hat{\Theta}^{-1}+\frac{A\Lambda A^{T}}{N}\right]^{-1}
\]
finally, averaging in feature space we see that $\underline{m}:=\hat{\Theta}\hat{R}_{3}=\hat{\Theta}\frac{1}{N}\tr\left[\frac{\delta h_{3}}{\delta u_{3}}\right]$
is the Stieltjes transform of $\frac{A\Lambda A}{N}$ with an argument
$w=-\hat{\Theta}^{-1}=-\frac{z-1}{\neff}$
\[
\underline{m}=\tr\left[\frac{A\Lambda A^{T}}{N}-w\right]^{-1}.
\]
We can now reinterpret the self consistent equation for $\hat{R}_{3}$
in terms of $\underline{m},w$
\[
\frac{1}{\hat{R}_{3}\left(z\right)}=1+\frac{1}{\nu}\left\langle \frac{\lambda_{k}\eta_{\text{eff}}}{z-1+\lambda_{k}\eta_{\text{eff}}\hat{R}_{3}\left(z\right)}\right\rangle _{k}\iff w=-\frac{1}{\underline{m}}+\frac{1}{\nu}\left\langle \frac{\lambda_{k}}{1+\lambda_{k}\underline{m}}\right\rangle _{k}
\]
which is the Silverstein equation for the companion Stieltjes transform
of $\Lambda^{1/2}A^{T}A\Lambda^{1/2}/N$, i.e. the Silverstein equation of the matrix
$\frac{A\Lambda A}{N}$ written in input space ($d$); noting that the two matrices share the same non-zero spectrum.

This perspective gives a particularly sharp interpretation for the
edge of stability phase condition, and justify its name. The largest
eigenvalue of a Wishart is given by $\underline{m}(-1)$ that solves
\[
1=\frac{1}{\nu}\left\langle \frac{\lambda_{k}^{2}\underline{m}^{2}}{\left(1+\lambda_{k} \underline{m}\right)^{2}}\right\rangle 
\]
and given this solution for $\underline{m}$, the largest eigenvalue $\lambda_{\text{max}}$ is explicit
\[
\lambda_{\text{max}}=-\frac{1}{\underline{m}}+\frac{1}{\nu}\left\langle \frac{\lambda_{k}}{1+\lambda_{k} \underline{m}}\right\rangle .
\]
Now rewriting our conditions in terms of $w,\underline{m}$ 
\[
\nu=\left\langle \left(\frac{\lambda_{k}\frac{\eta_{\text{eff}}}{2}\hR\left(-1\right)}{-1+\lambda_{k}\frac{\eta_{\text{eff}}}{2}\hR\left(-1\right)}\right)^{2}\right\rangle \iff1=\frac{1}{\nu}\left\langle \left(\frac{\lambda_{k}\underline{m}\left(z=-1\right)}{1+\lambda_{k}\underline{m}\left(z=-1\right)}\right)^{2}\right\rangle 
\]
\[
r_{-}^{-1}=1+\frac{1}{\nu}\left\langle \frac{\lambda_{k}\eta_{\text{eff}}}{-2+\lambda_{k}\eta_{\text{eff}}r_{-}}\right\rangle _{k}\iff\frac{2}{\neff}=-\frac{1}{\underline{m}}+\frac{1}{\nu}\left\langle \frac{\lambda_{k}}{1+\lambda_{k}\underline{m}}\right\rangle _{k}
\]
so we find that the effective learning is exactly $\neff=\frac{2}{\lambda_{\text{max}}}$
- the canonical EoS, expressed self-consistently for a random matrix.

\section{Long Time Behavior and Effective Learning Rate Steady State}
\label{app:pole_hunt}
As discussed in the main text, after a certain amount of time, the solution to the DMFT equations in section~\ref{subapp:DMFT_closed} settles on a steady state for the effective learning rate. Our results uncover three possible phases for this steady state: (1) edge of stability, (2) edge of stochastic stability, and (3) width-limited regime. 

To find these steady states, we assume $\neff^t \equiv \neffi$ and solve for $\neffi$ self-consistently. Note that this assumption is \emph{not} limited to a fixed point; for example, in a limit cycle, the $\neffi$ we find would be its zero-frequency component.  For the rest of this section, we drop the $\infty$ from $\neffi$ for notational brevity.

To solve these equations, we use the z-transform, which diagonalizes time operators that are TTI, i.e. $A(t,s)=A(t-s)$. 
We start by z-transforming the time integration operator $\Theta$. Writing the operator $\Theta^{ts}$ as a function function of the time lag $u=t-s$
\[
\Theta^{u}=\eta_{\text{eff}}\sum_{v>0}\delta^{u,v}
\]
where only positive time lags contribute and using the definition of the Z-transform we find
\[
\hat{\Theta}=\eta_{\text{eff}}\sum_{u}\sum_{v>0}\delta^{u,v}z^{-u}=\eta_{\text{eff}}\sum_{u>0}z^{-u}=\eta_{\text{eff}}\frac{1}{z-1}.
\]
Inspecting the self-consistent equation for $R_3$ we see that if $\Theta$ is TTI then $R_3$ must be TTI too, transforming Eq. ~\ref{eq:time_domain_R3_eq} we find
\[
\hat{R}_{3}(z)=1-\frac{1}{\nu d}\sum_{k=1}^{d}\frac{\lambda_{k}\eta_{\text{eff}}\frac{1}{z-1}\hat{R}_{3}(z)}{1+\lambda_{k}\eta_{\text{eff}}\frac{1}{z-1}\hat{R}_{3}(z)}
\]
where the physical solution lives on the branch connected to $\lim_{z\to\infty}\hat{R}_{3}=1\iff R_{3}\left(t-s=0\right)=1$.

Similarly, we can use the double z-transform to describe the correlation functions, which are not TTI.\footnote{To see that, consider a system with the fact that $C_0^{tt}=\mathcal{L}^t$; if it were TTI $C_0^{tt}\equiv {\rm const}$, so the system could never move.} We use the notation
\begin{align}
\tilde{C}_k(z_1,z_2) &:= \sum_{t_1,t_2} C^{t_1,t_2}_j z_1^{-t_1} z_2^{-t_2} 
\\
\hat{c}_i(\zeta) &:= \sum_{t} C^{t,t}_i \zeta^{-t}
\end{align}
and we have the important relation between the two
\[
\hat{c}_j = \intop_0^{2\pi} \frac{d \theta}{2\pi }\tilde{C}_{j}\left(\sqrt{\zeta} e^{i\theta},\sqrt{\zeta} e^{-i \theta} \right),
\]
as shown in App.~\ref{app:z-transform}.

Using the variables $\zeta,w$ with $z_1=\sqrt{\zeta} w$ and $z_2=\sqrt{\zeta}/w$, we then have the Z-transforms of $C_0,C_3$
\begin{align}
\tilde{C}_{0}\left(\sqrt{\zeta} w,\sqrt{\zeta}/w\right)&=\left\langle \lambda_{k}\beta_{k}^{2}\widehat{H_{k}j}\left(\sqrt{\zeta} w\right)\widehat{H_{k}j}\left(\sqrt{\zeta}/w\right)\right\rangle _{k}\nonumber\\&+\frac{\eta_{\text{eff}}^{2}\left\langle \lambda_{k}\hat{H_{k}}\left(\sqrt{\zeta} w\right)\hat{H}_{k}\left(\sqrt{\zeta}/w\right)\right\rangle }{\left(\sqrt{\zeta} w-1\right)\left(\sqrt{\zeta}/w-1\right)}\tilde{C}_{3}\left(\sqrt{\zeta} w,\sqrt{\zeta}/w\right)\nonumber\\&+\frac{\eta_{\text{eff}}^{2}\hat{R}_{3}\left(\sqrt{\zeta} w\right)\hat{R}_{3}\left(\sqrt{\zeta}/w\right)\left\langle \lambda_{k}^{2}\hat{H_{k}}\left(\sqrt{\zeta} w\right)\hat{H}_{k}\left(\sqrt{\zeta}/w\right)\right\rangle }{\left(\sqrt{\zeta} w-1\right)\left(\sqrt{\zeta}/w-1\right)}\hat{c}_{1}\left(\zeta\right)\nonumber\\\tilde{C}_{3}\left(\sqrt{\zeta} w,\sqrt{\zeta}/w\right)&=\frac{\hat{R}_{3}\left(\sqrt{\zeta} w\right)\hat{R}_{3}\left(\sqrt{\zeta}/w\right)}{\nu-\eta_{\text{eff}}^{2}\frac{\hat{R}_{3}\left(\sqrt{\zeta} w\right)}{\sqrt{\zeta} w-1}\frac{\hat{R}_{3}\left(\sqrt{\zeta}/w\right)}{\sqrt{\zeta}/w-1}\left\langle \lambda_{k}^{2}\hat{H_{k}}\left(\sqrt{\zeta} w\right)\hat{H}_{k}\left(\sqrt{\zeta}/w\right)\right\rangle _{k}}\nonumber\\&\times\left(\left\langle \lambda_{k}^{2}\widehat{H_{k}j}\left(\sqrt{\zeta} w\right)\widehat{H_{k}j}\left(\sqrt{\zeta}/w\right)\right\rangle _{k}+\left\langle \lambda_{k}\hat{H_{k}}\left(\sqrt{\zeta} w\right)\hat{H}_{k}\left(\sqrt{\zeta}/w\right)\right\rangle _{k}\hat{c}_{1}\left(\zeta\right)\right)
\end{align}
Using the relation $c_{1}^{t}=\frac{1}{\alpha}C_{0}^{tt}$ in \eqref{eq:C0_C1_c1_relation} we find the expression for the loss $\hat{c}_1$ and the gradient norm squared $\hat{c}_3$
\begin{align}
    \hat{c}_{1}\left(\zeta\right) &=\frac{\hat{\mathcal{B}}\left(\zeta\right)}{1-\hat{\mathcal{K}}\left(\zeta\right)}
    \nonumber\\
    \hat{c}_{3}\left(\zeta\right) &=
    \hat{\mathcal{S}}\left(\zeta\right)+\hat{\mathcal{T}}\left(\zeta\right)\frac{\hat{\mathcal{B}}\left(\zeta\right)}{1-\hat{\mathcal{K}}\left(\zeta\right)}
\end{align}
with the following definitions:
\begin{align*}\hat{\mathcal{B}}\left(\zeta\right)&:=\frac{1}{\alpha}\oint\frac{dw}{2\pi iw}\left(B_{1}+\frac{M_{1}}{\nu-M_{2}}B_{2}\right)\\\hat{\mathcal{K}}\left(\zeta\right)&:=\frac{1}{\alpha}\oint\frac{dw}{2\pi iw}\left(\frac{\left(\sqrt{\zeta}w-1\right)\left(\sqrt{\zeta}/w-1\right)}{\eta_{\text{eff}}^{2}\hat{R}\left(\sqrt{\zeta}w\right)\hat{R}_{3}\left(\sqrt{\zeta}/w\right)}\frac{M_{1}^{2}\left(\sqrt{\zeta}w,\sqrt{\zeta}/w\right)}{\nu-M_{2}\left(\sqrt{\zeta}w,\sqrt{\zeta}/w\right)}+M_{2}\left(\sqrt{\zeta}w,\sqrt{\zeta}/w\right)\right)\\\hat{\mathcal{S}} (\zeta)&:=\oint\frac{dw}{2\pi iw}\frac{\hat{R}\left(\sqrt{\zeta}w\right)\hat{R}_{3}\left(\sqrt{\zeta}/w\right)B_{2}\left(\sqrt{\zeta}w,\sqrt{\zeta}/w\right)}{\nu-M_{2}\left(\sqrt{\zeta}w,\sqrt{\zeta}/w\right)}\\\hat{\mathcal{T}} (\zeta)&:=\oint\frac{dw}{2\pi iw}\frac{\left(\sqrt{\zeta}w-1\right)\left(\sqrt{\zeta}/w-1\right)}{\eta_{\text{eff}}^{2}}\frac{M_{1}\left(\sqrt{\zeta}w,\sqrt{\zeta}/w\right)}{\nu-M_{2}\left(\sqrt{\zeta}w,\sqrt{\zeta}/w\right)}
\\
B_{a}\left(z_{1},z_{2}\right)&:=\left\langle \lambda_{k}^{a}\beta_{k}^{2}\frac{z_{1}}{z_{1}-1+\lambda_{k}\eta_{\text{eff}}\hat{R}_{3}\left(z_{1}\right)}\frac{z_{2}}{z_{2}-1+\lambda_{k}\eta_{\text{eff}}\hat{R}_{3}\left(z_{2}\right)}\right\rangle \\M_{a}\left(z_{1},z_{2}\right)&:=\left\langle \lambda_{k}^{a}\frac{\eta_{\text{eff}}\hat{R}_{3}\left(z_{1}\right)}{z_{1}-1+\lambda_{k}\eta_{\text{eff}}\hat{R}_{3}\left(z_{1}\right)}\frac{\eta_{\text{eff}}\hat{R}_{3}\left(z_{2}\right)}{z_{2}-1+\lambda_{k}\eta_{\text{eff}}\hat{R}_{3}\left(z_{2}\right)}\right\rangle .
\end{align*}

As explained in Appendix~\ref{app:z-transform}, we can find the long-time behavior using the final value theorem
\[
\neffi = \eta/\sqrt{c_3(\infty)} = \eta / \sqrt{\lim_{\zeta\to 1} (\zeta-1)\hat{c}_3(\zeta)}  
\]
giving us a self-consistency condition and allowing us to determine $\neffi$ (where we kept the $\infty$ superscript explicit to emphasize the steady-state condition).

Thus, self-consistency requires a simple pole at $\zeta=1$. Inspecting the terms we find three candidates
\begin{align*}
    (1)~~ &
    \lim _{\zeta \to 1} \left|e^{i\theta}-1+\lambda_{k}\eta_{\text{eff}}\hat{R}_{3}\right|^{2}=0
    ;\\
    (2)~~ &
    \lim _{\zeta \to 1} \nu-\left\langle \frac{\lambda_{k}^{2}}{\left|e^{i\theta}-1+\lambda_{k}\eta_{\text{eff}}\hat{R}_{3}\right|^{2}}\right\rangle \eta_{\text{eff}}^{2}\left|\hat{R}_{3}\right|^{2}=0
    ;\\
    (3)~~ &
    \lim _{\zeta \to 1}\hat{\mathcal{K}}\left(\zeta\right)-1=0.
\end{align*}
As we show below, these three conditions correspond to the three phases from the main text. Before detailing the phases, we evaluate the contour integral in $\hB,\hK,\hS,\hT$ as they would be used throughout.

\subsection{Evaluating the contour integral in $\hK,\hT$}
\label{subapp:contour_int_eval}

A central quantity we will use in this section is 
\[
g_{k}\left(z\right):=\frac{\lambda_{k}}{q-\lambda_{k}};\quad q:=\frac{1-z}{\neff\hat{R}_{3}\left(z\right)}
\]

first we show five identities 

\paragraph{(Identity 1)}

Multiplying the fraction in the definition ofc$g_{k}$ by the conjugate
of the denominator

\begin{equation}
g_{k}\left(z\right)=\frac{\lambda_{k}\left(\bar{q}-\lambda_{k}\right)}{\left|q-\lambda_{k}\right|^{2}}=\frac{\lambda_{k}\bar{q}}{\left|q-\lambda_{k}\right|^{2}}-\frac{\lambda_{k}^{2}}{\left|q-\lambda_{k}\right|^{2}}=\frac{\bar{q}}{\lambda_{k}}\left|g_{k}\right|^{2}-\left|g_{k}\right|^{2}\label{eq:identity_i}
\end{equation}

\paragraph{(Identity II)}

Taking the conjugate of \ref{eq:identity_i} and subtracting it from
\ref{eq:identity_i} while noting that $\overline{g_{k}\left(z\right)}=g_{k}\left(\overline{z}\right)$,
we find
\begin{equation}
g_{k}\left(z\right)-g_{k}\left(\overline{z}\right)=\left(\overline{q}-q\right)\frac{\left|g_{k}\right|^{2}}{\lambda_{k}}\label{eq:identity_ii}
\end{equation}

\paragraph{(Identity III)}

similarly multiplying \ref{eq:identity_i} by $q$ and the conjugate
by $\bar{q}$ and subtracting
\begin{equation}
qg_{k}\left(z\right)-\bar{q}g_{k}\left(\overline{z}\right)=\frac{\left|q\right|^{2}}{\lambda_{k}}\left|g_{k}\right|^{2}-q\left|g_{k}\right|^{2}-\left(\frac{\left|q\right|^{2}}{\lambda_{k}}\left|g_{k}\right|^{2}-\bar{q}\left|g_{k}\right|^{2}\right)=\left(\bar{q}-q\right)\left|g_{k}\right|^{2}.\label{eq:identity_iii}
\end{equation}

\paragraph{(Identity IV)}

another useful identify is found by writing the $\hR$ self-consistent
equation in terms of $g$ we find for $\hR\neq0$
\begin{equation}
\hat{R}_{3}\left(z\right)=\left(1-\frac{1}{\nu}\frac{1}{\hR}\left\langle g_{k}\right\rangle _{k}\right)^{-1}\Rightarrow q\left\langle g_{k}\right\rangle _{k}=q\nu\left(\hat{R}_{3}\left(z\right)-1\right)=\nu\left(\frac{1-z}{\neff}-q\right)\label{eq:identity_iv}
\end{equation}

\paragraph{(Identity V)}

finally, using \ref{eq:identity_iii} and then identity \ref{eq:identity_iv}
we find
\[
\nu-\left\langle \left|g_{k}\right|^{2}\right\rangle =\nu-\left\langle \frac{qg_{k}\left(z\right)-\bar{q}g_{k}\left(\overline{z}\right)}{\left(\bar{q}-q\right)}\right\rangle =\nu-\frac{\nu\left(\frac{1-z}{\neff}-q\right)-\nu\left(\frac{1-\bar{z}}{\neff}-\bar{q}\right)}{\left(\bar{q}-q\right)}
\]
which can be simplified to
\begin{equation}
\nu-\left\langle \left|g_{k}\right|^{2}\right\rangle =-\frac{\nu\left(\bar{z}-z\right)}{\neff\left(\bar{q}-q\right)}\label{eq:identity_v}
\end{equation}

\subsubsection{Simplifying $\protect\hT$}

we can now simplify $\hT$ on the contour integral where $z_{1}=\bar{z}_{2}$.
Replacing the denominator by the identity \ref{eq:identity_v} and
then rewriting using $g_{k}$
\[
\hT=\oint\frac{dz}{2\pi iz}\frac{\left|z-1\right|^{2}}{-\frac{\nu\left(\bar{z}-z\right)}{\neff\left(\bar{q}-q\right)}}\left\langle \frac{\lambda_{k}\left|\hat{R}_{3}\right|^{2}}{\left|z-1+\lambda_{k}\eta_{\text{eff}}\hat{R}_{3}\right|^{2}}\right\rangle =\oint\frac{dz}{2\pi iz}\frac{\left|z-1\right|^{2}}{-\frac{\nu\left(\bar{z}-z\right)}{\left(\bar{q}-q\right)}}\frac{1}{\neff}\left\langle \frac{\left|g_{k}\right|^{2}}{\lambda_{k}}\right\rangle 
\]
using \ref{eq:identity_ii} and then \ref{eq:identity_iv}
\[
\hT=\oint\frac{dz}{2\pi iz}\frac{\left|z-1\right|^{2}}{-\nu\left(\bar{z}-z\right)}\frac{1}{\neff}\left\langle g_{k}\left(z\right)-g_{k}\left(\overline{z}\right)\right\rangle =\oint\frac{dz}{2\pi iz}\frac{\left|z-1\right|^{2}}{-\nu\left(\bar{z}-z\right)}\frac{1}{\neff}\left(\nu\left(\hat{R}_{3}\left(z\right)-1\right)-\nu\left(\overline{\hat{R}_{3}\left(z\right)}-1\right)\right)
\]
simplifying
\[
\hT=\oint\frac{dz}{2\pi iz}\frac{\left|z-1\right|^{2}}{\neff}\left(\frac{\hat{R}_{3}\left(\overline{z}\right)-\hat{R}_{3}\left(z\right)}{\bar{z}-z}\right)
\]

\subsubsection{Simplifying $\protect\hK$}

Where we also used the fact $\overline{\hat{R}_{3}\left(z\right)}=\hat{R}_{3}\left(\overline{z}\right)$
which can be seen since the Laurent coefficients of $\hR$ are real
(as there are time steps in the z-transform and $R_{3}$ is real!).

Similarly for $\hK$
\[
\hat{\mathcal{K}}\left(\zeta=1\right)=\frac{1}{\alpha}\oint\frac{dz}{2\pi iz}\left(\frac{\left|e^{i\theta}-1\right|^{2}\left\langle \frac{\lambda_{k}}{\left|e^{i\theta}-1+\lambda_{k}\eta_{\text{eff}}\hat{R}_{3}\right|^{2}}\right\rangle ^{2}\eta_{\text{eff}}^{2}\left|\hat{R}_{3}\right|^{2}}{\nu-\left\langle \frac{\lambda_{k}^{2}}{\left|e^{i\theta}-1+\lambda_{k}\eta_{\text{eff}}\hat{R}_{3}\right|^{2}}\right\rangle \eta_{\text{eff}}^{2}\left|\hat{R}_{3}\right|^{2}}+\left\langle \frac{\lambda_{k}^{2}}{\left|e^{i\theta}-1+\lambda_{k}\eta_{\text{eff}}\hat{R}_{3}\right|^{2}}\right\rangle \eta_{\text{eff}}^{2}\left|\hat{R}_{3}\right|^{2}\right).
\]
\[
\hat{\mathcal{K}}\left(\zeta=1\right)=\frac{1}{\alpha}\oint\frac{dz}{2\pi iz}\frac{\left|z-1\right|^{2}\left\langle \frac{\left|g_{k}\right|^{2}}{\lambda_{k}}\right\rangle ^{2}}{\nu-\left\langle \left|g_{k}\right|^{2}\right\rangle }+\left\langle \left|g_{k}\right|^{2}\right\rangle =\frac{1}{\alpha}\oint\frac{dz}{2\pi iz}\frac{\left|q\right|^{2}\left\langle \frac{\left|g_{k}\right|^{2}}{\lambda_{k}}\right\rangle ^{2}+\nu\left\langle \left|g_{k}\right|^{2}\right\rangle -\left\langle \left|g_{k}\right|^{2}\right\rangle ^{2}}{\nu-\left\langle \left|g_{k}\right|^{2}\right\rangle }
\]

Going further requires a sixth, more involved identity

\paragraph{(Identity VI)}

We would need an additional identity 
\begin{equation}
\left(\overline{q}-q\right)\left(\left|q\right|^{2}\left\langle \frac{\left|g_{k}\right|^{2}}{\lambda_{k}}\right\rangle ^{2}-\left\langle \left|g_{k}\right|^{2}\right\rangle ^{2}\right)=q\left\langle g_{k}\left(z\right)\right\rangle \left\langle g_{k}\left(z\right)\right\rangle -\bar{q}\left\langle g_{k}\left(\overline{z}\right)\right\rangle \left\langle g_{k}\left(\bar{z}\right)\right\rangle \label{eq:identity_vi}
\end{equation}
we prove it below:taking \ref{eq:identity_ii} multiplying by $q$
and averaging, and multiplying it's conjugate by $\bar{q}$ and averaging
and then taking the product
\[
q\left(\left\langle g_{k}\left(z\right)\right\rangle -\left\langle g_{k}\left(\overline{z}\right)\right\rangle \right)\bar{q}\left(\left\langle g_{k}\left(\bar{z}\right)\right\rangle -\left\langle g_{k}\left(z\right)\right\rangle \right)=q\left(\overline{q}-q\right)\left\langle \frac{\left|g_{k}\right|^{2}}{\lambda_{k}}\right\rangle \bar{q}\left(q-\bar{q}\right)\left\langle \frac{\left|g_{k}\right|^{2}}{\lambda_{k}}\right\rangle 
\]
\[
\left|q\right|^{2}\left(\left\langle g_{k}\left(z\right)\right\rangle -\left\langle g_{k}\left(\overline{z}\right)\right\rangle \right)\left(\left\langle g_{k}\left(\bar{z}\right)\right\rangle -\left\langle g_{k}\left(z\right)\right\rangle \right)=\left|q\right|^{2}\left(2\left|q\right|^{2}-\bar{q}^{2}-q^{2}\right)\left(q-\bar{q}\right)\left\langle \frac{\left|g_{k}\right|^{2}}{\lambda_{k}}\right\rangle ^{2}
\]
\[
\left|q\right|^{2}\left(2\left\langle g_{k}\left(z\right)\right\rangle \left\langle g_{k}\left(\bar{z}\right)\right\rangle -\left\langle g_{k}\left(z\right)\right\rangle \left\langle g_{k}\left(z\right)\right\rangle -\left\langle g_{k}\left(\overline{z}\right)\right\rangle \left\langle g_{k}\left(\bar{z}\right)\right\rangle \right)=\left|q\right|^{2}\left(2\left|q\right|^{2}-\bar{q}^{2}-q^{2}\right)\left\langle \frac{\left|g_{k}\right|^{2}}{\lambda_{k}}\right\rangle ^{2}
\]
now averaging \ref{eq:identity_iii} and taking its product with its
conjugate
\[
\left(q\left\langle g_{k}\left(z\right)\right\rangle -\bar{q}\left\langle g_{k}\left(\overline{z}\right)\right\rangle \right)\left(\bar{q}\left\langle g_{k}\left(\bar{z}\right)\right\rangle -q\left\langle g_{k}\left(z\right)\right\rangle \right)=\left(\bar{q}-q\right)\left(q-\bar{q}\right)\left\langle \left|g_{k}\right|^{2}\right\rangle ^{2}
\]
\[
\left|q\right|^{2}\left\langle g_{k}\left(z\right)\right\rangle \left\langle g_{k}\left(\bar{z}\right)\right\rangle -q^{2}\left\langle g_{k}\left(z\right)\right\rangle \left\langle g_{k}\left(z\right)\right\rangle -\bar{q}^{2}\left\langle g_{k}\left(\overline{z}\right)\right\rangle \left\langle g_{k}\left(\bar{z}\right)\right\rangle +\left|q\right|^{2}\left\langle g_{k}\left(\overline{z}\right)\right\rangle \left\langle g_{k}\left(z\right)\right\rangle =\left(2\left|q\right|^{2}-\bar{q}^{2}-q^{2}\right)\left\langle \left|g_{k}\right|^{2}\right\rangle ^{2}
\]
\[
2\left|q\right|^{2}\left\langle g_{k}\left(z\right)\right\rangle \left\langle g_{k}\left(\bar{z}\right)\right\rangle -q^{2}\left\langle g_{k}\left(z\right)\right\rangle \left\langle g_{k}\left(z\right)\right\rangle -\bar{q}^{2}\left\langle g_{k}\left(\overline{z}\right)\right\rangle \left\langle g_{k}\left(\bar{z}\right)\right\rangle =\left(2\left|q\right|^{2}-\bar{q}^{2}-q^{2}\right)\left\langle \left|g_{k}\right|^{2}\right\rangle ^{2}.
\]

Subtracting the two results and simplifying yields the desired result
\[
\left(\overline{q}-q\right)\left(\left|q\right|^{2}\left\langle \frac{\left|g_{k}\right|^{2}}{\lambda_{k}}\right\rangle ^{2}-\left\langle \left|g_{k}\right|^{2}\right\rangle ^{2}\right)=q\left\langle g_{k}\left(z\right)\right\rangle \left\langle g_{k}\left(z\right)\right\rangle -\bar{q}\left\langle g_{k}\left(\overline{z}\right)\right\rangle \left\langle g_{k}\left(\bar{z}\right)\right\rangle 
\]

\paragraph{Simplifying $\protect\hK$ }

substituting into $\hK$ identities \ref{eq:identity_vi}, \ref{eq:identity_v},\ref{eq:identity_iii}
\[
\frac{1}{\alpha}\oint\frac{dz}{2\pi iz}\frac{\frac{q\left\langle g_{k}\left(z\right)\right\rangle \left\langle g_{k}\left(z\right)\right\rangle -\bar{q}\left\langle g_{k}\left(\overline{z}\right)\right\rangle \left\langle g_{k}\left(\bar{z}\right)\right\rangle }{\left(\overline{q}-q\right)}+\nu\frac{q\left\langle g_{k}\left(z\right)\right\rangle -\bar{q}\left\langle g_{k}\left(\overline{z}\right)\right\rangle }{\left(\bar{q}-q\right)}}{-\frac{\nu\left(\bar{z}-z\right)}{\neff\left(\bar{q}-q\right)}}
\]
we can now simplify
\[
\frac{1}{\alpha}\oint\frac{dz}{2\pi iz}\frac{q\left\langle g_{k}\left(z\right)\right\rangle \left(\left\langle g_{k}\left(z\right)\right\rangle +\nu\right)-\bar{q}\left\langle g_{k}\left(\overline{z}\right)\right\rangle \left(\left\langle g_{k}\left(\overline{z}\right)\right\rangle +\nu\right)}{-\frac{\nu\left(\bar{z}-z\right)}{\neff}}
\]
using \ref{eq:identity_iv} in the form $q\left(\left\langle g_{k}\right\rangle _{k}+\nu\right)=\nu\left(\frac{1-z}{\neff}\right)$
\[
\frac{1}{\alpha}\oint\frac{dz}{2\pi iz}\frac{\left\langle g_{k}\left(z\right)\right\rangle \nu\left(\frac{1-z}{\neff}\right)-\left\langle g_{k}\left(\overline{z}\right)\right\rangle \nu\left(\frac{1-\bar{z}}{\neff}\right)}{-\frac{\nu\left(\bar{z}-z\right)}{\neff}}
\]
which can be simplified to
\[
\frac{1}{\alpha}\oint\frac{dz}{2\pi iz}\frac{\left\langle g_{k}\left(\overline{z}\right)\right\rangle \left(1-\bar{z}\right)-\left\langle g_{k}\left(z\right)\right\rangle \left(1-z\right)}{\bar{z}-z}
\]
using \ref{eq:identity_iv} again in the form $\left\langle g_{k}\right\rangle _{k}=\nu\left(\hat{R}_{3}\left(z\right)-1\right)$
\[
\hat{\mathcal{K}}=\frac{\nu}{\alpha}\oint\frac{dz}{2\pi iz}\frac{\left(\hat{R}_{3}\left(\bar{z}\right)-1\right)\left(1-\bar{z}\right)-\left(\hat{R}_{3}\left(z\right)-1\right)\left(1-z\right)}{\bar{z}-z}.
\]

\subsubsection{Evaluating the contour integral for $\protect\hK$ at $\zeta=1$}

We can expand $\hR$ in a Laurent series back to time domain $\hR=\sum_{t=0}^{1}R_{t}z^{-t}=1+\sum_{t=1}^{1}R_{t}z^{-t}$
, additionally at $\zeta=1$ we have $\bar{z}=z^{-1}$, so after some simplification the expression becomes
\[
\hat{\mathcal{K}}\left(\zeta=1\right)=-\frac{\nu}{\alpha}\oint\frac{dz}{2\pi iz}\frac{\sum_{t=1}^{\infty}R_{t}z^{t}+\sum_{t=1}^{1}R_{t}z^{-\left(t-1\right)}}{1+z}
\]
pulling out a shared $\sum_{t=1}^{1}R_{t}z^{-\left(t-1\right)}$ and expanding out the polynomial
\[
-\frac{\nu}{\alpha}\sum_{t=1}^{\infty}R_{t}z^{-\left(t-1\right)}\oint\frac{dz}{2\pi iz}\frac{z^{2t-1}+1}{1+z}=-\frac{\nu}{\alpha}\sum_{t=1}^{\infty}R_{t}\sum_{j=0}^{2t-2}\left(-1\right)^{j}\oint\frac{dz}{2\pi iz}z^{j-t+1}
\]
which we can finally directly integrate using $\oint\frac{d\theta}{2\pi}e^{i\theta\left(j-t+1\right)}=\delta_{j-t+1,0}$
with $\delta$ the Kronecker delta which gives after the $j$ summation
and pushing in the minus from outside
\[
\frac{\nu}{\alpha}\sum_{t=1}^{\infty}R_{t}\left(-1\right)^{t}=\frac{\nu}{\alpha}\left(\hR\left(-1\right)-1\right)
\]
which we recognize at $\hR-1$ evaluated at at $z=-1$. Using identity
\ref{eq:identity_iv} once more we reach the final result
\[
\hat{\mathcal{K}}\left(\zeta=1\right)=\frac{\left\langle g_{k}\left(z=-1\right)\right\rangle _{k}}{\alpha}.
\]

\subsubsection{Evaluating the contour integral for $\protect\hT$ at $\zeta=1$}

Expanding in a Laurent series again
\[
\hT=\oint\frac{dz}{2\pi iz}\frac{\left|z-1\right|^{2}}{\neff}\left(\frac{\sum_{t=1}^{\infty}R_{t}z^{t}-\sum_{t=1}^{\infty}R_{t}z^{-t}}{\bar{z}-z}\right)=\frac{1}{\neff}\sum_{t=1}^{\infty}R_{t}\oint\frac{dz}{2\pi iz}\frac{\left(z-1\right)\left(z^{-1}-1\right)}{\left(z^{-1}-z\right)}\left(z^{t}-z^{-t}\right)
\]
calculating the integral explicitly again and resumming we arrive
at 
\[
\hT\left(\zeta=1\right)=\frac{2\left(\hR\left(-1\right)-1\right)}{\neff}=\frac{2\left\langle g_{k}\right\rangle _{k}}{\nu\neff}
\]

\subsection{Width Limited Pole $\lim _{\zeta \to 1} \left|e^{i\theta}-1+\lambda_{k}\eta_{\rm{eff}}\hat{R}_{3}\right|^{2}=0$}

This term can vanish in two separate ways, either for a specific $k$, or for all $k$. 
The former admits no consistent solution
and here we focus here on the latter.
For this expression to vanish for all $k$, we require  $e^{i\theta}-1$ and $\neff \hat{R}_{3}$ vanish simultaneously.

Requiring $e^{i\theta}-1$ vanishes immediately implies $\theta=0$ or equivalently $z=1$. Using the notation
$r_:=R_{3}\left(1\right)$ 
we find two cases. For $r>0$ substituting z=1 into 
\[ \hat{R}_{3}\left(z\right)=\left(1+\frac{1}{\nu}\left\langle \frac{\lambda_{k}\eta_{\text{eff}}}{z-1+\lambda_{k}\eta_{\text{eff}}\hat{R}_{3}\left(z\right)}\right\rangle _{k}\right)^{-1}
\]
we immediately find
\[
r=1-\frac{1}{\nu}
\]
so a solution exists only for $\nu>1$. Another solution can be found for $r=\rho\left(z-1\right)$ in this case 
\[
1=\lim_{z\to1}\rho\left(z-1\right)+\frac{1}{\nu}\left\langle \frac{\lambda_{k}\eta_{\text{eff}}\rho\left(z-1\right)}{z-1+\lambda_{k}\eta_{\text{eff}}\rho\left(z-1\right)}\right\rangle _{k}
\]
which gives the condition
\[
\left\langle \frac{\lambda_{k}\eta_{\text{eff}}\rho}{1+\lambda_{k}\eta_{\text{eff}}\rho}\right\rangle _{k}=\nu.
\]
The physical branch is the one where $\rho>0$ so this condition can only be satisfied for $\nu<1$ as the denominator is greater than the numerator and $\nu$ is a positive real quantity (with $\nu=1$ a borderline case giving a square root divergence).

The above conditions determine the product $\neff \rho$, but to isolate each one individually, we must determine $\neff$ from the residue of $\hat{c}_3$ at $\zeta=1$. Inspecting $\hat{\mathcal{S}}$, $\hat{\mathcal{T}}$, and $\hat{\mathcal{K}}$ the singularity cancels in all of them and only $\hat{\mathcal{B}}$ maintains a singular part, so
\[
\lim_{\zeta\rightarrow1}(\zeta-1)\hat{c}_{3}(\zeta)
    =\frac{\hat{\mathcal{T}}\left(1\right)}{1-\hat{\mathcal{K}}\left(1\right)}\lim_{\zeta\to1}\left(\zeta-1\right)\hat{\mathcal{B}}\left(\zeta\right)
    \nonumber.
\]
Evaluating it at the limit
\begin{align}
    \lim_{\zeta\to1}\left(\zeta-1\right)\hat{\mathcal{B}}\left(\zeta\right)&=\lim_{\zeta\to1}\left(\zeta-1\right)\frac{1}{\alpha}\intop_{0}^{2\pi}\frac{d\theta}{2\pi}\left(\left\langle \frac{\lambda_{k}\beta_{k}^{2}}{\left|1+\lambda_{k}\eta_{\text{eff}}\rho\right|^{2}}\right\rangle \frac{1}{\left|\sqrt{\zeta}e^{i\theta}-1\right|^{2}}
    \right.
    \\&\qquad\qquad\qquad\qquad
    \left.
    +\frac{\left\langle \frac{\lambda_{k}}{\left|1+\lambda_{k}\eta_{\text{eff}}\rho\right|^{2}}\right\rangle \eta_{\text{eff}}^{2}\rho^{2}}{\nu-\left\langle \frac{\lambda_{k}^{2}}{\left|1+\lambda_{k}\eta_{\text{eff}}\rho\right|^{2}}\right\rangle \eta_{\text{eff}}^{2}\rho^{2}}\left\langle \frac{\lambda_{k}^{2}\beta_{k}^{2}}{\left|1+\lambda_{k}\eta_{\text{eff}}\rho\right|^{2}}\right\rangle \frac{1}{\left|\sqrt{\zeta}e^{i\theta}-1\right|^{2}}\right)
    \nonumber
\end{align}
and after integration the limit yields
\begin{align}
    \lim_{\zeta\to1}\left(\zeta-1\right)\hat{\mathcal{B}}\left(\zeta\right)&=
    \frac{1}{\alpha}\left(\left\langle \frac{\lambda_{k}\beta_{k}^{2}}{\left(1+\lambda_{k}\eta_{\text{eff}}\rho\right)^{2}}\right\rangle +\frac{\left\langle \frac{\lambda_{k}}{\left(1+\lambda_{k}\eta_{\text{eff}}\rho\right)^{2}}\right\rangle \eta_{\text{eff}}^{2}\rho^{2}}{\nu-\left\langle \frac{\lambda_{k}^{2}}{\left(1+\lambda_{k}\eta_{\text{eff}}\rho\right)^{2}}\right\rangle \eta_{\text{eff}}^{2}\rho^{2}}\left\langle \frac{\lambda_{k}^{2}\beta_{k}^{2}}{\left(1+\lambda_{k}\eta_{\text{eff}}\rho\right)^{2}}\right\rangle \right)
    \nonumber\\
    &=\frac{1}{\alpha}\left\langle \frac{\lambda_{k}\beta_{k}^{2}}{1+\lambda_{k}\eta_{\text{eff}}\rho}\right\rangle. 
\end{align}
The final result is therefore
\begin{align}
    \lim_{\zeta\rightarrow1}(\zeta-1)\hat{c}_{3}(\zeta)
    &=\frac{2\left(\hat{R}_{3}(-1)-1\right)}{\alpha\eta_{\text{eff }}\left[1-\frac{\nu}{\alpha}\left(\hat{R}_{3}(-1)-1\right)\right]}\left\langle \frac{\lambda_{k}\beta_{k}^{2}}{1+\lambda_{k}\eta_{\text{eff }}\rho}\right\rangle _{k}
    \nonumber\\
    \nu 
    &=
    \left\langle \frac{\lambda_{k}\eta_{\text{eff}}\rho}{1+\lambda_{k}\eta_{\text{eff}}\rho}\right\rangle _{k}
    \nonumber\\
    \hat{R}_{3}\left(-1\right)&=\left(1+\frac{1}{\nu}\left\langle \frac{\lambda_{k}\eta_{\text{eff}}}{\lambda_{k}\eta_{\text{eff}}\hat{R}_{3}\left(-1\right) -2}\right\rangle _{k}\right)^{-1}.
\label{eq:bias_eta}
\end{align}

\subsection{Edge of Stability Pole $\lim _{\zeta \to 1} \nu-\left\langle \frac{\lambda_{k}^{2}}{\left|e^{i\theta}-1+\lambda_{k}\eta_{\rm{eff}}\hat{R}_{3}\right|^{2}}\right\rangle \eta_{\rm{eff}}^{2}\left|\hat{R}_{3}\right|^{2}=0$}

In this limit, we have
\[
\left\langle \frac{\lambda_{k}^{2}}{\left|e^{i\theta}-1+\lambda_{k}\eta_{\text{eff}}\hat{R}_{3}\right|^{2}}\right\rangle \eta_{\text{eff}}^{2}\left|\hat{R}_{3}\right|^{2}=\nu
\]

Using the identity \ref{eq:identity_iv}
\[
\frac{z-1}{\eta_{\text{eff}}}=\frac{z-1}{\eta_{\text{eff}}\hat{R}_{3}\left(z\right)}\left(1+\frac{1}{\nu}\left\langle \frac{\lambda_{k}}{\frac{1-z}{\eta_{\text{eff}}\hat{R}_{3}\left(z\right)}-\lambda_{k}}\right\rangle _{k}\right)
\]
and taking the imaginary part
\[
\Im\left[\frac{1-z}{\eta_{\text{eff}}}\right]=\Im\left[\frac{1-z}{\eta_{\text{eff}}\hat{R}_{3}\left(z\right)}\right]\left(\underbrace{1-\frac{1}{\nu}\eta_{\text{eff}}^{2}\left|\hat{R}_{3}\right|^{2}\left\langle \frac{\lambda_{k}^{2}}{\left|z-1+\lambda_{k}\eta_{\text{eff}}^{2}\left|\hat{R}_{3}\right|^{2}\right|^{2}}\right\rangle _{k}}_{=0}\right)=0,
\]
where the last term is zero by assumption, we find that
$\Im\left[\frac{1-z}{\eta_{\text{eff}}}\right]=0$, therefore $z=\pm1$, but only the case $z=-1$ gives a physically consistent solution.

Letting $r_{-}:=\hat{R}_{3}\left(z=-1\right)$ we find the condition
\[
\left\langle \frac{\lambda_{k}^{2}\eta_{\text{eff}}^{2}\left|r_{-}\right|^{2}}{\left|-2+\lambda_{k}\eta_{\text{eff}}r_{-}\right|^{2}}\right\rangle =\nu
\]
which is the condition for Phase 1 - EoS. This condition has to be solved together with the self-consistent equation for $\hR$ in Eq.~\ref{eq:R3_theta}
\[
r_{-}^{-1}=1+\frac{1}{\nu}\left\langle \frac{\lambda_{k}\eta_{\text{eff}}}{-2+\lambda_{k}\eta_{\text{eff}}r_{-}}\right\rangle _{k}.
\]
Practically, we can first solve for the product $\eta_{\text{eff}}r_{-}$, and then solve the second one for the ratio, which determines the physical $\neffi$. In Appendix~\ref{app:rmt_interp} we show that the result can be interpreted as $\neffi=2/\lambda_{\rm max}$ with $\lambda_{\rm max}$ the largest eigenvalue of the random matrix $A\tfrac{\Lambda}{N}A^T$.

\subsection{Edge of Stochastic Stability Pole $\lim _{\zeta \to 1}\hat{\mathcal{K}}\left(\zeta\right)-1=0$}

We already evaluated $\hK(\zeta=1)$ in Appendix~\ref{subapp:contour_int_eval} and found
\[
\hat{\mathcal{K}}\left(\zeta=1\right)=\frac{\left\langle g_{k}\left(z=-1\right)\right\rangle _{k}}{\alpha}.
\]
Our EoSS pole assumption thus directly implies
\[
\left\langle \frac{\lambda_{k}\eta_{\text{eff}}r_{-}}{2-\lambda_{k}\eta_{\text{eff}}r_{-}}\right\rangle _{k}\stackrel{!}{=}\alpha
\]
which has to be solved together with the $r_{-}$ equation
\[
r_{-}=\left(1+\frac{1}{\nu}\left\langle \frac{\lambda_{k}\eta_{\text{eff}}}{-2+\lambda_{k}\eta_{\text{eff}}r_{-}}\right\rangle _{k}\right)^{-1}.
\]
\[
\nu\left(1-r_{-}\right)=\left\langle \frac{\lambda_{k}\eta_{\text{eff}}r_{-}}{-2+\lambda_{k}\eta_{\text{eff}}r_{-}}\right\rangle _{k}.
\]

Given the relation 
\[
\lim_{\zeta\to1} (\zeta-1) c_3 = \hT(1) \lim_{\zeta\to1} (\zeta-1) c_1
\]
We can also deduce the final loss using
\begin{equation}
\lim_{\zeta\to1} (\zeta-1) c_1 = \frac{1}{\hT(1)} \frac{\eta^2}{\neff^2}
\label{eq:EoSS_loss_floor}
\end{equation}
with the $\neff$ that solves the condition above.

\section{Short Derivation of the GF Scaling Law}\label{app:transient_normSGD}
Here we give a simple heuristic derivation for the scaling law $L(\tau)\sim \tau^{r_{\rm SGD}}=\tau^{-\frac{2a+b-1}{b}}$.
Writing the loss as a sum over spectral modes $e_k(t):=w^*_k-w_k(t)$ and using the solution $e_k(t) = e_k(0) \exp(-2 \lambda_k t)$
\[
    L=\frac{1}{2}\sum_{k}\lambda_{k}e_{k}^{2}(t)=\sum_{k}\lambda_{k}e_{k}^{2} (0)\exp\left(-2\lambda_{k}\tau\right)
\]
and plugging in $\lambda_k = k^{-b}$, $e_k(0) = k^{-a}$ we find
\[
\sum_{k}k^{-2a-b}\exp\left(-2k^{-b}\tau\right).
\]
From this we can conclude that at time $\tau$ approximately down to $\frac{1}{\tau}$ are learned. Plugging the eigenvalue scaling we find that modes up to $k_*$ are learned
\[
\tau_{k}\sim\frac{1}{2\lambda_{k}}\sim k^{b}\Rightarrow k_{*}=\tau^{1/b}.
\]
Summing only modes higher than $k_*$ we finally get an explicit scaling with time
\begin{align}
    L\left(\tau\right)&=
    \frac{1}{2}\sum_{k=k_{*}\left(\tau\right)}^{\infty}k^{-2a-b}
    \approx\frac{1}{2}\intop_{\tau^{1/b}}^{\infty}k^{-\left(2a+b\right)}\,dk
    =\frac{1}{2}\frac{1}{1-\left(2a+b\right)}\left[k^{-\left(2a+b-1\right)}\right]_{\tau^{1/b}}^{\infty}\\
    &\sim\tau^{-\frac{2a+b-1}{b}} = \tau^{-r_{\text{SGD}}}.
\end{align}
In the same way, the gradient norm in the $\tau$ time is
\begin{equation}
    C_3 (\tau) = \sum_k \lambda_k^2 e_k^2 (\tau) = \sum_k k^{-2(a+b)}\exp(-2k^{-b}\tau).
\end{equation}
Using the same cutoff $\kappa_\star \sim \tau^{1/b}$ leads to
\begin{equation}
    C_3(\tau) \approx \int_{k_\star}^\infty k^{-2(a+b)}dk \sim \tau^{-\frac{2a+2b-1}{b}}= \tau^{-(r_{\text{SGD}}+1)}.
\end{equation}
We now return to the physical time $t$ of normalized gradient flow. Since by construction $\frac{dt}{d\tau} = \sqrt{C_3(\tau)}$, therefore
\begin{equation}
    \frac{dt}{d\tau} = \tau^{-\frac{r_{\text{SGD}}+1}{2}}.
\end{equation}
The resulting relation between $\tau$ and $t$ depends on whether $r_{\mathrm{SGD}}$ is smaller than, equal to, or larger than one. Since
\[
r_{\mathrm{SGD}}-1=\frac{2a-1}{b},
\]
these cases correspond respectively to $a<1/2$, $a=1/2$, and $a>1/2$.
For $a<1/2$, we have $r_{\mathrm{SGD}}<1$, and hence
\[
t(\tau)\sim
\int^\tau du\,u^{-\frac{r_{\mathrm{SGD}}+1}{2}}
\sim\tau^{\frac{1-r_{\mathrm{SGD}}}{2}}.
\]
Therefore,
\[
\tau(t)\sim t^{\frac{2}{1-r_{\mathrm{SGD}}}},
\]
and substituting into $L(\tau)\sim\tau^{-r_{\mathrm{SGD}}}$ gives
\[
L(t)\sim t^{-\frac{2r_{\mathrm{SGD}}}{1-r_{\mathrm{SGD}}}}.
\]

At $a=1/2$, we have $r_{\mathrm{SGD}}=1$, so
\[
\frac{dt}{d\tau}\sim\frac1{\tau},
\qquad
t\sim\log\tau.
\]
Thus $\tau$ grows exponentially with physical time and the loss decays exponentially. Retaining the prefactor in the continuum integral gives
\[
C_3(\tau)\sim\frac{1}{4b\tau^2},
\qquad
\tau(t)\sim e^{2\sqrt b\,t},
\qquad
L(t)\sim e^{-2\sqrt b\,t}.
\]

Finally, for $a>1/2$, we have $r_{\mathrm{SGD}}>1$, and the clock integral converges to a finite value $T$. Its remaining tail satisfies
\[
T-t(\tau)
\sim
\int_\tau^\infty du\,u^{-\frac{r_{\mathrm{SGD}}+1}{2}}
\sim\tau^{-\frac{r_{\mathrm{SGD}}-1}{2}}.
\]
Consequently
\[
\tau(t)\sim
(T-t)^{-\frac{2}{r_{\mathrm{SGD}}-1}},
\]
and
\[
L(t)\sim
(T-t)^{\frac{2r_{\mathrm{SGD}}}{r_{\mathrm{SGD}}-1}}.
\]

Collecting the three regimes
\begin{equation}
L(t)\sim
\begin{cases}
(T-t)^{\frac{2r_{\mathrm{SGD}}}{r_{\mathrm{SGD}}-1}},
& a>\frac12,\\[2mm]
e^{-2\sqrt b\,t},
& a=\frac12,\\[2mm]
t^{-\frac{2r_{\mathrm{SGD}}}{1-r_{\mathrm{SGD}}}},
& a<\frac12.
\end{cases}
\end{equation}

\section{Effective Learning Rate Scaling with Batch Size and Width}
\subsection{Effective learning rate and batch scaling in Phase 2}
We derive the stationary effective learning rate $\neffi$ on the Phase 2 (EoSS) branch, and quantify its departure from linear scaling with batch size. That is, we quantify when increasing batch size can lead to the same loss with fewer optimization steps and nearly the same compute budget -- close to linear scaling of $\neffi$ with $\alpha$. We also show that super-linear scaling is never possible in our setting, meaning the best possible outcome is to break even in terms of compute. 

Concretely, we study how $\neffi$ varies with $\alpha = B/d$ at fixed $\nu = N/d$ and fixed input spectrum $\{ \lambda_k \}_k$. 

First, let $r_{-} = \hat{R}_3 (-1)$. The response equation and the Phase 2 condition together read
\begin{equation}
    \frac{1}{r_{-}} = 1-\frac{1}{\nu}\Big< \frac{\lambda_k \neff}{2-\lambda_k \neff r_-} \Big>_k, \quad \alpha =\Big< \frac{\lambda_k \neff}{2-\lambda_k \neff r_-} \Big>_k r_-. 
\end{equation}
Since $r_{-}$ is independent of $k$, this just leads to 
\begin{equation}
    r_{-} = 1+\frac{\alpha}{\nu}.
\end{equation}
Given that $\neffi>0$ and $\alpha \geq 0$, this implies that
\begin{equation}
    \alpha \geq \frac{\mu_1}{2}\neffi(1+\alpha/\nu),\quad 0\leq \neffi\leq \frac{2\alpha}{\mu_1(1+\alpha/\nu)}
\end{equation}
with $\mu_1 = \langle \lambda_k\rangle_k$. Thus, at fixed spectrum, the spectral products tend to zero and the denominators can be expanded geometrically. Keeping every term through order $\alpha^2$ gives
\begin{equation}
    \alpha = \frac{\mu_1}{2}\neffi + \frac{\mu_1}{2\nu}\alpha \neffi + \frac{\mu_2}{4}\neffi^2 + \mathcal{O}(\alpha^3).
\end{equation}
Matching the coefficients to obtain $\neffi(\alpha)$ yields to
\begin{equation}
    \neffi(\alpha) = \frac{2\alpha}{\mu_1}\left[1-\alpha \left( \frac{\mu_2}{\mu_1^2}+\frac{1}{\nu}\right) \right] + \mathcal{O}(\alpha^3).
\end{equation}
Then, the leading term in $\alpha$ sets the linear scaling while the quadratic term determines its first relative correction
\begin{equation}
    \frac{\neffi}{\eta^{\infty}_{\text{lin}}} = 1-C \alpha + \mathcal O (\alpha^2)
\end{equation}
with $C = \frac{\mu_2}{\mu_1^2}+1/\nu$. Notice the sign of this correction is always negative, so the growth of $\neffi$ is always sublinear, and larger $C$s give smaller tipping points or smaller batch size in the sense of~\citep{mccandlish2018empirical}.

The logarithmic slope measures the fractional increase in effective
learning rate produced by a fractional increase in batch size
\begin{equation}
    p(\alpha)
    = \frac{d\log\neffi}{d\log\alpha}.
\end{equation}
The equality of the two logarithmic derivatives follows from keeping
$d$ fixed. Linear scaling corresponds to $p=1$, whereas $0<p<1$
corresponds to sublinear growth. Differentiating the expansion above gives
\begin{equation}
    \frac{d\neffi}{d\alpha}
    = \frac{2}{\mu_1}
      \left[1-2C\alpha+\mathcal{O}(\alpha^2)\right]
\end{equation}
and therefore
\begin{equation}
    p(\alpha)
    = 1-C \alpha
      +\mathcal{O}(\alpha^2).
\end{equation}
Thus, the coefficient $C$ controls the initial departure from linear
scaling. We can trade off some compute efficiency for fewer optimization steps by introducing a small tolerance $\delta$; the local condition
$p\geq1-\delta$ gives
\begin{equation}
    \alpha_\delta
    = \frac{\delta}{\mu_2/\mu_1^2+1/\nu}
      +\mathcal{O}(\delta^2)
\end{equation}
where the expansion is taken as $\delta\to0$ at fixed spectrum and $\nu$; yielding a mathematically concrete notion of the tradeoff mentioned above.

For the power law spectrum $\lambda_k=k^{-b}$, define
\begin{equation}
    S_j(d,b)=\sum_{k=1}^{d}k^{-jb},
    \qquad
    \mu_1=\frac{S_1}{d},
    \qquad
    \frac{\mu_2}{\mu_1^2}=\frac{dS_2}{S_1^2}.
\end{equation}
Using the asymptotics of power-law sums, we obtain
\begin{equation}
    \frac{\mu_2}{\mu_1^2}\asymp
    \begin{cases}
        1,                 & 0\le b<\tfrac12\\
        \log d,            & b=\tfrac12\\
        d^{\,2b-1},        & \tfrac12<b<1\\
        d/(\log d)^2,      & b=1\\
        d,                 & b>1.
    \end{cases}
\end{equation}
Consequently, at fixed $\nu$, the coefficient of the initial slope
deficit $1-p$ remains finite for $b<1/2$, grows logarithmically for
$b=1/2$ and grows with dimension for $b>1/2$. The linear scaling window size is inversely proportional to that, i.e., macroscopic for $b<1/2$ and vanishing for $b>1$.

For completeness, the leading effective learning rate is
\begin{equation}
    \eta_{\mathrm{lin}}^\infty
    = \frac{2\alpha}{\mu_1}
    \sim
    \begin{cases}
        2(1-b)\alpha d^b, & 0\leq b<1 \\
        2\alpha d/\log d, & b=1 \\
        2\alpha d/\zeta(b), & b>1.
    \end{cases}
\end{equation}
Here $\zeta(b)=\sum_{k=1}^{\infty}k^{-b}$ denotes the Riemann
zeta function, whose value is finite for $\mathbb{R}\ni b>1$. Together, these expressions specify the leading effective learning rate and the dimension dependence of its first relative correction in every spectral regime.
\subsection{Critical batch ratio separating EoSS and EoS}
We determine the critical batch ratio $\alpha_c=B_c/d$ by imposing the Phase 2 and Phase 3 conditions simultaneously. We consider $\lambda_k=k^{-b}$ and derive large $d$ asymptotics at fixed $\nu>0$ and fixed $b$. 

At the transition, define
\begin{equation}
    s_c=\frac{\neffi(\alpha_c)}{2}
        \left(1+\frac{\alpha_c}{\nu}\right).
\end{equation}
The two phase conditions become
\begin{equation}
    \nu d
    =\sum_{k=1}^{d}
      \left(\frac{s_c}{k^b-s_c}\right)^2,
    \qquad
    \alpha_c
    =\frac1d\sum_{k=1}^{d}
      \frac{s_c}{k^b-s_c}.
    \label{eq:critical_batch_conditions}
\end{equation}
with $0<s_c<1$. These equations determine the transition within the stationary
stability-limited regime. For fixed $b>0$, we first separate the contribution of the largest
eigenvalue, corresponding to $k=1$. The squared phase condition reads
\begin{equation}
    \nu d
    =\left(\frac{s_c}{1-s_c}\right)^2
     +\sum_{k=2}^{d}\left(\frac{s_c}{k^b-s_c}\right)^2.
\end{equation}
The remaining squared terms grow more slowly than $d$. Thus, at large
$d$, the first term supplies the leading contribution, giving
\begin{equation}
    \frac{s_c}{1-s_c}\simeq\sqrt{\nu d},
    \qquad s_c\to1.
\end{equation}

We now use this result in the unsquared phase condition. Keeping the
top-mode contribution and setting $s_c\simeq1$ in the remaining terms
gives
\begin{equation}
    B_c
    \simeq\sqrt{\nu d}
          +\sum_{k=2}^{d}\frac{1}{k^b-1}.
\end{equation}
The first term is the contribution of the largest eigenvalue, while the
second collects the remaining spectrum.

For $0<b<1$, the spectral sum grows as $d^{1-b}/(1-b)$.
It grows logarithmically at $b=1$ and approaches a finite constant
for $b>1$. Comparing this contribution with $\sqrt{\nu d}$ gives
the leading critical batch ratio:
\begin{equation}
    \alpha_c(\nu;b,d)\simeq
    \begin{cases}
        \sqrt{\nu},
            & b=0 \\
        \dfrac{d^{-b}}{1-b},
            & 0<b<\tfrac12 \\
        \dfrac{2+\sqrt{\nu}}{\sqrt d},
            & b=\tfrac12 \\[8pt]
        \sqrt{\dfrac{\nu}{d}},
            & b>\tfrac12.
    \end{cases}
    \label{eq:critical_batch_cases}
\end{equation}
The flat-spectrum result $b=0$ is exact. The remaining expressions
are leading approximations as $d\to\infty$ at fixed $b>0$ and $\nu>0$. 

For $b<1/2$, the remaining spectrum dominates the critical batch,
and the dependence on $\nu$ enters at smaller order. At $b=1/2$, both contributions have the same order and must be added. For $b>1/2$, the largest eigenvalue determines the leading result $B_c\simeq\sqrt{\nu d}=\sqrt N$, while the dependence on $b$ appears in subleading corrections.

\begin{figure}[t]
    \centering
    \includegraphics[width=0.44\linewidth]{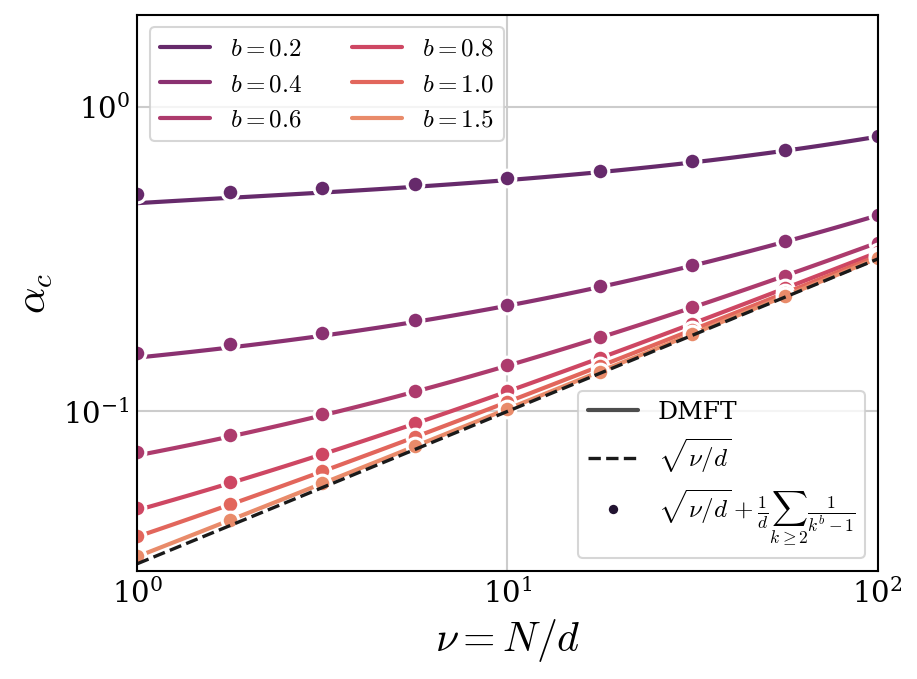}
    \caption{Critical batch ratio $\alpha_c$ versus width ratio $\nu$. Solid curves show DMFT predictions, the dashed line shows the leading top-mode estimate $\sqrt{\nu/d}$, and markers show the large-dimension approximation $\sqrt{\nu/d}+\frac{1}{d}\sum_{k=2}^{d}\frac{1}{k^b-1}$. The remaining spectrum raises the saturation threshold substantially for smaller $b$.}
    \label{fig:alpha_vs_nu}
\end{figure}

\subsection{Scaling with Width at Phase 1}
\label{appendix:width_speedup}

For the width-limited (Phase 1) branch, define
\begin{align}
    &\mu_1 = \frac{1}{d}\sum_k \lambda_k\\
    &\Phi (\neffi r_{-}) = \frac{1}{d}\sum_k \frac{\lambda_k \neffi r_{-}}{2-\lambda_k \neffi r_{-}}\\
    &P_{\nu} = \frac{1}{d}\sum_k \frac{\lambda_k \beta_k^2}{1+\lambda_k \neffi \rho}.
\end{align}
The three conditions to be satisfied can be then read as
\begin{align}
    &\nu = \frac{1}{d}\sum_k \frac{\lambda_k \neffi \rho}{1+\lambda_k \neffi \rho} \\
    &r_{-} = \Big(1-\frac{1}{\nu} \frac{\Phi (\neffi r_{-})}{r_{-}}\Big)^{-1}\\
    &\frac{\eta^2}{\nu \neffi} = \frac{2 (r_{-} -1)}{\alpha \Big( 1-\frac{\nu}{\alpha} (r_{-}-1)\Big)} P_{\nu}.
\end{align}
From the response function equation, one can then solve for $r_{-}$ and get 
\begin{equation}
    r_{-} = 1+ \frac{\Phi (\neffi r_{-})}{\nu}
\end{equation}
from which the third equation simplify as
\begin{equation}
    \alpha = \Phi (\neffi r_{-}) \Big(1+ \frac{2 \neffi P_{\nu}}{\eta^2}\Big).
\end{equation}
We first specify the physical branch on which the expansion is performed. Writing $y=\neffi r_-$, the EoS condition derived above is
\begin{equation}
    G(y_c)=\nu,
    \qquad
    G(y)=\frac1d\sum_k
    \left(\frac{\lambda_k y}{2-\lambda_k y}\right)^2.
\end{equation}
On $0\leq y<2/\lambda_{\max}$, $G(y)$ increases continuously from $0$ to $\infty$. Hence its unique root in this interval satisfies $0<y_c<2/\lambda_{\max}$. The stable Phase-1 branch lies below this EoS endpoint, so $0<y<y_c$ and therefore
\begin{equation}
    0<\lambda_k y<2
    \qquad\forall k.
\end{equation}
Each denominator in $\Phi(y)$ is thus positive and no larger than $2$, giving
\begin{equation}
    \Phi(y)
    =\frac1d\sum_k\frac{\lambda_k y}{2-\lambda_k y}
    \geq\frac1d\sum_k\frac{\lambda_k y}{2}
    =\frac{\mu_1y}{2}.
\end{equation}
Moreover, $\Phi(y)>0$ implies $r_-=1+\Phi(y)/\nu\geq1$. Substituting $y=\neffi r_-$ we obtain
\begin{equation}
    \Phi(\neffi r_-)
    \geq\frac{\mu_1\neffi r_-}{2}
    \geq\frac{\mu_1\neffi}{2}.
\end{equation}
This bound follows from the stability branch selection. Using this bound in the stationary closure gives
\begin{equation}
    \alpha\geq\frac{\mu_1P_\nu}{\eta^2}(\neffi)^2.
\end{equation}
Rearranging 
\begin{equation}
    0<\neffi\leq\eta\sqrt{\frac{\alpha}{\mu_1P_\nu}}.
\end{equation}
Since $\alpha$, $\mu_1>0$ and $P_\nu>0$ remain fixed,
$\eta\to0$ therefore implies $\neffi\to0$. The stationary closure gives $0\leq\Phi(y)\leq\alpha$, hence $1\leq r_-\leq1+\alpha/\nu$. Therefore $y=\neffi r_-\to0$, implying $\Phi(y)\to0$ and $r_-\to1$. Expanding $\Phi$ near zero then gives
\begin{equation}
    \frac{\Phi(y)}{\neffi}
    =\frac{\mu_1}{2}r_-+o(1)
    \longrightarrow\frac{\mu_1}{2}.
\end{equation}
Rearranging the exact stationary closure
\begin{equation}
    \frac{(\neffi)^2}{\eta^2}
    =\frac{\alpha-\Phi(y)}
    {2P_\nu\,[\Phi(y)/\neffi]}
    \longrightarrow\frac{\alpha}{\mu_1P_\nu},
\end{equation}
and taking the positive square root yields
\begin{equation}
    \neffi(\nu)\sim\eta\sqrt{\frac{\alpha}{\mu_1P_\nu}}.
\label{eq:small_eta_width_limited_neff}
\end{equation}

Consequently, the leading width efficiency is
\begin{equation}
    Q_\nu(\nu)=\frac{\neffi(\nu)}{\nu}
    \sim \frac{\eta}{\nu}
    \sqrt{\frac{\alpha}{\mu_1P_\nu}}.
\end{equation}
Since at fixed input spectrum, $\mu_1$ is independent of $\nu$, the remaining width dependence is determined by $P_\nu$.

For the power-law spectrum
\begin{equation}
    \mu_1 = \begin{cases}
        \frac{d^{-b}}{1-b}, &0\leq b < 1\\
        \frac{\log d}{d}, &b = 1\\
        \frac{\zeta(b)}{d}, &b>1.
    \end{cases}
\end{equation}
This is the same $\mu_1$ that appears in the small-$\alpha$ EoSS branch calculation. 
To extract the width dependence, we now specialize to
$\lambda_k=k^{-b}$ and $\beta_k=k^{-a}$.
Writing $m_\nu=\neffi\rho$, the width equation and the
residual target power become
\begin{align}
    \nu &= \frac{1}{d}\sum_{k=1}^{d}
    \frac{m_\nu}{k^b+m_\nu}\\
    P_\nu &= \frac{1}{d}\sum_{k=1}^{d}
    \frac{k^{-2a}}{k^b+m_\nu}.
\end{align}
The first equation determines $m_\nu$ independently of the
teacher exponent $a$. Once $m_\nu$ is known, the second
equation determines the dependence on both $a$ and $b$.

For $\lambda_k=k^{-b}$, the denominator $k^b+m_\nu$ changes behavior around $k\sim m_\nu^{1/b}$: below this index it is dominated by $m_\nu$, and above it by $k^b$. To obtain asymptotic width scalings, we focus on an intermediate-width regime where this crossover is well separated from both spectral endpoints, i.e.
\begin{equation}
    1\ll m_\nu^{1/b}\ll d.
\end{equation}

Both ranges then contain many modes, allowing their contributions to the spectral sums to be estimated separately. This condition specifies a window of $\nu$ given $d$, it is not required by the full stationary equations.

For modes with $k\ll m_\nu^{1/b}$, the denominator is
dominated by $m_\nu$, whereas for $k\gg m_\nu^{1/b}$
it is dominated by $k^b$. Splitting the sums into these
two ranges gives the scaling estimates
\begin{align}
    d\nu &\asymp m_\nu^{1/b}
    +m_\nu\sum_{k\gtrsim m_\nu^{1/b}}^d k^{-b},\\
    dP_\nu &\asymp
    \frac{1}{m_\nu}\sum_{k\lesssim m_\nu^{1/b}}k^{-2a}
    +\sum_{k\gtrsim m_\nu^{1/b}}^d k^{-(2a+b)}.
\end{align}
Here $\asymp$ denotes equality of leading scaling,
without specifying the multiplicative constant.

Now, to determine $m_\nu$, we approximate the width equation by
\begin{equation}
    d\nu \asymp m_\nu^{1/b}
    +m_\nu\int_{m_\nu^{1/b}}^d k^{-b}\,dk.
\end{equation}
For $0<b<1$, the integral is dominated by its upper endpoint
and gives $m_\nu d^{1-b}$, which dominates the first term.

\begin{figure}[h]
    \centering
    \includegraphics[width=0.67\linewidth]{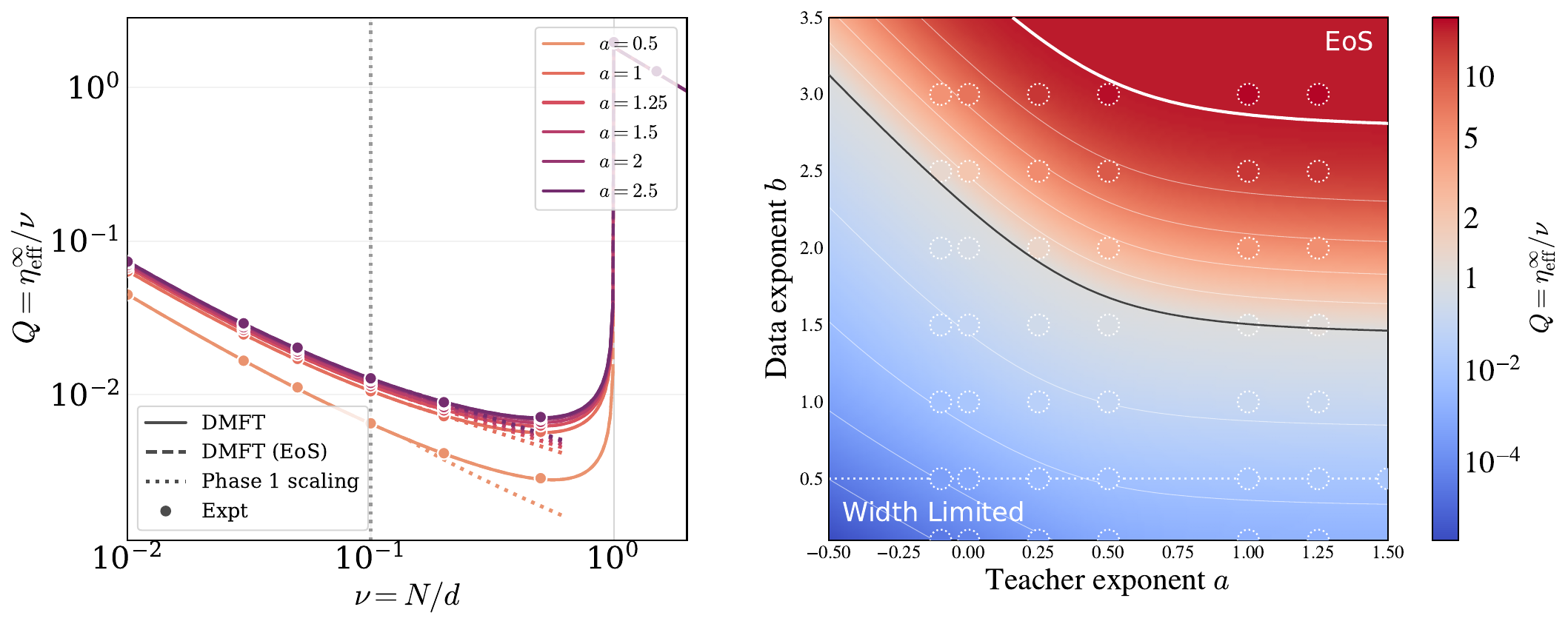}
    \caption{\textbf{Width efficiency and spectral dependence.} \textbf{Left:} $Q_\nu=\eta_{\mathrm{eff}}^\infty/\nu$ versus width for $b=0.5$ and different teacher exponents $a$. DMFT predictions (solid; dashed in EoS) are compared with simulations (symbols) and anchored Phase-1 scalings (dotted), which predict $Q_\nu\propto\nu^{-1/2}$ for $a>1/2$, with a logarithmic correction at $a=1/2$. \textbf{Right:} $Q_\nu$ across $(a,b)$ at $\nu=0.1$, with theory and simulation markers sharing the color scale. The thick white curve marks the EoS boundary; the dark contour marks $Q_\nu=1$. For these parameters, no interior efficiency maximum is observed for $\nu<1$: $Q_\nu(\nu)$ decreases to a minimum before rising toward a peak near the interpolation threshold $\nu=1$.
}
    \label{fig:width_b_0.5}
\end{figure}

\begin{figure}[h]
    \centering
    \includegraphics[width=0.67\linewidth]{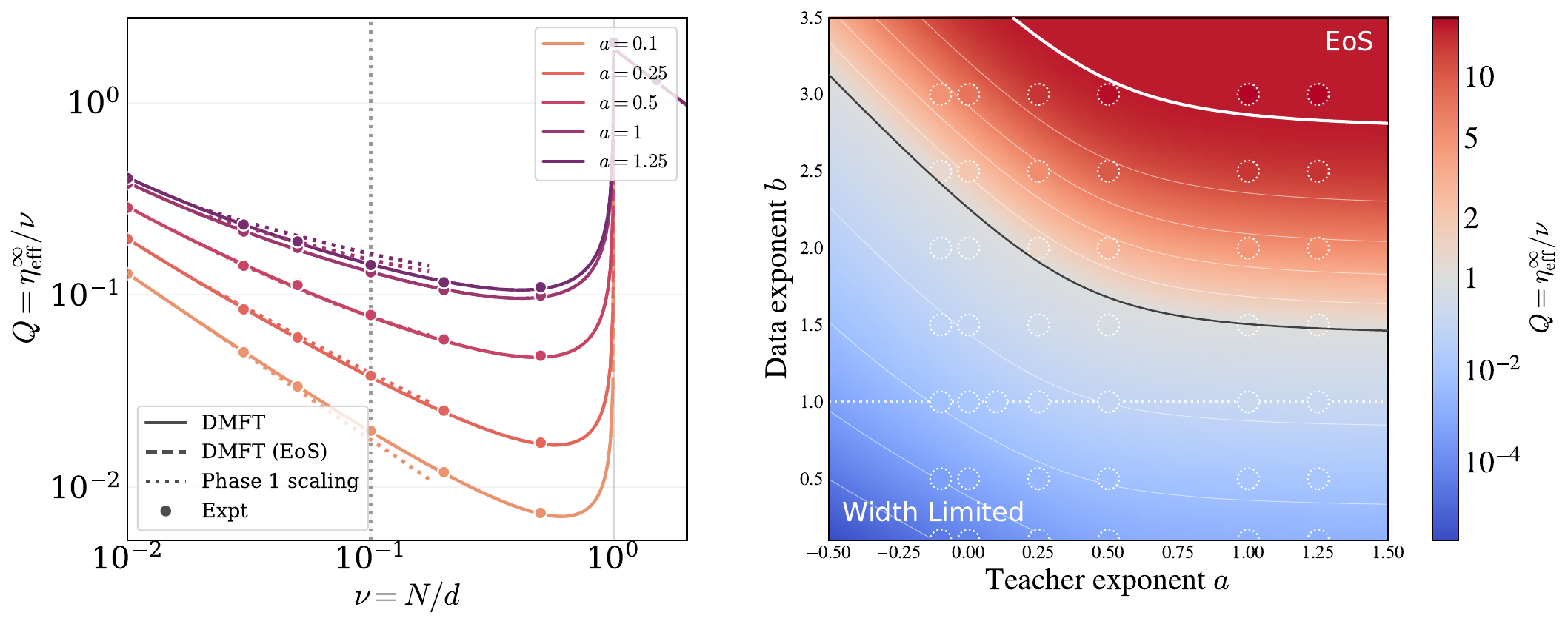}
    \caption{\textbf{Width efficiency at $b=1$.} \textbf{Left:} $Q_\nu=\eta_{\mathrm{eff}}^\infty/\nu$ versus width for different teacher exponents $a$. DMFT predictions (solid; dashed in EoS) agree with simulations (symbols). Dotted curves show anchored Phase-1 scalings, including the logarithmic corrections at $b=1$; for $a>1/2$, $Q_\nu\propto[\nu\log(1/\nu)]^{-1/2}$. Efficiency initially decreases, reaches a minimum, and rises toward a peak near $\nu=1$. \textbf{Right:} $Q_\nu$ across $(a,b)$ at $\nu=0.1$, with theory and simulation markers sharing the color scale. The thick white curve marks the EoS boundary, and the dark contour marks $Q_\nu=1$. The vertical and horizontal dotted reference lines indicate $\nu=0.1$ and $b=1$, respectively.}
    \label{fig:width_b_1}
\end{figure}

\begin{figure}[h]
    \centering
    \includegraphics[width=0.45\linewidth]{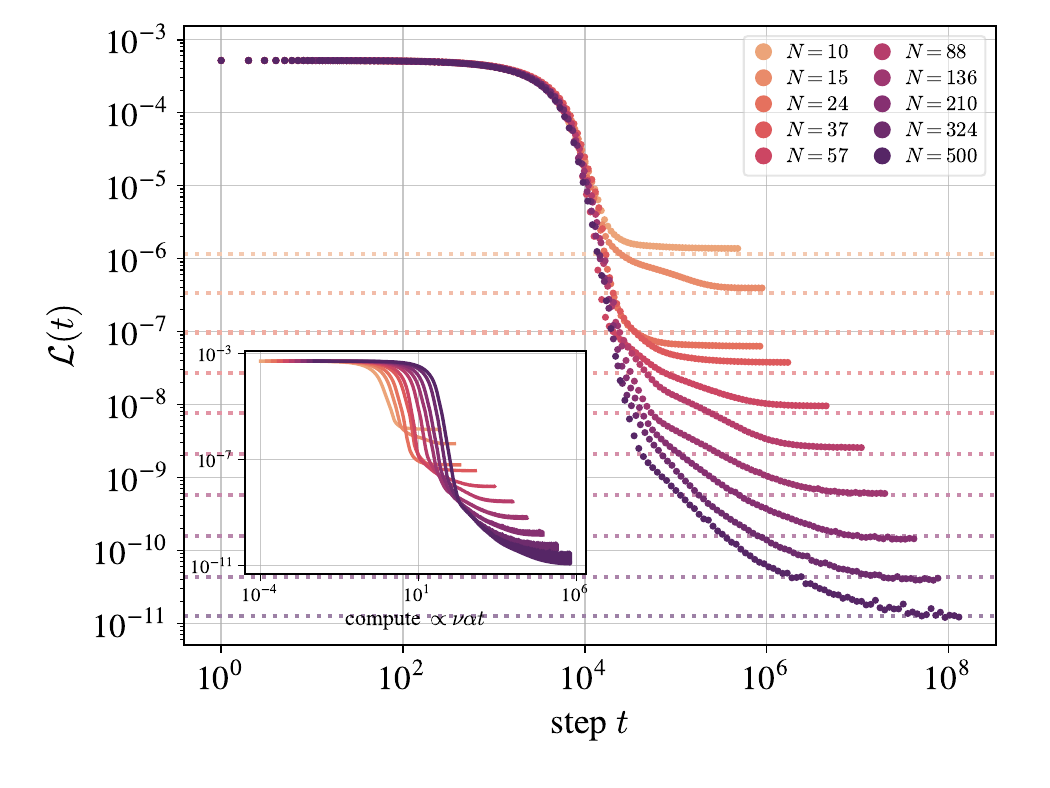}
    \caption{Complimentary figure to~\ref{fig:width_dep} right panel. Here we train a larger range of widths to convergence, showing that the width speedup can sometimes be subsumed by the loss floor, making larger widths for compute and speed efficient.} 
    \label{fig:width_speed_all}
\end{figure}

For $b=1$, it gives $m_\nu\log(d/m_\nu)$.
For $b>1$, the lower endpoint dominates, giving a contribution
of order $m_\nu^{1/b}$, comparable to the first term.
Therefore,
\begin{equation}\label{eq::m_nu}
    \nu \asymp
    \begin{cases}
        m_\nu d^{-b}, & 0<b<1\\
        \displaystyle \frac{m_\nu}{d}\log\!\left(\frac{d}{m_\nu}\right), & b=1\\
        \displaystyle \frac{m_\nu^{1/b}}{d}, & b>1.
    \end{cases}
\end{equation}
Inverting these relations yields
\begin{equation}
    m_\nu \asymp
    \begin{cases}
        \nu d^b, & 0<b<1\\
        \displaystyle \frac{\nu d}{\log(1/\nu)}, & b=1\\
        (\nu d)^b, & b>1.
    \end{cases}
\end{equation}
We next evaluate the two contributions to $P_\nu$.
The contribution from modes below the crossover satisfies
\begin{equation}
    \frac{1}{m_\nu}\sum_{k\lesssim m_\nu^{1/b}}k^{-2a}
    \asymp
    \begin{cases}
        m_\nu^{(1-2a-b)/b}, & a<1/2\\[0.4em]
        m_\nu^{-1}\log m_\nu, & a=1/2\\[0.4em]
        m_\nu^{-1}, & a>1/2.
    \end{cases}
\end{equation}
Indeed, the sum grows as its upper cutoff to the power
$1-2a$ for $a<1/2$, grows logarithmically for $a=1/2$,
and converges for $a>1/2$.

We assume $2a+b>1$, ensuring that
$\sum_k\lambda_k\beta_k^2$ converges as $d\to\infty$.
Consequently, the contribution from modes above the crossover is
\begin{equation}
    \sum_{k\gtrsim m_\nu^{1/b}}^d k^{-(2a+b)}
    \asymp m_\nu^{-(2a+b-1)/b}.
\end{equation}
Comparing the two contributions, we obtain
\begin{equation}
    P_\nu\asymp\frac1d
    \begin{cases}
        m_\nu^{-(2a+b-1)/b}, & \displaystyle \frac{1-b}{2}<a<\frac12\\[0.6em]
        m_\nu^{-1}\log m_\nu, & a=\frac12\\[0.4em]
        m_\nu^{-1}, & a>\frac12.
    \end{cases}
\end{equation}

Substituting the residual-power scaling into
$Q_\nu(\nu)$ gives
\begin{equation}
    Q_\nu(\nu)\asymp
    \frac{\eta}{\nu}\sqrt{\frac{\alpha d}{\mu_1}}
    \begin{cases}
        m_\nu^{(2a+b-1)/(2b)}, & (1-b)/2<a<1/2\\
        \sqrt{m_\nu/\log m_\nu}, & a=1/2\\
        \sqrt{m_\nu}, & a>1/2.
    \end{cases}
\end{equation}
For $b>1$, substituting the corresponding $m_\nu$ form Eq.~\ref{eq::m_nu} yields, at fixed $d,\alpha,\eta$
\begin{equation}
    Q_\nu(\nu)\asymp
    \begin{cases}
        \nu^{(2a+b-3)/2}, & (1-b)/2<a<1/2\\
        \nu^{(b-2)/2}/\sqrt{\log(\nu d)}, & a=1/2\\
        \nu^{(b-2)/2}, & a>1/2
    \end{cases}
\end{equation}
up to factors independent of $\nu$. Thus efficiency grows with width
when $2a+b>3$ for $a<1/2$, or $b>2$ for $a>1/2$, with a
logarithmic correction at $a=1/2$. This scaling alone does not determine the optimal width. The gain is eventually reversed by stability: on the EoS branch, $Q_\nu'(\nu)<0$. When this branch is reached below $\nu=1$, the increasing and decreasing regimes imply an efficiency maximum at an underparameterized width. Its location is determined by the full stationary equations.

For $0<b<1$, substituting $m_\nu$ gives
\begin{equation}
    Q_\nu(\nu)\asymp
    \begin{cases}
        \nu^{(2a-b-1)/(2b)}, & (1-b)/2<a<1/2\\
        \displaystyle
        \frac{\nu^{-1/2}}{\sqrt{\log\!\big((1-b)\nu d^b\big)}},
        & a=1/2\\
        \nu^{-1/2}, & a>1/2.
    \end{cases}
\end{equation}
Thus all three regimes give decreasing $Q_\nu(\nu)$. The marginal spectrum $b=1$ requires a separate treatment. The width constraint becomes
\begin{equation}
    \nu d\sim m_\nu\log\!\left(\frac{d}{m_\nu}\right).
\end{equation}
Writing $\ell_\nu=\log(1/\nu)$, its leading logarithmic solution is
\begin{equation}
    m_\nu\sim\frac{\nu d}{\ell_\nu},
    \qquad \nu\ll1.
\end{equation}
Substitution then gives
\begin{equation}
    Q_\nu(\nu)\asymp
    \begin{cases}
        \displaystyle \frac{\nu^{a-1}}{\ell_\nu^a},
        & 0<a<1/2\\
        \displaystyle
        \frac{1}{\sqrt{\nu\ell_\nu
        \log\!\big(\nu d/\ell_\nu\big)}},
        & a=1/2\\
        \displaystyle \frac{1}{\sqrt{\nu\ell_\nu}},
        & a>1/2.
    \end{cases}
\end{equation}
Here too, efficiency decreases with width within the asymptotic regime and the negative powers of $\nu$ dominate the logarithmic corrections. 

\subsection{Absence of approximately equal-compute speedup.}
\label{subapp:no_speedup}
Approximately equal-compute speedup requires the stationary effective learning rate to grow approximately linearly with the resource being increased. We examine the complementary resource directions to those permitting this behavior.

\paragraph{Batch size in Phase 1.}
At fixed width and spectrum, the width constraint fixes $\neffi\rho$, so $P_\nu$ is independent of batch size. The small-step asymptotics in \eqref{eq:small_eta_width_limited_neff} therefore gives

$$
    \neffi\sim\eta\sqrt{\frac{\alpha}{\mu_1P_\nu}},
    \qquad
    \frac{\partial\log\neffi}{\partial\log\alpha}
    \longrightarrow\frac12.
$$

Thus the gain is square-root rather than approximately linear. For example, doubling the batch reduces the updates needed to accumulate the same effective time by only a factor $\sqrt{2}$, increasing total compute by a factor $\sqrt{2}$. Increasing batch size therefore does not provide approximately equal-compute speedup in this regime.

\paragraph{Width in Phase 2.}
At fixed batch size and spectrum, the EoSS condition $\Phi(y)=\alpha$ fixes $y$ independently of width. The response equation gives

$$
    \neffi=\frac{\nu y}{\nu+\alpha},
    \qquad
    \frac{\partial\log\neffi}{\partial\log\nu}
    =\frac{\alpha}{\nu+\alpha}.
$$

The physical Phase-2 branch satisfies $\nu>1$ and $G(y)<\nu$. Since $\Phi(y)^2\leq G(y)$, we obtain $\alpha<\sqrt{\nu}<\nu$, and hence

$$
    \frac{\partial\log\neffi}{\partial\log\nu}<\frac12.
$$

Width gains are therefore slower than square-root throughout Phase 2. In particular, doubling width while remaining in this phase increases the compute required for the same effective time by more than a factor $\sqrt{2}$.

\paragraph{Batch size and width in Phase 3.}
At fixed width, the deterministic stability threshold
$\neffi=2/\lambda_{\max}(\nu)$ is independent of batch size. Additional batching therefore provides no stationary speedup.

For width, the response equation and the EoS condition give

$$
    \neffi=\frac{\nu y}{\nu+\Phi(y)},
    \qquad
    G(y)=\nu.
$$

Using $y\Phi'(y)=\Phi(y)+G(y)$ and differentiating along the EoS branch yields

$$
    \frac{\partial\log\neffi}{\partial\log\nu}
    =\frac{\Phi(y)}{\nu+\Phi(y)}.
$$

For $\nu\geq1$, the inequality $\Phi(y)^2\leq G(y)=\nu$ immediately bounds this slope by $1/(1+\sqrt{\nu})\leq1/2$. For $\nu<1$, Phase 3 is reached from the width-limited branch. At the transition to EoS, the Phase-1 closure gives
\[
    \alpha
    =
    \Phi(y)\left(1+\frac{2\neffi P_\nu}{\eta^2}\right)
    \geq \Phi(y).
\]
Hence throughout Phase 3, $\Phi(y)\leq\alpha$. Assuming $N>B$, equivalently $\nu>\alpha$, we therefore obtain
\[
    \frac{\partial\log\neffi}{\partial\log\nu}
    =\frac{\Phi(y)}{\nu+\Phi(y)}
    \leq\frac{\alpha}{\nu+\alpha}
    <\frac12.
\]
Thus, also for $\nu<1$, increasing width yields at most square-root gains in the stationary effective learning rate when $N>B$. We do not treat here the case of an under-parameterized network where the batch size $\alpha$ is larger than the total parameter count $\nu$, and we do not exclude speedups in this case; instead we note that this case is far removed from modern practice.

\section{Experimental Details}
\label{app:experiments}

\paragraph{Task and data.} We train on CIFAR-5M with the binary
animate/inanimate target, $y=+1$ for classes $2$--$7$ and $y=-1$ otherwise, so
that $\mathbb{E}[y^{2}]=1$. We rescaled each coordinate's color scale to $[-1,1]$ and used online training with no repetition: each of the $5\times10^{6}$ training images is used exactly
once, so every step draws a fresh batch such that the total number of steps satisfies $T$ $BT=5\times10^{6}$ in every run.

\paragraph{Architecture.} We use the residual CNN
of~\citep{bordelonDynamicalModelNeural2024}: a $3\times3$ stem convolution with
$w$ channels, three residual groups of one block each with the third at stride
$2$, average pooling, and a scalar readout. A block is
$\mathrm{Relu}\to\mathrm{Conv}_{3\times3}\to\mathrm{Relu}\to\mathrm{Conv}_{3\times3}$
summed with its shortcut. The width $w$ is constant across groups and there is
no batch normalization. The number of trainable parameters is
\begin{equation}
    N(k)=63w^{2}+28w .
\end{equation}

\paragraph{Linearized training.} We train the linearization of the network
around its initialization, $f(\delta\bm\theta)=\bm J(\bm\theta_{0})\,
\delta\bm\theta$, on the half mean-squared error, with the normalized update
$\delta\bm\theta\leftarrow\delta\bm\theta-\eta\,\bm g/\|\bm g\|_{2}$. Accordingly we use
$\neff^{t}=\eta/\|\bm g^{t}\|_{2}$. We note that this follows our dimension-free update described in App.~\ref{app:d-free} and differs from our proportional regime where we used $\sqrt{d} \eta$ for the update.

\paragraph{Spectral exponents.} We take $\lambda_{k}$ and $\beta_{k}$ from the
analytic infinite-width NTK, evaluated on $M=16{,}384$ images and diagonalized in
numerically; we find
$\operatorname{tr}\bm\Lambda=0.690$ and by the defintion of the labels
$\sum_{k}\lambda_{k}\beta_{k}^{2}=\mathbb{E}[y^{2}]=1$. We fit $b$ from the decay
of $\lambda_{k}$ and $2a+b-1$ from the decay of the residual target power
\[
1-C(k):=\tfrac{\sum_{k'=k}^M \lambda_{k'}\beta_{k'}^{2}}{\sum_{k'=1}^M \lambda_{k'}\beta_{k'}^{2}},
\]
both over the window $k\in[20,512]$ where we find a good power fit with parameters
\begin{equation}
    b=1.81,
    \qquad
    2a+b-1=0.17,
    \qquad
    a=-0.32,
    \qquad
    \lambda_{\max}=0.44 ,
\end{equation}
close to the values reported in~\citep{bordelonDynamicalModelNeural2024} for the
same architecture. Hence $\rsgd=0.095$, with predicted transient exponent
$2\rsgd/(1-\rsgd)=0.21$.

The DMFT takes the measured spectrum for $k\le1024$ and the fitted power law beyond it, continued to $d=10^{10}$. We found that taking larger $d$ made a negligible difference. 

\paragraph{Sweep.} We vary width and batch size over

\begin{center}
\begin{tabular}{lccccc}
\hline
$w$ & $4$ & $16$ & $64$ & $128$ & $256$ \\
$N$ & $1{,}120$ & $16{,}576$ & $259{,}840$ & $1{,}035{,}776$ & $4{,}135{,}936$ \\
\hline
$B$ & $2$ & $10$ & $100$ & $1{,}000$ & $10{,}000$ \\
$T$ & $2.5\cdot10^{6}$ & $5\cdot10^{5}$ & $5\cdot10^{4}$ & $5\cdot10^{3}$ & $500$ \\
\hline
$\eta$ & $0.01$ & $0.1$ & $0.2$ & $0.5$  \\
\hline
\end{tabular}
\end{center}

\noindent
with two seeds per cell, for $241$ runs in total.

\paragraph{Evaluation.} We report the mean-squared error on a fixed held-out set
of $4{,}096$ images, evaluated at $150$ logarithmically spaced steps, and record
$\|\bm g^{t}\|_{2}$ and $\neff^{t}$ at every step. 
The DMFT is integrated to up to $T=3800$ steps due to memory constraints.

\section{The Dimension-Free Limit}
\label{app:d-free}

The main text focuses on the proportional limit $d,N,B\to\infty$ at fixed $\nu=N/d$ and $\alpha=B/d$, here we extend the results to the case of taking the limit $d\to\infty$ first, and only then taking $N,B\gg 1$ -- a dimension free setting. This limit corresponds to the kernel settings and our NTK and linearized networks experiments where the representations live in an infinite-dimensional kernel space. We show that at fixed $N,B$ the
theory has a regular limit there in which the dimension drops out of every equation, leaving $N$, $B$, $\eta$ and the spectrum. We then translate the steady state of App.~\ref{app:pole_hunt} into this setting, where phase 1 (width-limited) and phase 3 (EoS) survive but phase 2 (EoSS) does not.
The main change is that we no longer average over $k$ modes, but take the trace (sun), so with $\lambda_{k}=k^{-b}$ and $\beta_{k}=k^{-a}$,
\begin{equation}
    \tr\Lambda=\zeta\left(b\right),
    \qquad
    \mathcal{L}^{0}=\sum_{k}\lambda_{k}\beta_{k}^{2}=\zeta\left(2a+b\right),
\end{equation}
and the limit exists for a trace-class kernel, $b>1$, alongside the usual $2a+b>1$.

To define this limit, we also need to modify the update and remove the $\sqrt d$ scaling,
\begin{equation}
    \bm w^{t+1}=\bm w^{t}-\eta\left(\bm g^{\top}\bm g\right)^{-1/2}\bm g,
    \qquad
    \eta_{\mathrm{eff}}^{t}=\frac{\eta}{\left\Vert \bm g^{t}\right\Vert _{2}}.
\end{equation}
which should be contrasted with \eqref{eq:model_def}.

\subsection{The Dimension-Free DMFT Equations}

In App.~\ref{subapp:DMFT_closed} the dimension enters only through
$\left\langle \cdot\right\rangle _{k}=d^{-1}\sum_{k}$ and the ratios $\nu,\alpha$, and these
appear only in the combinations $\nu^{-1}\left\langle \cdot\right\rangle _{k}$ and
$\alpha^{-1}\left\langle \cdot\right\rangle _{k}$. Multiplying every relation by $d$ therefore
removes the dimension and trades the intensive quantities for convergent ones (under this appendix's additional assumption $b>1$),
\begin{equation}
    \left\langle \cdot\right\rangle _{k}\mapsto\sum_{k},
    \quad
    \nu\mapsto N,
    \quad
    \alpha\mapsto B,
    \quad
    C_{1}\mapsto C_0/B,
    \quad
    \nu C_{3}\mapsto N C_3
    \quad
    dC_{2}\mapsto\Sigma_{2}.
\end{equation}
Keeping $H_{k}=\left[I+\lambda_{k}\Theta R_{3}\right]^{-1}$ and
$\Theta^{ts}=\eta\,1_{s<t}\left( N C_3^{ss}\right)^{-1/2}=\eta_{\mathrm{eff}}^{s}1_{s<t}$, the closed
set of App.~\ref{subapp:DMFT_closed} becomes
\begin{align}
    R_{3}&=I-\frac{1}{N}\sum_{k}\lambda_{k}\left(\Theta\left[I+\lambda_{k}R_{3}\Theta\right]^{-1}\right)R_{3}
    \nonumber\\
    _0&=\sum_{k}\lambda_{k}\left[\beta_{k}^{2}H_{k}j\left(H_{k}j\right)^{\top}
    +H_{k}\Theta\left(C_3+\lambda_{k}R_{3} c_1 \,R_{3}^{\top}\right)\left(H_{k}\Theta\right)^{\top}\right]
    \nonumber\\
    \Sigma_{2}&=\sum_{k}\left[\lambda_{k}^{2}\beta_{k}^{2}H_{k}j\left(H_{k}j\right)^{\top}
    +\lambda_{k}H_{k}\left(c_1 +\lambda_{k} \Theta C_3 \Theta^{\top}\right)H_{k}^{\top}\right],
    \qquad
\end{align}
with $j$ the all-ones time vector. These are the $\nu,\alpha\to0$ limit of the main-text closure at
fixed $N$ and $B$, and are exact for $1\ll N,B\ll d$. The same substitution in
~\eqref{eq:R3_theta} gives the stationary response
\begin{equation}
    \hat{R}_{3}\left(z\right)=\left(1+\frac{1}{N}\sum_{k}\frac{\lambda_{k}\neffi}{z-1+\lambda_{k}\neffi\hat{R}_{3}\left(z\right)}\right)^{-1}.
    \label{eq:dfree_R3}
\end{equation}

\subsection{Steady State and the Absence of the EoSS Phase}
Multiplying \eqref{eq:stability_phases} by $d$, the three conditions read
\begin{equation}
    \underbrace{\sum_{k}\frac{\lambda_{k} \neffi\rho}{1+\lambda_{k} \neffi\rho}=N}_{\text{Phase 1: width}},
    \qquad
    \underbrace{\operatorname{Tr} \left[ \frac{\Lambda\neffi\hR\left(-1\right)}{2-\Lambda \neffi\hR\left(-1\right)}\right] =B }_{\text{Phase 2: EoSS}},
    \qquad
    \underbrace{\operatorname{Tr} \left[ \left(\frac{\Lambda\neffi\hR\left(-1\right)}{2-\Lambda \neffi\hR\left(-1\right)} \right)^2 \right] =N}_{\text{Phase 3: EoS}},
\end{equation}
Taking the $z=-1$ response equation to the same limit gives
\begin{equation}
    \operatorname{Tr}\left[\frac{\Lambda\neffi\hR\left(-1\right)}{2-\Lambda\neffi\hR\left(-1\right)}\right]
    =N\left(\hR\left(-1\right)-1\right),
\end{equation}
so phase 2 reads $\hR\left(-1\right)=1+B/N$, while multiplying \eqref{eq:bias_eta} by $d$
and using $\neffi=\eta/\sqrt{NC_{3}^{\infty}}$ puts phase 1 in the form
\begin{equation}
    \eta^{2}=\frac{2\neffi N\left(\hR\left(-1\right)-1\right)}{B-N\left(\hR\left(-1\right)-1\right)}
    \sum_{k}\frac{\lambda_{k}\beta_{k}^{2}}{1+\lambda_{k}\neffi\rho},
    \label{eq:dfree_width}
\end{equation}
which always has a solution: its mode sum now runs over infinitely many modes and is no longer
bounded by $d$, which is the origin of the $\nu<1$ condition in the main text. The right-hand
side of \eqref{eq:dfree_width} grows continuously from zero and diverges precisely at
$\hR\left(-1\right)=1+B/N$, so it matches any $\eta$ at a smaller effective learning rate than
phase 2 requires. The width-limited state is therefore reached first for every $N$, $B$ and
$\eta$, and phase 2 is excluded. The only transition left is width-limited to EoS, at a
critical batch size obtained by evaluating \eqref{eq:dfree_width} on the EoS branch,
\begin{equation}
    B_{1 \to 3}=N\left(\hR\left(-1\right)-1\right)
    \left(1+\frac{2\neffi}{\eta^{2}}\sum_{k}\frac{\lambda_{k}\beta_{k}^{2}}{1+\lambda_{k}\neffi\rho}\right).
\end{equation}
\end{document}